\documentclass{article}

\usepackage[numbers]{natbib} 

\usepackage[final]{neurips_2025}

\usepackage[utf8]{inputenc} 
\usepackage[T1]{fontenc}    
\usepackage{hyperref}       
\usepackage{url}            
\usepackage{booktabs}       
\usepackage{amsfonts}       
\usepackage{nicefrac}       
\usepackage{microtype}      
\usepackage{xcolor}         
\usepackage{multirow}
\usepackage{amsmath}   
\usepackage{amssymb}   
\usepackage{graphicx}
\usepackage{subcaption}
\usepackage{array}
\usepackage{float}
\usepackage[linesnumbered,ruled,vlined]{algorithm2e}
\usepackage{algpseudocode}  
\usepackage{xcolor}         
\usepackage{listings}

\title{Balanced Conic Rectified Flow}

\author{%
  Shin seong Kim\\
  Yonsei University\\
  \texttt{tltydl2@yonsei.ac.kr}
  \And
  Mingi Kwon\\
  Yonsei University\\
  \texttt{kwonmingi@yonsei.ac.kr}
  \And
  Jaeseok Jeong\\
  Yonsei University\\
  \texttt{jete\_jeong@yonsei.ac.kr}
  \And
  Youngjung Uh\thanks{Corresponding author.} \\
  Yonsei University \\
  \texttt{yj.uh@yonsei.ac.kr}
}

\begin{document}

\maketitle

\begin{abstract}

Rectified flow is a generative model that learns smooth transport mappings between two distributions through an ordinary differential equation (ODE). The model learns a straight ODE by reflow steps which iteratively update the supervisory flow. It allows for a relatively simple and efficient generation of high-quality images. However, rectified flow still faces several challenges. 1) The reflow process is slow because it requires a large number of generated pairs to model the target distribution. 2) It is well known that the use of suboptimal fake samples in reflow can lead to performance degradation of the learned flow model. This issue is further exacerbated by error accumulation across reflow steps and model collapse in denoising autoencoder models caused by self-consuming training. In this work, we go one step further and empirically demonstrate that the reflow process causes the learned model to drift away from the target distribution, which in turn leads to a growing discrepancy in reconstruction error between fake and real images. We reveal the drift problem and design a new reflow step, namely the \textit{conic reflow}. It supervises the model by the inversions of real data points through the previously learned model and its interpolation with random initial points. Our conic reflow leads to multiple advantages. 1) It keeps the ODE paths toward real samples, evaluated by reconstruction. 2) We use only a small number of generated samples instead of large generated samples, 600K and 4M, respectively. 3) The learned model generates images with higher quality evaluated by FID, IS, and Recall. 4) The learned flow is more straight than others, evaluated by curvature. We achieve much lower FID in both one-step and full-step generation in CIFAR-10. The conic reflow generalizes to various datasets such as LSUN Bedroom and ImageNet. The project page is available at \url{https://grainsack.github.io/BC_rectified_flow_project_page/}.

\end{abstract}

\section{Introduction}
\label{Intro}
Rectified flow \citep{sd3,liu2023instaflow,lee2024improving, li2024flowdreamer, lee2023minimizing} demonstrates state-of-the-art image generation with fewer sampling steps than diffusion models \citep{de2021diffusion,vargas2021solving,song2020score,ho2020denoising,tzen2019theoretical}.
$k$-rectified flow involves $k$ reflow steps that make ODE paths smooth and straight \citep{liu2022flow}. This allows the model to generate high-quality images simply and efficiently in just one or a few steps.
Intriguingly, \textit{all rectified flow models} (Flux, SD3, and AuraFlow) achieve state-of-the-art quality with \textit{1-rectified flow} and require about 30 NFEs (number of function evaluations) \citep{sd3}.

In this paper, we identify key limitations of $k$-rectified flow stemming from its reflow procedure, and propose to use real images and their inversions, combined with a Slerp-based perturbation loss, to address them. Our core insight is that the degradation in performance cannot be fully explained by error accumulation or vanishing weight norms alone \citep{kim2024simple, zhu2024analyzing}. Instead, we observe that the reflow process itself induces a drift away from the target distribution, leading to a measurable discrepancy between the model’s behavior on real and fake samples, which we later analyze through reconstruction error. 1) \textbf{The flow drifts away from the real distribution.} For example, in training a 2-rectified flow, random noise vectors and their generated images from the 1-rectified flow are reused as supervisory targets, which causes the ODE paths to diverge from real data. We empirically show that this results in a reconstruction performance gap between real and fake samples, and that Slerp-based supervision for real samples helps maintain alignment with the true distribution. 2) \textbf{The number of fake samples required for reflow is prohibitively large.} Our method leverages real samples and their conic neighbors to guide the flow with far fewer generated pairs, while maintaining competitive or superior performance. This efficiency makes the method scalable and less dependent on synthetic supervision. 3) \textbf{The reflow process degrades image quality in full-step generation.} By providing trajectory-level supervision rooted in real data geometry, our method enhances generation quality across 1-step, few-step, and full-step regimes, and mitigates the degradation in standard reflow training.

As a result, we successfully demonstrate better performance than existing $k$-rectified flow models. On CIFAR-10, we reduce the FID (Fréchet Inception Distance, \citep{heusel2017gans}) of the existing 2-rectified flow from 12.21 to 5.98 while using only 7.2\% of the generative pair, and show that the curvature of the ODE paths became straighter. We also introduce a new method for calculating curvature, which explains the time distribution sampling method. We provide various ablation studies and show that even with simple fine-tuning, the performance of existing $k$-rectified flow can be significantly improved.

\begin{figure*}[t]
  \vspace*{0pt}  
  \centering
  \includegraphics[width=0.9\textwidth]{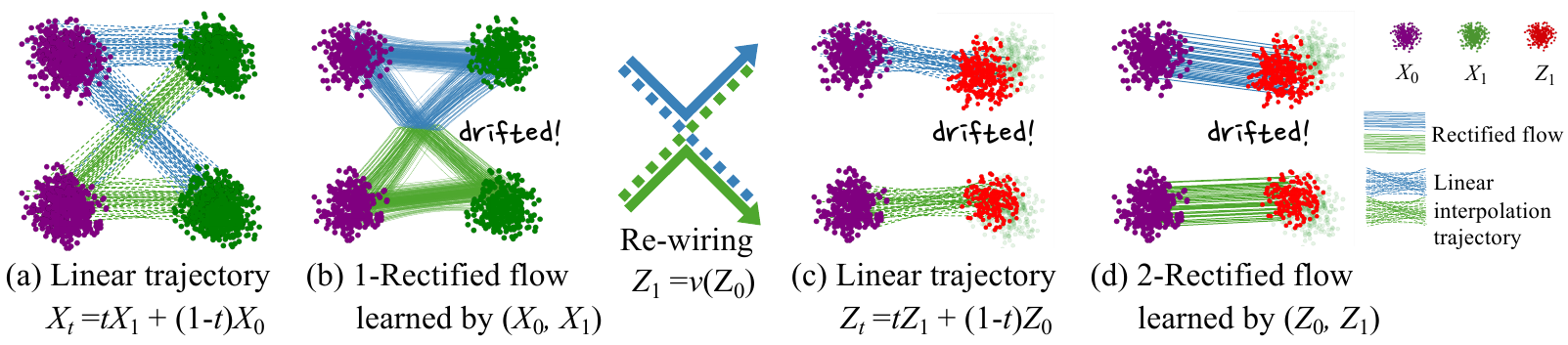}
  \caption{
    \textbf{Problem of rectified flow.}
    (a) By randomly pairing data $X_0 \sim \pi_0$ and $X_1 \sim \pi_1$, linear interpolation trajectories are defined.
    (b) The 1-rectified flow connects $X_0$ and $X_1$ with a learned velocity field which is potentially inaccurate. After the 1-rectified flow, the trajectories are rewired to avoid crossing.
    (c) The trajectories for reflow are defined as linear interpolation trajectories between $Z_0$ and the generated $Z_1 = v(Z_0)$. Note that $Z_1$ is drifted away from $\pi_1$.
    (d) Consequently, the 2-rectified flow has a velocity field drifted away from $X_1$.
  }
  \vspace{-1.5em}
  \label{fig:flow_overview}
\end{figure*}

\section{Rectified Flow}
\label{Rectified}

Rectified flow~\citep{liu2022flow} is a generative model that solves an ordinary differential equation (ODE) to induce a transition trajectory between two given data distributions $\pi_0$ and $\pi_1$. Data $X_0 \sim \pi_0$ and $X_1 \sim \pi_1$ define linear trajectories $X_t = (1 - t) X_0 + t X_1$ for $t \in [0, 1]$, as illustrated in Figure~\ref{fig:flow_overview}a. Then, a rectified flow $v$ is an ODE on time $t$ parameterized by $\theta$:
\vspace{-0.3em}

\begin{equation}
\frac{d Z_t}{dt} = v_\theta(Z_t, t) := \frac{1}{t}(Z_t - \mathbb{E}[(X_1 - X_0) | X_t = Z_t])
\end{equation}

\vspace{-0.3em}
We omit $\theta$ for brevity.
\citet{liu2022flow} propose a simplified mean squared error (MSE) loss for an ODE neural network to train velocity field $v : \mathbb{R}^n \rightarrow \mathbb{R}^n$ as follows:
\vspace{-0.5em}

\begin{align}
\label{eq:training}
\arg \min_{\theta} \mathbb{E} \Big[ \big\| X_1 - X_0 - v(t X_1 + (1 - t) X_0, t) \big\|^2 \Big]
\end{align}

\vspace{-0.5em}
$\text{With } t \sim \text{Uniform}([0, 1])$. In image generation tasks, $X_0 \sim \pi_0$ and $X_1 \sim \pi_1$ are random noise and real images, sampled from a Gaussian distribution and the data distribution, respectively. Once the model has learned the velocity field, the rectified flow rewires the trajectories in a non-crossing manner due to the inherent properties of ODEs, which enforce uniqueness and smoothness in trajectory evolution, preventing paths from intersecting in the continuous-time dynamics, as depicted in Figure~\ref{fig:flow_overview}b. It constitutes a 1-rectified flow model denoted by $\mathbf{Z} = \texttt{RectFlow}((X_0, X_1))$.

The $k$-rectified flow model learns a straighter sampling trajectory by repeating \textit{reflow procedure} k times as follows. 
Following $\mathbf{Z}^k$ from $Z_0^k$ induces a generated pair $(Z_0^k, Z_1^k)$ where $(Z_0^0, Z_1^0) = (X_0, X_1)$. It redefines the linear interpolation trajectory $\mathbf{Z}_t^{k+1} = (1 - t) Z_0^k + t Z_1^k$ for $t \in [0, 1]$, as shown in Figure~\ref{fig:flow_overview}c. Then, fine-tuning a velocity field $v$ using Eq.~\eqref{eq:training} with $(Z_0^k, Z_1^k)$ instead of $(X_0, X_1)$ constitutes $\mathbf{Z}^{k+1} = \texttt{RectFlow}((Z_0^k, Z_1^k))$.

According to optimal transport theory, \citep{OT1,OT2,OT3,OT4} coupling the generated pairs $(Z_0, v(Z_0))$ ensures that the interpolation trajectory preserves the marginal distributions of the original and target domains, as well as the linear interpolation trajectory between them \citep{kurtz2011equivalence,ambrosio2008existence}.
$k$-rectified flow has superior quality of few-step sampling by straighter sampling trajectory, as shown in Figure~\ref{fig:flow_overview}d.

    

\section{Improved Techniques for Reflow Step}

\begin{figure}[t!]
\centering
\vspace*{0pt} 

\begin{minipage}[t]{0.6\textwidth}
    \centering
    \includegraphics[width=\linewidth]{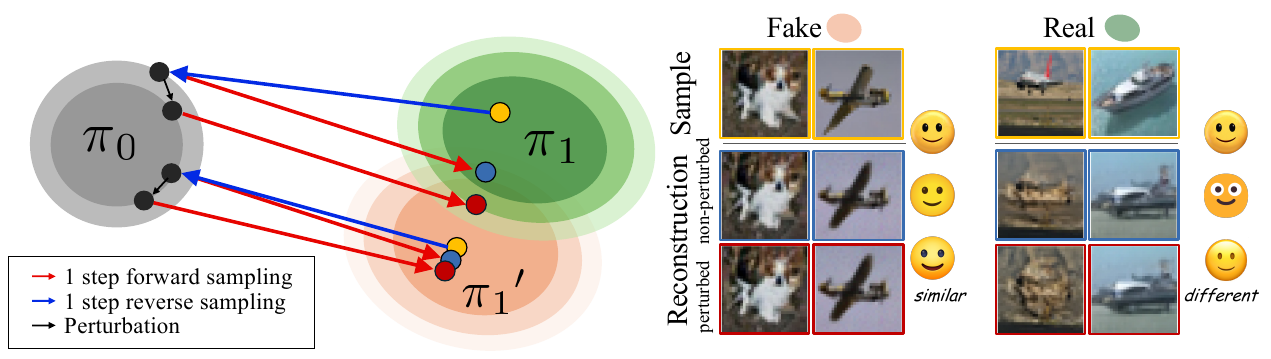}
\end{minipage}\hfill
\begin{minipage}[t]{0.4\textwidth}
    \centering
    \includegraphics[width=\linewidth]{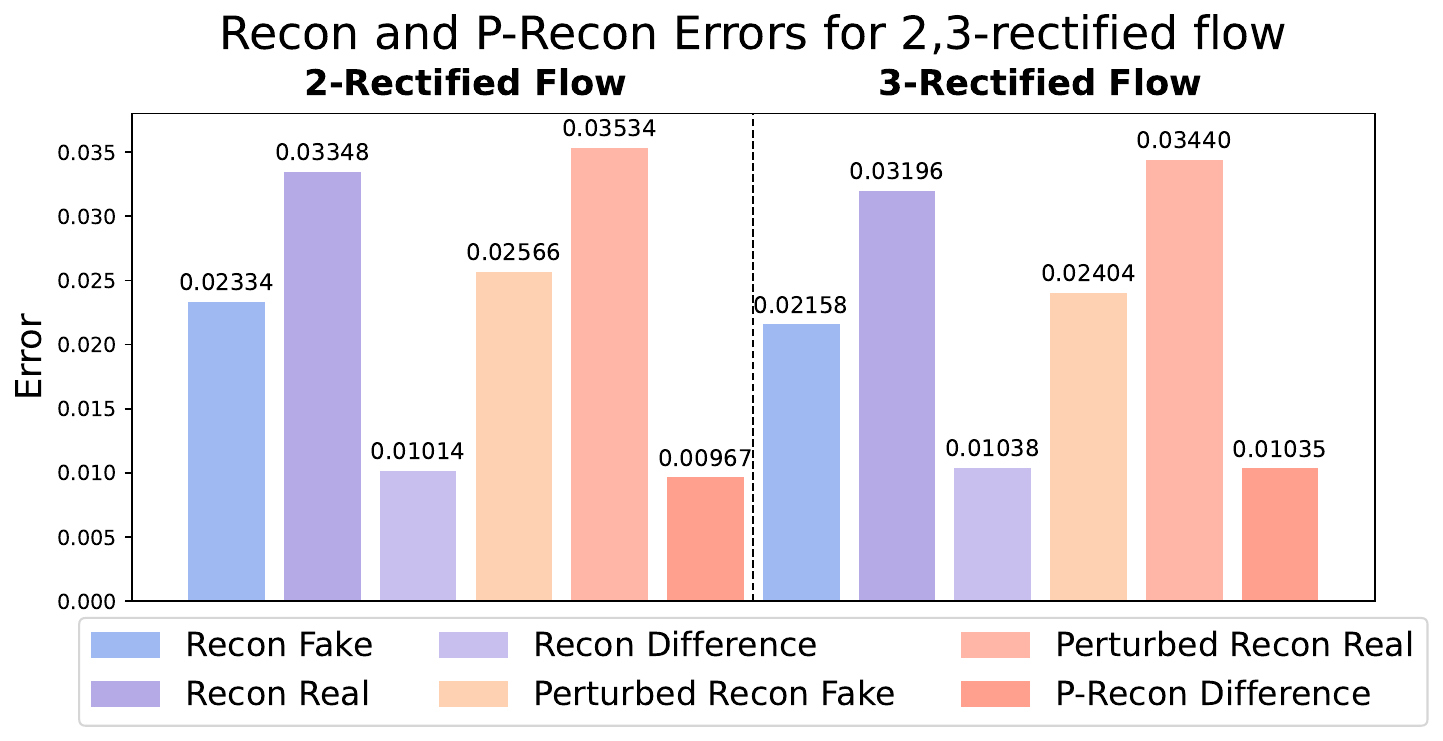}
\end{minipage}

\caption{\textbf{(a) 2-rectified flow overfits fake samples.}
Following the reverse and forward 2-rectified flow, fake images inherently return at similar images with or without perturbation at $\pi_0$. In contrast, real images return at different images and it is worse with perturbation, implying overfitting. \textbf{(b)} Reconstruction discrepancies emerge between real and fake images due to the use of fake-only pairs.}
\label{fig:compare_recon}
\vspace{-1em}
\end{figure}

In this section, we discuss the generated fake pairs in the original rectified flow and their problems. 
Then, we introduce real pairs and balanced-conic reflow which directly supervise the flow to reach the real data distribution. Finally, we provide detailed training configurations.

\subsection{Reflow steps drift the flow away from the real distribution.}
\label{sec:origin_reflow_drift}

As shown by \citet{liu2022flow}, the trajectory $Z_t^k$ between the generated pairs in the reflow process becomes smoother and straighter with each iteration of reflow, because the ODE induces a deterministic smooth solution while preserving the same marginal distribution as the original trajectory. This straightened path is essential for generating high-quality images with a small number of sampling steps rather than SDE-based generative models \citep{ho2020denoising,song2020score,song2020denoising}. We use subscript $F$ to denote the \textbf{fake pairs} from the original rectified flow as follows:
\begin{equation}
(Z_{0,F}^k, Z_{1,F}^k) := (Z_0^k, Z_1^k)
\end{equation}
$\text{Where } Z_0^k \sim \pi_{0} \text{ and } Z_1^k = v(Z_0^k).\footnote{For brevity, we denote the forward generation process at the t-th sampling step as $X_0 + \int^{1}_{0} v_{t}(X_t, \cdot) \, dt:= v(X_0)$ and backward process as $X_1 + \int^{1}_{0}v_{t}^{-1}(X_t, \cdot)\, dt:=v^{-1}(X_1)$, where $v^{-1} = -v$.}$ To simplify notation and avoid confusion, we will denote the fake pair as $(Z_{0, F}, Z_{1, F})$ when we do not need to consider the reflow step and, denote the $k$-th order of the rectified flow
as $(Z^k_{0,F}, Z_{1,F}^k)$. 

Beyond known issues such as error accumulation and model collapse \citep{kim2024simple, zhu2024analyzing}, we provide empirical evidence that reflow steps cause a distributional drift, observed through growing reconstruction discrepancies between real and fake samples. We provide empirical evidence of the accumulating drifts and suboptimality of the learned rectified flow. Figure~\ref{fig:compare_recon}a illustrates faithfulness and continuity\footnote{We use the notion of continuity as in Lipschitz continuity: the generated images should be similar with the similar latents.} of a 2-rectified flow on both fake samples and real samples. As expected, fake images are mostly reconstructed by inversion and generation following the 2-rectified flow, i.e., $Z_1 \simeq v(v^{-1}(Z_1))$. Also, an inversion of a fake sample and its perturbation land at similar images, i.e., $v(v^{-1}(Z_1)) \simeq v(v^{-1}(Z_1) + \varepsilon z)$ where we perturb it with $\varepsilon = 0.05$. In contrast, real images lose the main object by inversion and generation following the 2-rectified flow, i.e., $X_1 \neq v(v^{-1}(X_1))$. Furthermore, real images are vulnerable to perturbation on their inversion, i.e., $v(v^{-1}(X_1)) \neq v(v^{-1}(X_1) + \varepsilon z)$. To evaluate the faithfulness of a rectified flow to a dataset $X$, we measure the error between the sample and its reconstruction \citep{candes2006stable,ravishankar2019image} via the flow:
\vspace{-0.1em}
\begin{equation}
    L_2^\text{recon}(X) = \mathbb{E}_{x \sim X} \left[ \lVert x - v(v^{-1}(x)) \rVert_2 \right]
\end{equation}

\vspace{-0.1em}
Instead of Lipschitz continuity, we practically evaluate the continuity of a rectified flow near samples from a dataset $X$ by measuring a perturbed reconstruction error:
\vspace{-0.1em}
\begin{equation}
    L_2^\text{p-recon}(X, \varepsilon) = \mathbb{E}_{x \sim X, z \sim \pi_0}\lVert x - v(v^{-1}(x) + \varepsilon z) \rVert_2,
\end{equation}

\vspace{-0.1em}
where $\varepsilon$ is the strength of perturbation. 
The lower perturbed reconstruction error near the real samples $L_2^\text{p-recon}(X_1)$ indicates the more continuous generative model near the real samples.\footnote{We measure the reconstruction and perturbed reconstruction error with 1-step Euler sampling.} Figure~\ref{fig:compare_recon}b compares $L_2^\text{recon}$ of real and fake samples, and $L_2^\text{p-recon}$ near real and fake samples. $L_2^\text{recon}$ is higher at the real samples than the fake samples. It indicates that the 2,3-rectified flow drifts away from the real samples. Furthermore, $L_2^\text{p-recon}$ is lower near the fake samples than the real samples. It indicates that the 2-rectified flow suffers from crossing between real samples.

\begin{figure*}[t!]
    \centering
    \begin{minipage}[t]{0.5\textwidth}
        \centering
        \includegraphics[width=\linewidth]{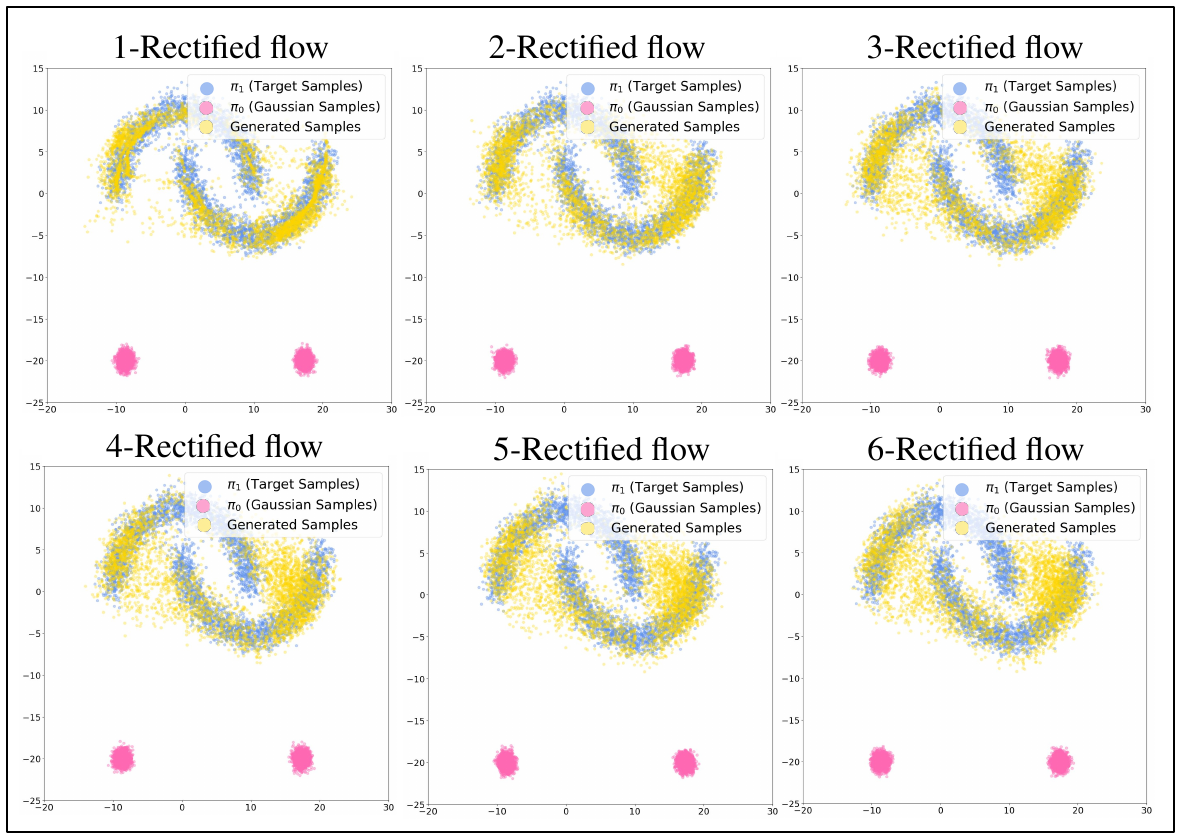}
    \end{minipage}\hfill
    \begin{minipage}[t]{0.49\textwidth}
        \centering
        \includegraphics[width=\linewidth]{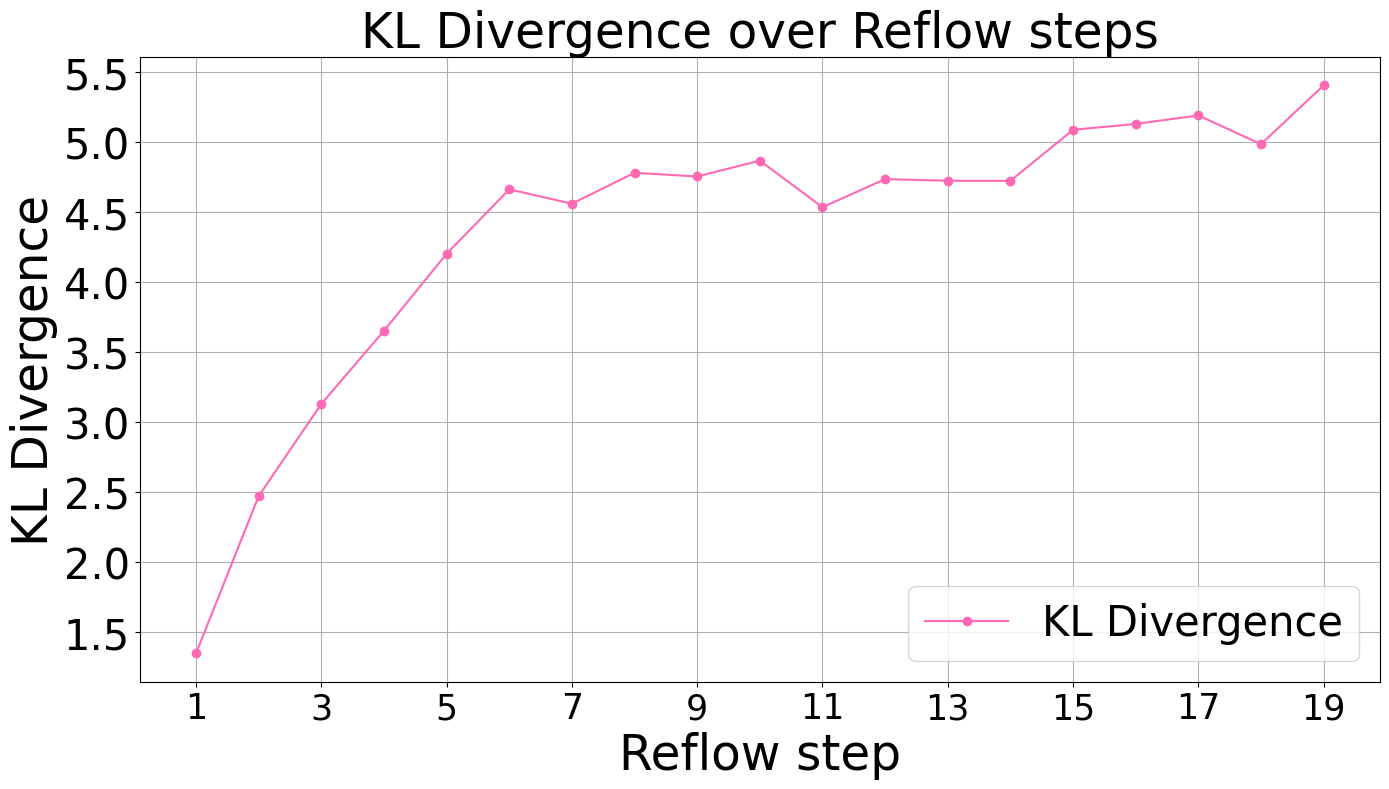}
    \end{minipage}
    \caption{
\textbf{(a)} As reflow steps increase, generated samples diverge from the target distribution. \textbf{(b)} This drift is further evidenced by the rising KL divergence from the real data distribution.}
    \vspace{-0.5em}
    \label{fig:reflow_drift_all}
\end{figure*}

Critically, the original reflow accumulates the drift over recursive reflow steps. This drift is an innate phenomenon because the supervision from a shifted distribution does not steer the flow toward the real distribution. Figure \ref{fig:reflow_drift_all}a provides empirical evidence of the accumulating drift in a toy \texttt{two moons} dataset \citep{scikit-learn}. The successive reflow steps cause the fake (yellow) data to diverge further from the real (blue) target distribution. Furthermore, Figure \ref{fig:reflow_drift_all}b illustrates the progressive increase in KL divergence between the fake distribution and the target distribution, providing clear evidence of this phenomenon.\footnote{The KL divergence is measured between Gaussian mixture approximations of the fake and real samples.} We provide a solution to mitigate this issue in Section \ref{asec:conic} and \ref{asec:bconic}.





\subsection{Real pair}
\label{sec:kernelpair}

The previous subsection has unveiled the pitfall of supervision using fake pairs: samples from the domain distribution, e.g., Gaussian, and their codomain following a rectified flow. Instead of the fake pairs, we propose to use the real samples and their inverse following a reverse rectified flow, defined by :
\vspace{-0.5pt}
\begin{equation}
\text{Real pair} (Z_{0,R}, X_1) := (v^{-1}(X_1), X_1)
\end{equation}
$\text{with } X_1 \sim \pi_{1}$ where $v$ is the 1-rectified flow and we abuse the term \textit{real pair} although the $Z_{0,R}$ is not real.
As in the original rectified flow, where it was optionally provided, it is safe and easy to use reverse rectified flow without stochasticity because it inherently produces a deterministic solution, and using real images does not contradict the original purpose because the noise ${v}^{-1}({X}_1)$ is generated using ${v}^{-1}$. To avoid confusion, from now on, we will refer to $(X_0, v(X_0))$ as a \textit{fake pair} (generated pair), where $v(X_0)$ is a fake (generated) image, and $(v^{-1}(X_1), X_1)$ as a \textit{real pair}, where $X_1$ is a real image.

\subsection{Conic reflow}
\label{asec:conic}

Building upon the basic pairing of real samples with their reverse noises, we introduce conic reflow, which expands their influence on the domain distribution
to their neighboring areas, as shown in Figure~\ref{fig:conicreflow}b. When we train the model, we use spherical linear interpolation (Slerp) between the reverse noise $Z_{0,R}$ and a randomly sampled noise $\epsilon\sim  \mathcal{N}(0, I)$ with the interpolation ratio $\zeta$:

\begin{equation}
    \text{Slerp}(Z_{0,R}, \epsilon, \zeta)= \frac{\sin((1 - \zeta) \phi)}{\sin(\phi)} Z_{0,R} + \frac{\sin(\zeta \phi)}{\sin(\phi)} \epsilon,
\end{equation}

where $\phi = \arccos(Z_{0,R} \cdot \epsilon)$ denotes the angle between $Z_{0,R}$ and $\epsilon$. Then we define a conic inverse from a real sample $X_1$:
\vspace{-0.3em}
\begin{equation}
    \text{Conic}(X_1, \epsilon, \zeta,t)=tX_{1} + (1-t)\text{slerp}(Z_{0,R}, \epsilon, \zeta),
\end{equation}
where $Z_{0,R}=v_\theta^{-1}(X_1)$, and $t\in [0,1]$. During training, we sample $\epsilon$ and $\zeta$ multiple times to let the target flow stochastically cover the nearby domain. As the collection of the paths over multiple iterations looks like a cone, we name our method as \textit{conic reflow}. The schedule of interpolation weight $\zeta$ is deferred to Section~\ref{sec:detail}. Our training objective with conic reflow is:
\vspace{-0.3em}
\begin{figure}[!t]
\vspace*{0pt}  
\begin{center}
    \includegraphics[width=0.8\textwidth]{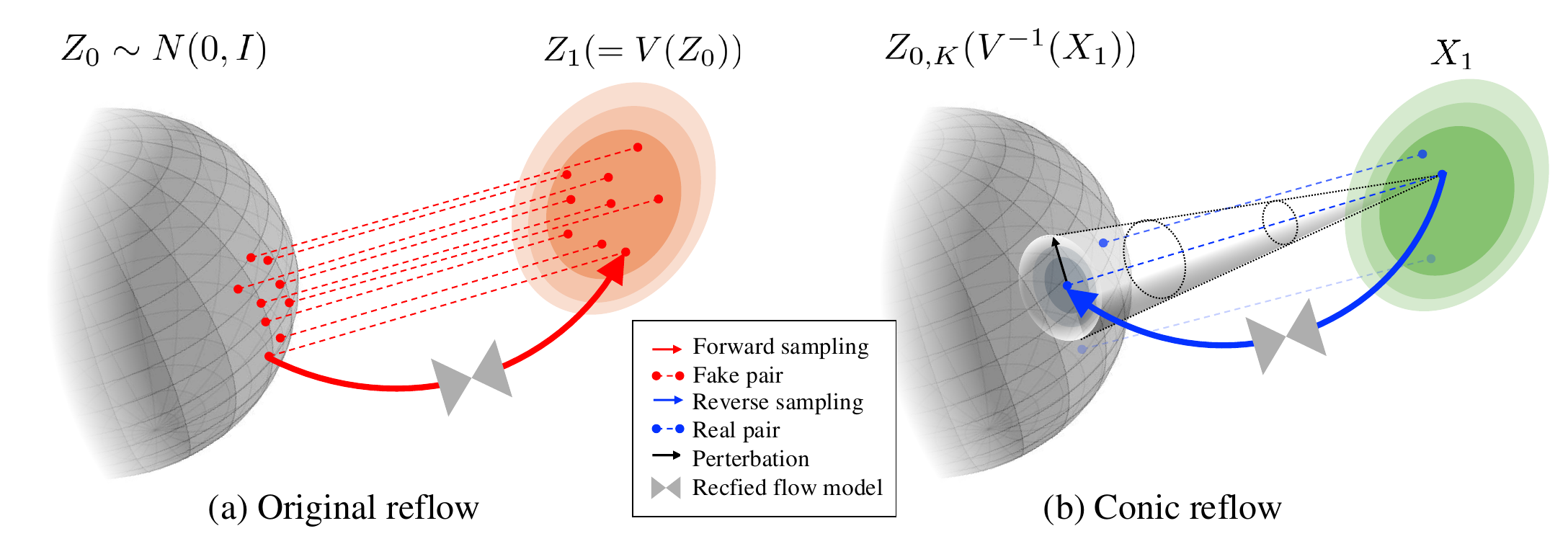}
    \caption{\textbf{Illustration of original fake pairs and our real pairs.} (a) The original rectified flow supervises 2-rectified flow with fake pairs $(Z_0, v_\theta(Z_0))$. (b) Our conic reflow supervises 2-rectified flow with real pairs $(v_\theta^{-1}(X_1), X_1)$ and their conic neighbors.}
    \vspace{-1em}
    \label{fig:conicreflow}
\end{center}
\end{figure}

\begin{equation}
\begin{aligned}
\hat{\theta} = \arg \min_{\theta} \int_0^1 \mathbb{E} \Big[ w_t \Big\| X_1 
- \text{slerp}(Z_{0,R}, \epsilon, \zeta)
- v_{\theta}\big(\text{Conic}(X_1, \epsilon, \zeta, t)\big) \Big\|^{2} \Big] dt
\end{aligned}
\end{equation}

Where $t \sim \text{exp}([0, 1])$, $\zeta \sim \text{slerp schedule}([0, 1])$, $\epsilon \sim \mathcal{N}(0, I)$, and $w_t$ is weighting function (default=1). Slerp is commonly used for interpolation in the noise space of generative models, as it preserves vector magnitudes on the Gaussian hypersphere and enables smooth semantic transitions \citep{wang2023interpolating, jang2024spherical}. In our case, Slerp serves as a geometry-aware regularizer that improves the alignment between numerically inverted real samples and the underlying data manifold. Rather than simply reusing generated samples, we supervise the model using real samples and their perturbed inversions to reduce reconstruction loss discrepancies observed during reflow. This approach not only mitigates distributional drift but is also aligned with adversarial robustness strategies studied in inverse problems and reconstruction-based neural networks \citep{antun2020instabilities, adler2021task}. From this perspective conic reflow provides localized supervision that encourages stability and continuity around real data points in the noise space.


\subsection{Balanced conic rectified flow}
\label{asec:bconic}
We design a new reflow procedure which consists of our conic reflow and the original reflow. 
For each training iteration, we design different training schemes for real pairs and fake pairs. We alternate between conic reflow steps with real pairs $(Z_{0,R}, X_{1})$ and original reflow steps with fake pairs.
It encourages the trajectories to head toward the exact real distribution while fake pairs ensure the entire domain distribution to receive supervision. Fake Pairs: For fake pairs, we proceed with the reflow process exactly as it was done in the original rectified flow model. Its training objective $\hat{\theta}$ as follows:
\begin{equation}
\arg \min_{\theta} \mathbb{E} \left[ \| Z_{1,F} - Z_{0,F} - v_{\theta}(t Z_{1,F} + (1 - t) Z_{0,F}) \|^2 \right],
\end{equation}
\vspace{-0.1em}
$\text{with } t \sim \text{exp}([0, 1]).$ The entire training objective of our method is as follows for the given fake pair($Z_{0,F},Z_{1,F}$) and real pair ($Z_{0,R},X_{1}$):
\vspace{-0.5em}

\begin{equation}
\begin{gathered}
\min_v \int_0^1 \Big[ \Big\| \chi_{\text{fake}} \cdot \left(\dot{Z}_{t,F} - v_{\theta}(Z_{t,F})\right) + \chi_{\text{real}} \cdot \Big(X_1 - \text{slerp}(Z_{0,R}, \epsilon, \zeta) 
- v_{\theta}\big(\text{Conic}(X_1, \epsilon, \zeta, t)\big) \Big)
\Big\|^{2} \Big] dt
\end{gathered}
\end{equation}

where $\chi_{\text{fake}}$ and $\chi_{\text{real}}$ are indicator functions for given index subsets $U_{\text{real}}$ and $U_{\text{fake}}$ such that $U_{\text{real}} \cup U_{\text{fake}} = \mathbb{N}$. Then, for $i \in \mathbb{N}$:
\vspace{-0.5em}
\begin{equation}
\chi_{\text{fake}} = 
\begin{cases} 
1 & \text{if } i \in U_{\text{fake}}\\
0 & \text{else} 
\end{cases}, \quad\quad
\chi_{\text{real}} = 
\begin{cases} 
1 & \text{if } i \in U_{\text{real}}\\
0 & \text{else}.
\end{cases}
\end{equation}
The notation $\zeta$, Conic$(\cdot)$, $\epsilon$, and $w_t$ follows the definitions in Section~\ref{asec:conic}.

\subsection{Detailed training schemes}
\label{sec:detail}

In this section, we provide a more detailed explanation of our proposed training schemes, including visualizations of the Slerp scheduling and maximum Slerp noise magnitude. The maximum magnitude of the Slerp noise, $\zeta_{\text{max}} \in (0,0.5]$ is selected in accordance with our intuition that it is the point where the discrepancy between the perturbed reconstruction errors of real and fake samples is maximized. After a warm-up phase of training steps, we compute $\zeta_{\text{max}}$ using 10{,}000 images each from the real and fake. The value is selected based on the following hyperparameter $\zeta^{\text{max}}$:

\begin{equation}
\label{mrd}
\zeta^{\text{max}} := \max_{\zeta \in (0, 0.5]} \;
\mathop{\mathbb{E}}_{\substack{x \sim X_1 \\ z \sim Z_{1,F}}}
\left[
\left\| v_{\theta}\left( \text{Slerp}(z_{0,R}, \epsilon, \zeta) \right) - x \right\|_2
-
\left\| v_{\theta}\left( \text{Slerp}(z_{0,F}, \epsilon, \zeta) \right) - z \right\|_2
\right]
\end{equation}

\begin{figure*}[t!]
    \centering

    \begin{minipage}[t]{0.22\textwidth}
        \centering
        \includegraphics[width=\linewidth]{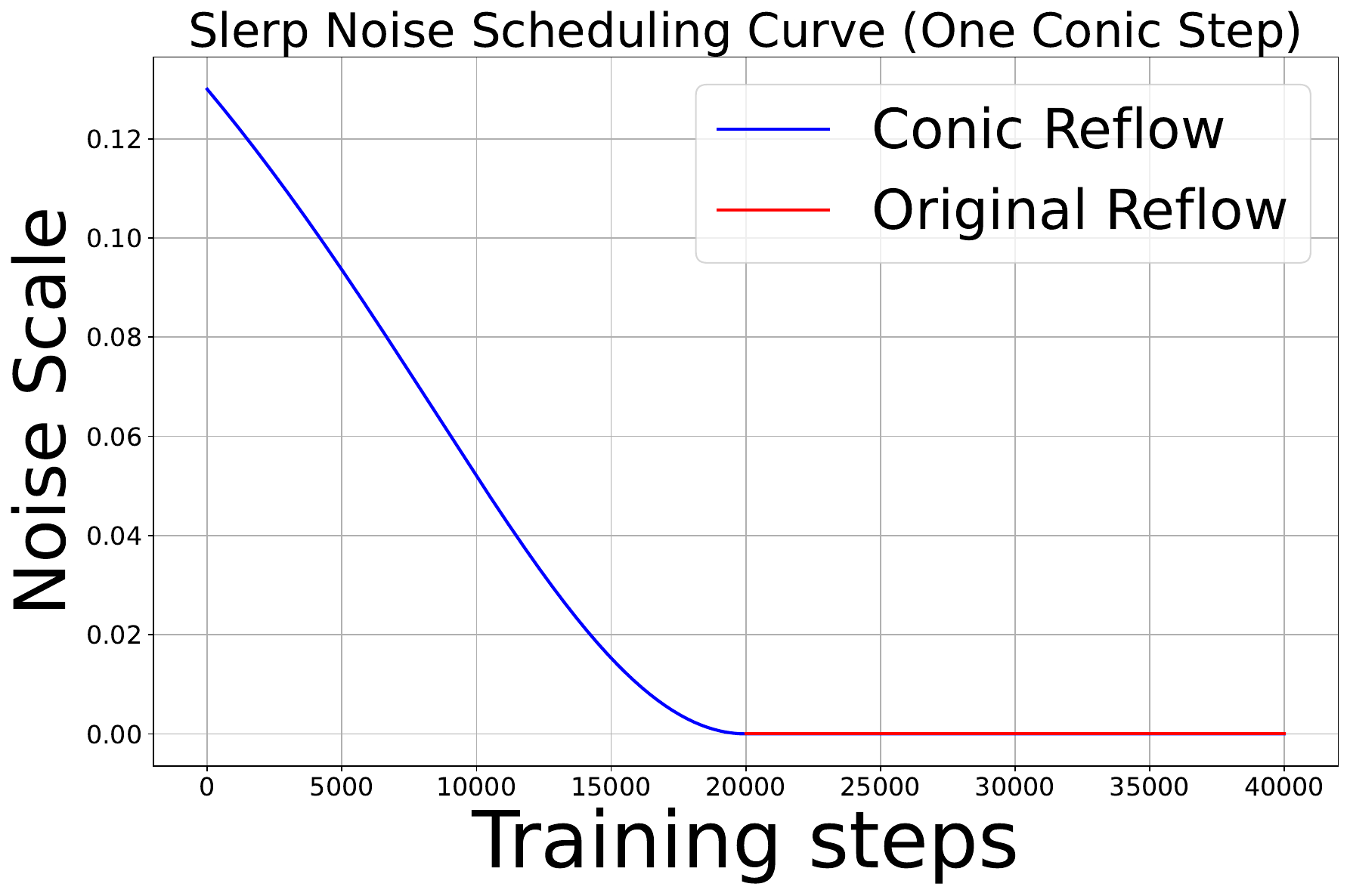}
    \end{minipage}\hfill
    \begin{minipage}[t]{0.31\textwidth}
        \centering
        \includegraphics[width=\linewidth]{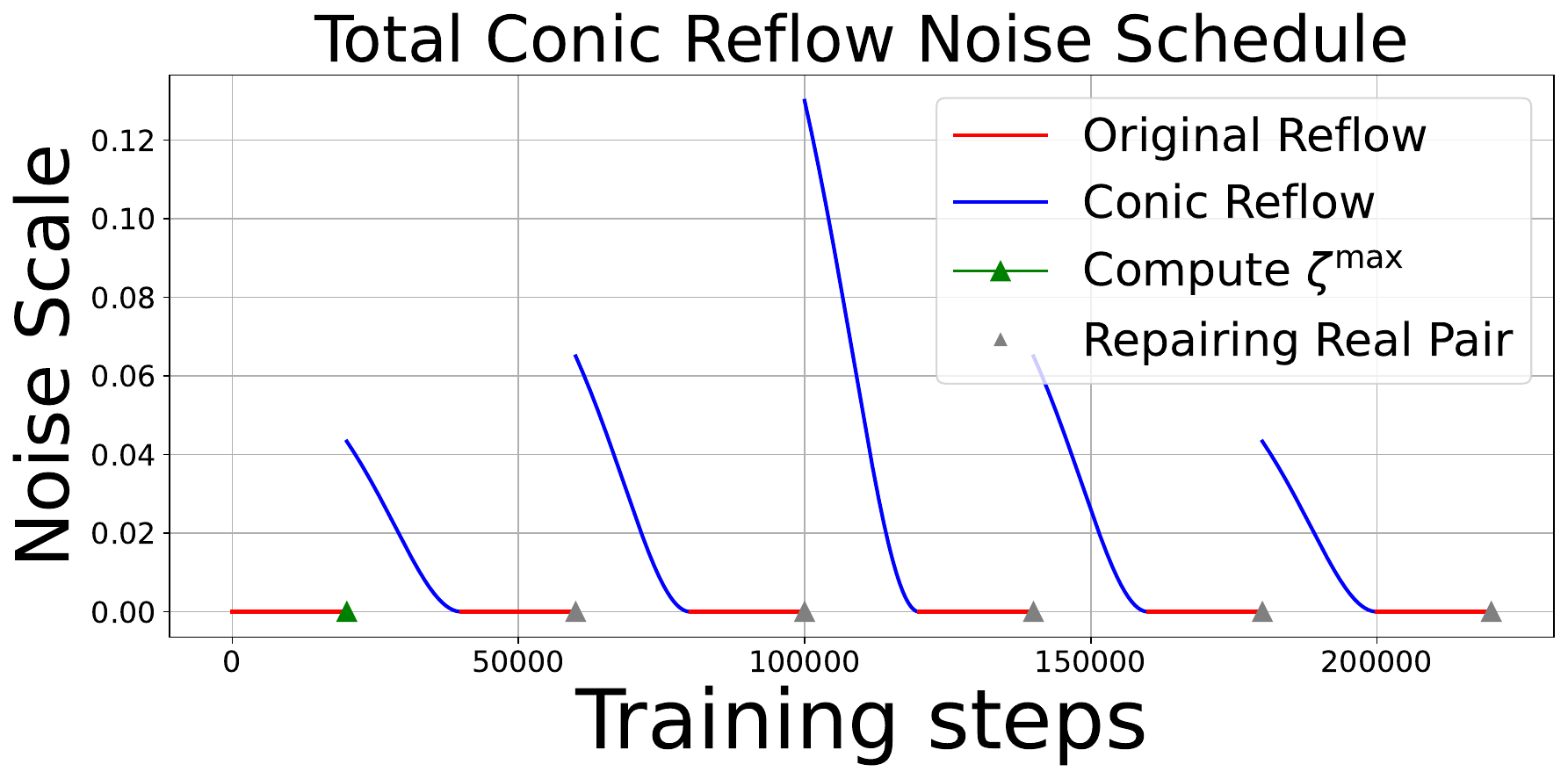}
    \end{minipage}\hfill
    \begin{minipage}[t]{0.22\textwidth}
        \centering
        \includegraphics[width=\linewidth]{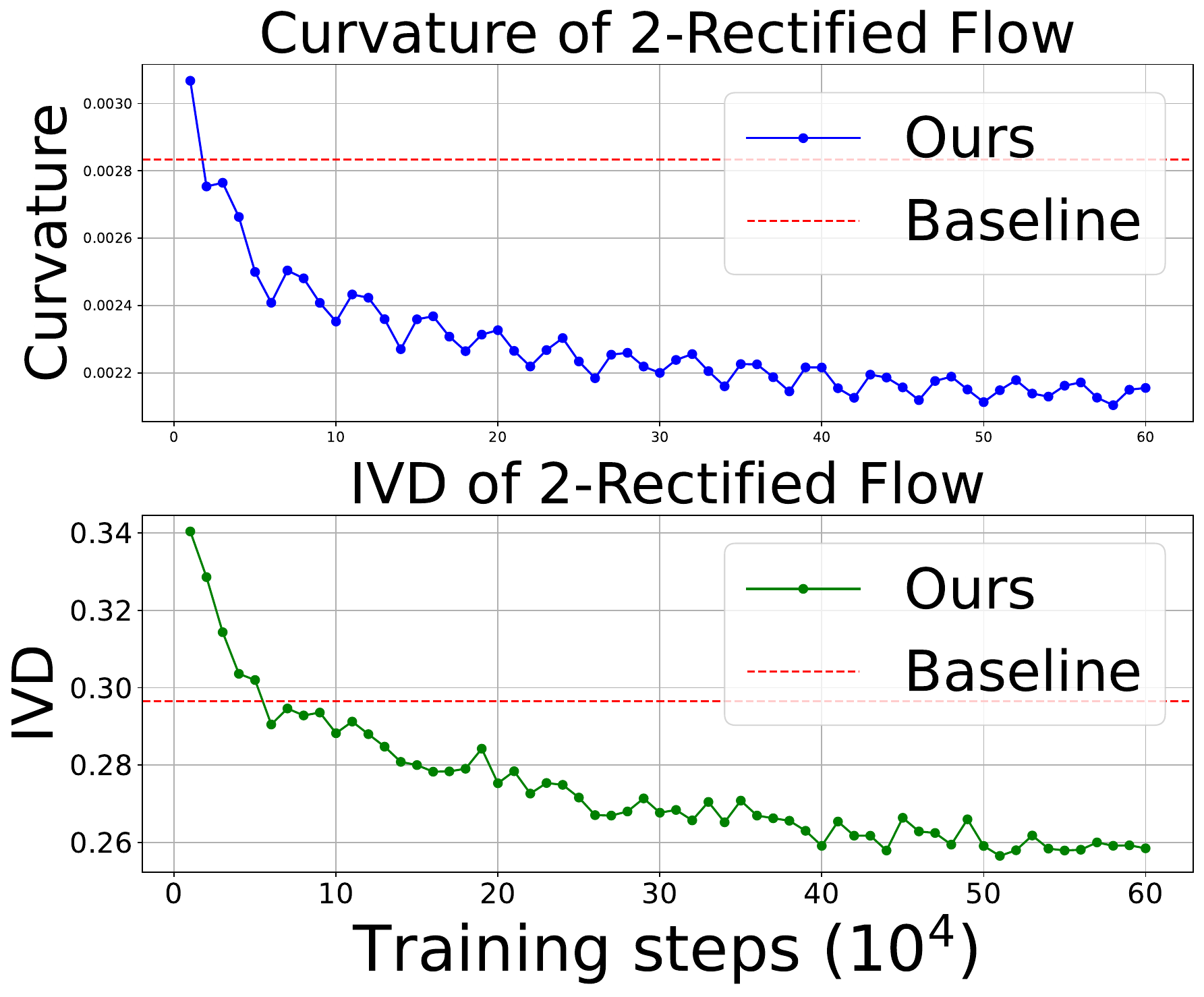}
    \end{minipage}\hfill
    \begin{minipage}[t]{0.22\textwidth}
        \centering
        \includegraphics[width=\linewidth]{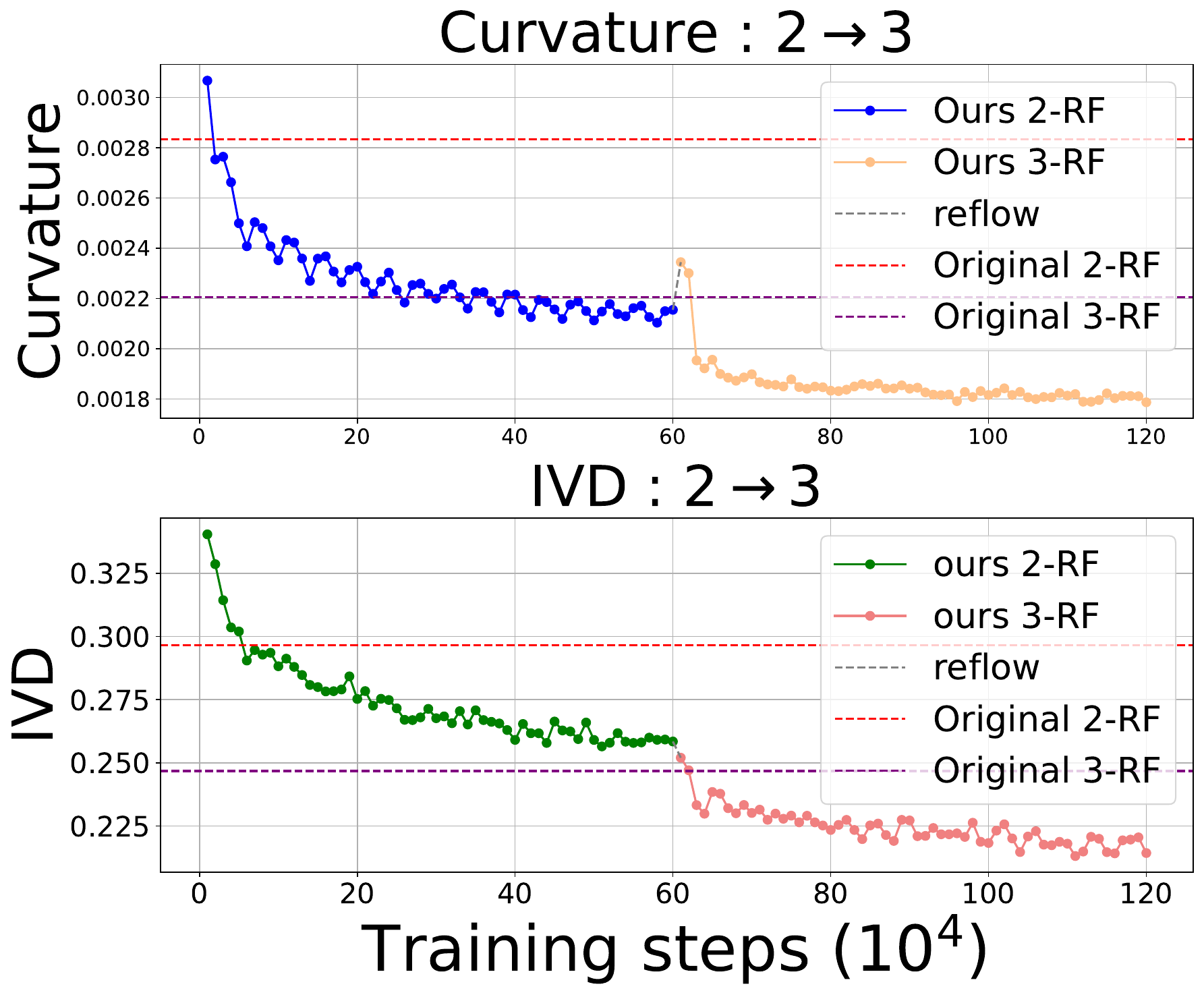}
    \end{minipage}
    \caption{
\textbf{(a)} Slerp noise schedule for single conic reflow and \textbf{(b)} Total slerp schedule. \textbf{(c)} Curvature and IVD during training, and \textbf{(d)} From 2- to 3-rectified flow.
    }
    \label{fig:combined_plots}
    \vspace{-0.5em}
\end{figure*}

\vspace{-0.5em}
where $\epsilon \sim \mathcal{N}(0, I)$, with $z_{0,R} = v_\theta^{-1}(x)$ and $z_{0,F} = v_\theta^{-1}(z_{1,F})$ denoting the inverse noise estimates for the real and fake samples, respectively. Figure~\ref{fig:combined_plots}(a) shows a single conic reflow noise schedule in training steps (For CIFAR-10 and ImageNet, we set $\zeta^{\text{max}}$ to 0.13 and 0.23, respectively). Each conic is trained to progressively reduce the noise scale over time.
Specifically, the Slerp noise schedule $\zeta(t')$ is defined as $\zeta(t') := \zeta^{\text{max}} \cdot \frac{2{t'}^2}{1 + {t'}^2}, \quad t' \in [0, 1],$ where $t' = 1$ corresponds to the start of training and $t' = 0$ to the end. This design follows the intuition from traditional diffusion models~\citep{ho2020denoising}, where noise is progressively reduced to guide samples toward realism. To strengthen the real image trajectory during training, we periodically update the real sample pairs used in the reflow process. Figure~\ref{fig:combined_plots}(b) illustrates an example of conic reflow with updated real pairs, assuming 220K total training steps for visualization.
When the number of updates is $2K$, we schedule the maximum noise magnitude $\zeta^{\text{max}}$ by scaling it according to the pattern $[K, K{-}1, \ldots, 1, 2, \ldots, K]$, assigning $\zeta^{\text{max}} / K$ to the smallest value and $\zeta^{\text{max}}$ to the midpoint. The noise increases linearly in the first half and decreases symmetrically in the second half. The full pseudocode for our training method is provided in Appendix~\ref{asec:algorithm}.

\section{Experiments}
\label{sec:experiments}

We conducted experiments to evaluate the effectiveness of our method. Our findings demonstrate: Superiority over original reflow in terms of (1) Quality of the results, (2) Straightness of the flow, (3) Mitigation of distribution shift, as well as (4) Ablation study, (5) Generalization to other datasets.


\paragraph{Experimental setup}

Most of our experiments are conducted on CIFAR-10 \citep{krizhevsky2009learning}. The IVD, curvature, reconstruction, and perturbed ($0.05\varepsilon$, 1-step) reconstruction error values reported were computed using 10,000 random samples, with the expectation taken over these samples. We employ Scipy's RK45\citep{virtanen2020scipy}, a 5(4) Runge-Kutta method with adaptive step size and step count determined by specified tolerances, following the same parameters \citep{song2020score}. Further details on the training configurations are provided in Appendix~\ref{asec:config}.

\subsection{Image quality}

\begin{figure}[t!]
    \centering
    \begin{minipage}[t]{0.33\linewidth}
        \vspace*{0pt}  
        \centering
        \scriptsize  
        \setlength{\tabcolsep}{2pt}
        \resizebox{\linewidth}{!}{%
        \begin{tabular}{lccc}
            \toprule
            \textbf{Method} & \textbf{NFE (↓)} & \textbf{IS (↑)} & \textbf{FID (↓)} \\ \midrule
    
            \multicolumn{4}{l}{\textbf{One-Step Generation (Euler solver, N=1)}} \\
            1-Rectified Flow & 1 & 1.13 (9.08) & 378 (6.18) \\ \midrule
    
            \multicolumn{4}{l}{\textit{2-Rectified Flow}} \\
            Original (\textit{+Distill}) & 1 & 8.08 (9.01) & 12.21 (4.85) \\
            \textbf{Ours} (\textit{+Distill}) & 1 & \textbf{8.79} (\textbf{9.11}) & \textbf{5.98} (\textbf{4.16}) \\ 
            $\text{Rf++}^\dagger$\citep{lee2024improving} & 1 & 8.87 & 4.43 \\ 
            $\text{Rf++}^\dagger$(\textbf{+ours}) & 1 & 8.87 & \textbf{4.22} \\\midrule
            \multicolumn{4}{l}{\textit{3-Rectified Flow}} \\
            Original (\textit{+Distill}) & 1 & 8.47 (8.79) & 8.15 (5.21) \\
            \textbf{Ours} (\textit{+Distill}) & 1 & \textbf{8.84} (\textbf{8.96}) & \textbf{5.48} (\textbf{4.68}) \\ \midrule
    
            \multicolumn{4}{l}{\textbf{Full Simulation (Runge–Kutta (RK45), Adaptive N)}} \\
            1-Rectified Flow & 127 & \textbf{9.60} & \textbf{2.58} \\ \midrule
    
            \multicolumn{4}{l}{\textit{2-Rectified Flow}} \\
            Original & 110 & 9.24 & 3.36 \\
            \textbf{Ours} & 104 & \textbf{9.30} & \textbf{3.24} \\ \midrule
    
            \multicolumn{4}{l}{\textit{3-Rectified Flow}} \\
            Original & 104 & 9.01 & 3.96 \\
            \textbf{Ours} & 98 & \textbf{9.14} & \textbf{3.70} \\
    
            \bottomrule
            \end{tabular}
        }
        \vspace{-0.5em}
       \captionof{table}{One-step and full-simulation comparison of 2,3 Rectified Flows on CIFAR-10.}
        \label{tab:quality_table}
    \end{minipage}
    \hspace{3em}
    \begin{minipage}[t]{0.55\linewidth}
        \vspace*{0pt}  
        \centering
        \includegraphics[width=\linewidth]{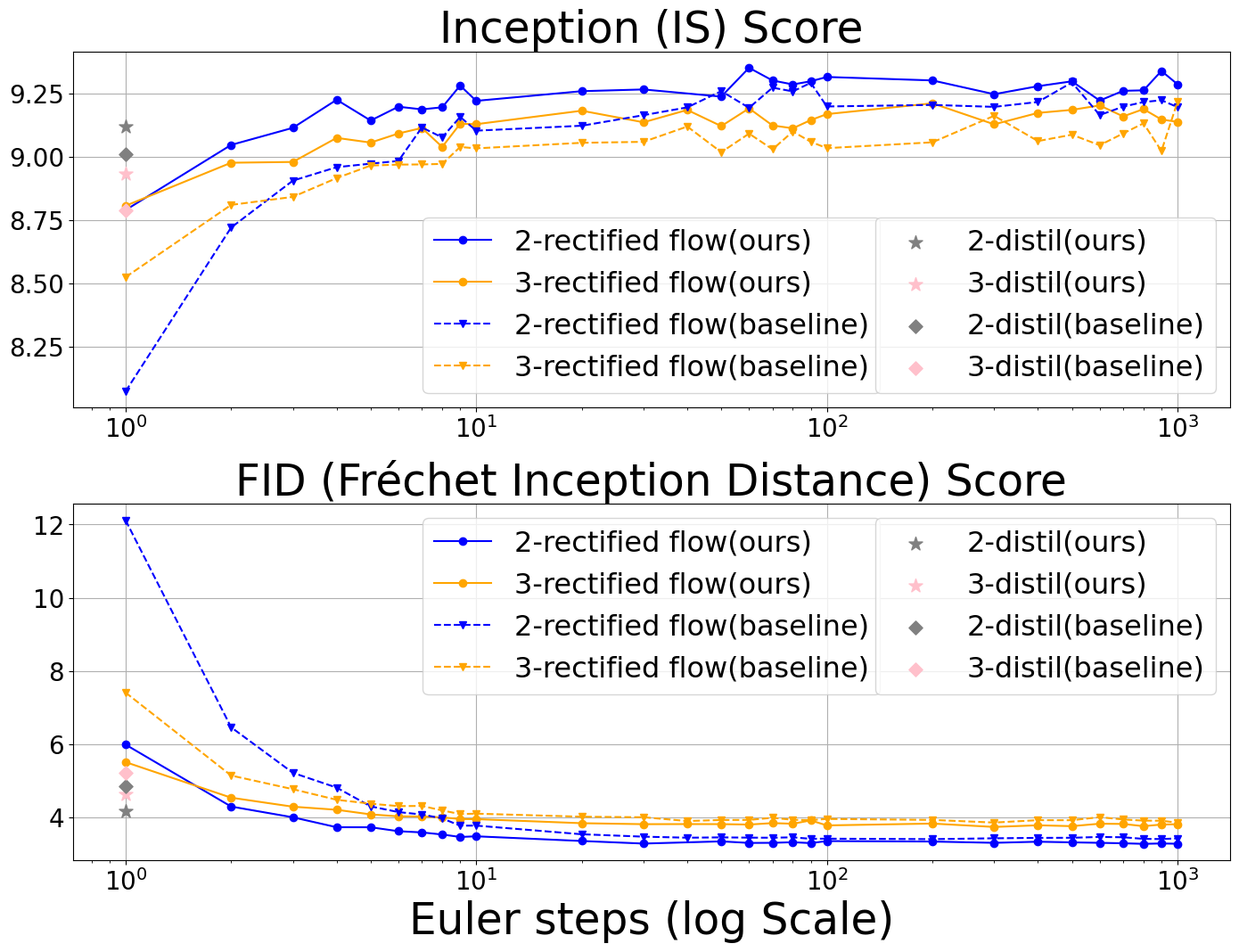}
       \parbox{0.85\linewidth}{\captionof{figure}{ CIFAR-10 generation quality across Euler steps.}
       \label{fig:quality_compare}}
    \end{minipage}
    \vspace{-2em}
\end{figure}

Our method achieves better FID and IS scores across all sampling steps, i.e., 1-step, few-step, and full-step generations, as shown in Table~\ref{tab:quality_table} and Figure~\ref{fig:quality_compare}. Notably, we use only 300K fake pairs, significantly fewer than the 4M fake pairs used in the original rectified flow, demonstrating the efficiency of our approach. Furthermore, our method outperforms the original in generation quality with RK sampling, even with a lower NFE. We also apply our method to Rectified++~\citep{lee2024improving} and compare generation quality. While RF++$^\dagger$ uses 800K synthetic pairs, our variant uses 600K fake pairs and 50K real images, achieving better FID. This indicates that our method generalizes well to reflow-based generative models. Detailed settings for RF++$^\dagger$ are provided in Appendix~\ref{asec:config}.

Additionally, our method produces images with superior quality even when applying the same distillation in the original rectified flow \citep{liu2022flow} as compared to solid (ours) and dashed (original) lines, and star (ours) and rectangle (original) in Figure~\ref{fig:quality_compare}.
These results suggest that our method produces a more favorable initial velocity field than the original rectified flow. A detailed comparison of precision and recall scores is provided in Appendix~\ref{asec:recall}. Furthermore, we show in Appendix~\ref{asec:finetune} that fine-tuning the original rectified flow using only real images and their reverse noise pairs effectively reduces reconstruction discrepancy and quickly improves generation quality. Additional performance comparisons on more complex datasets and higher-resolution images are presented in Sections~\ref{sec:imagenet} and \ref{sec:high}. To further contextualize the performance of reflow-based methods, we also report the unconditional generation quality of pretrained diffusion models on CIFAR-10 in Appendix~\ref{asec:diffusion}.

\vspace{-0.5em}

\subsection{Straightness}
We evaluate trajectory straightness using curvature, following prior works \citep{liu2022flow,lee2023minimizing}. Straighter trajectories reduce discretization error under few step solvers, improving sample quality \citep{stetter1973analysis,chen2023improved}. For any continuously differentiable process $\mathbf{Z} = \{{Z_t}\}$ , the curvature is measured by :
\vspace{-0.5em}

\begin{equation} 
S(\mathbf{Z}) = \int_0^1 \mathbb{E} \left[ \left\| \left( Z_1 - Z_0 \right) - \dot{Z_t} \right\|^2 \right] dt
\end{equation}
 Additionally, it is known that $\mathbf{S}(\mathbf{Z}) = 0$ indicates exact straightness.
\vspace{-0.5em}
\paragraph{Relationship between curvature and initial velocity delta (IVD)}
While curvature captures trajectory straightness, it does not fully explain 1-step sampling quality, which depends solely on the initial velocity. We propose Initial Velocity Delta (IVD) to directly evaluate the accuracy of the initial velocity and its impact on 1-step generation. The calculation method for IVD is provided in the equation below:
\vspace{-0.5em}
\begin{equation}
IVD(\mathbf{Z}, t_0) = \mathbb{E} \left[ \left\| (Z_1 - Z_0) -\dot{Z}_{t_{0}}\right\|^2 \right]
\end{equation}
\vspace{-0.5em}
\paragraph{Curvature and initial velocity delta.}
Our approach demonstrates improved trajectory straightening and better preservation of the initial velocity direction. As shown in Figure \ref{fig:combined_plots}(c), our method consistently achieves lower curvature and IVD values than the original, indicating a more stable trajectory even with fewer fake pairs. Furthermore, Figure \ref{fig:combined_plots}(d) highlights that applying an additional reflow step from 2-rectified to 3-rectified flow further reduces both curvature and IVD, reinforcing the effectiveness of our training method.

\subsection{Reconstruction with perturbation to address drift from the real distribution}
\vspace{-0.5em}
\begin{figure*}[b]
    \vspace{-1em}
    \centering
    \begin{subfigure}{0.45\textwidth}
    \vspace*{0pt}  
        \centering
        \includegraphics[width=\textwidth]{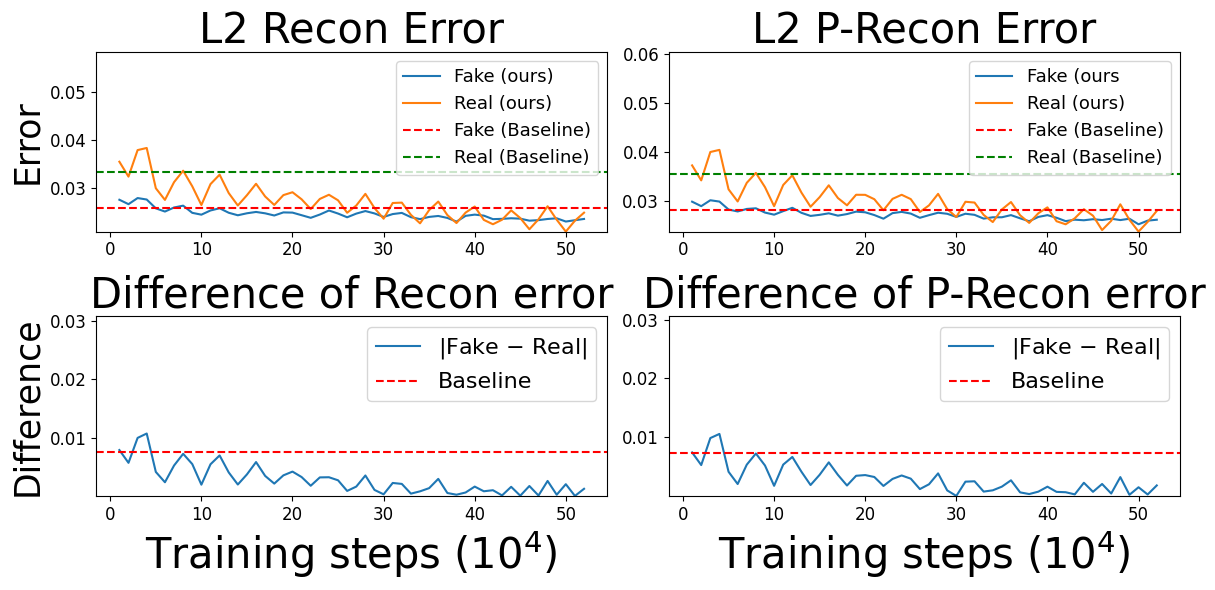}
    \end{subfigure}
    \begin{subfigure}{0.45\textwidth}
    \vspace*{0pt}  
        \centering
        \includegraphics[width=\textwidth]{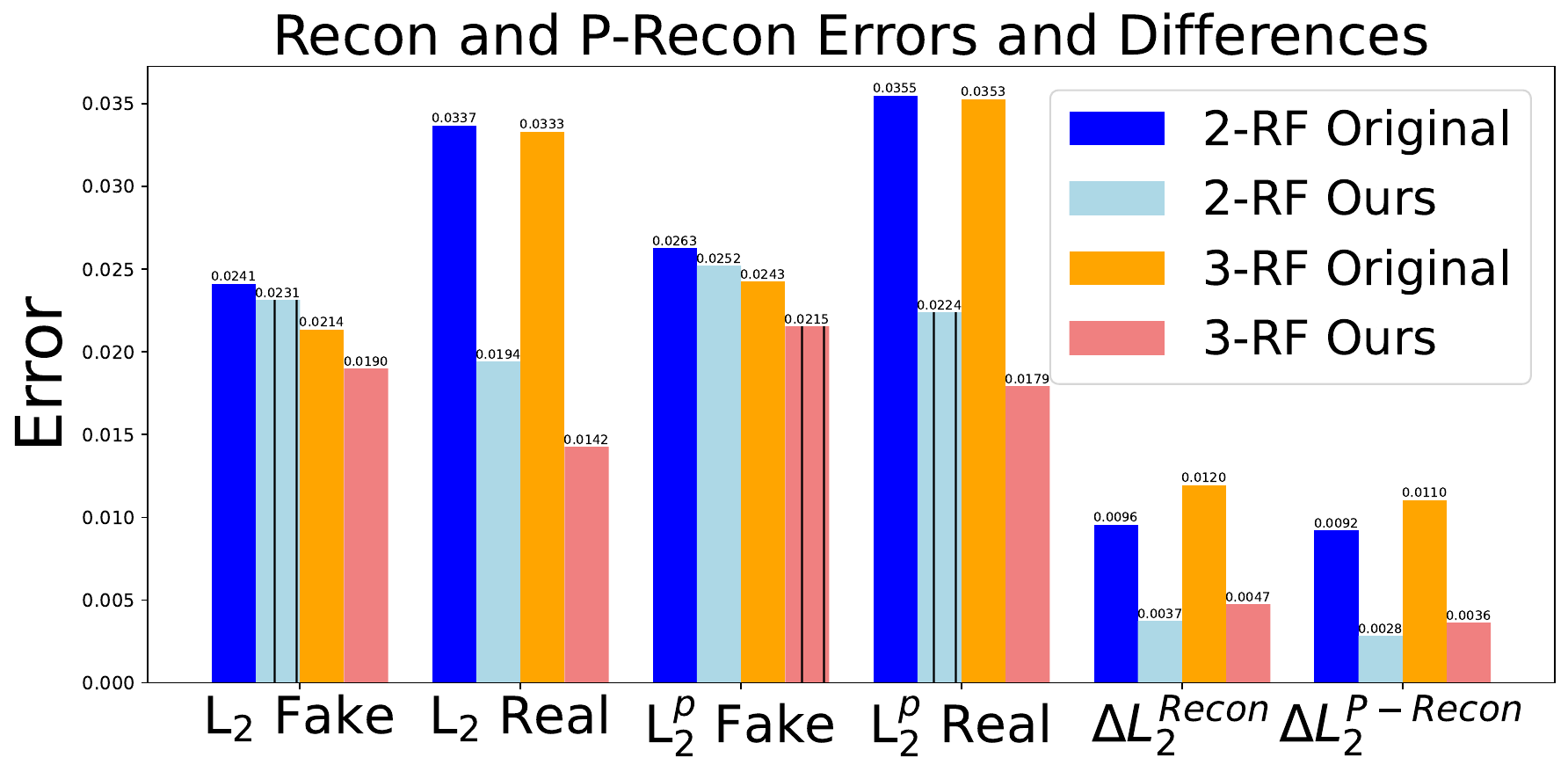}
    \end{subfigure}
    \caption{Reconstruction and perturbed reconstruction error across training iterations.}
    \label{fig:recon_error_ablation}
    \vspace{-0.5em}
\end{figure*}

We empirically show that incorporating real data pairs via Slerp-based supervision in the reflow step significantly improves the preservation of real image trajectories. As illustrated in Figure~\ref{fig:recon_error_ablation}(a), the original rectified flow exhibits a clear reconstruction error gap between real and fake images, whereas our method progressively narrows this gap during training. Figure~\ref{fig:recon_error_ablation}(b) highlights two key aspects. First, in terms of reconstruction error, our method better preserves real trajectories and mitigates overfitting to fake samples. Second, the lower perturbed reconstruction error suggests that the model more accurately aligns the velocity field around real images, leading to improved robustness against perturbations.




\subsection{Ablation study}

We report results from an ablation study on various settings of our proposed framework, comparing four configurations: (1) without Slerp noise; (2) reflow using just a single real pair; (3) our full method; and (4) the original 2-rectified flow. The comparison focuses on 1-step generation quality, curvature, IVD(Initial Velocity Delta), reconstruction error, and perturbed reconstruction error.

\subsubsection{Benefits of incorporating real data via slerp-based perturbation training}
\vspace{-0.2em}
\begin{table*}[htbp] 
\centering
\tiny 
\renewcommand{\arraystretch}{1} 
\setlength{\tabcolsep}{1pt} 
\resizebox{\textwidth}{!}{%
\begin{tabular}{lcccccccc}
\toprule
\textbf{Model} & \textbf{FID} & \textbf{IS} & \textbf{Curvature} & \textbf{IVD} & \textbf{Recon Real} & \textbf{Recon Fake} & \textbf{Perturbed Real} & \textbf{Perturbed Fake} \\ 
\midrule
Original           & 12.21 & 8.08 & 0.002837 & 0.295078 & 0.033668 & 0.024106 & 0.035481 & 0.026270 \\
\textbf{Ours* (Slerp + conic)} & \textbf{5.98} & \textbf{8.79} & \textbf{0.002295} & 0.253334 & \textbf{0.019404} & 0.023139 & \textbf{0.022382} & 0.025206 \\
Fixed Real Pair    & 6.69 & \underline{8.59} & \underline{0.002313} & \underline{0.242444} & \underline{0.020227} & \underline{0.020607} & \underline{0.022890} & \underline{0.022914} \\
No Slerp           & \underline{6.60} & 8.57 & 0.002322 & \textbf{0.240884} & 0.023380 & \textbf{0.020154} & 0.026063 & \textbf{0.022496} \\
\bottomrule
\end{tabular}%
}
\caption{An ablation table comparing various 2-rectified models across multiple metrics: FID, IS (1-step), Curvature, IVD, and errors (Recon Real/Fake, Perturbed Real/Fake).}
\label{tab:comparison_metrics}
\vspace{-0.5em}
\end{table*}



\paragraph{Benefits of adding noise via Slerp}

Adding noise via Slerp avoids trajectory crossover, preserving a straighter path relative to real images. This leads to improved trajectory quality and enhances 1-step sampling efficiency. Moreover, using Slerp results in lower reconstruction error for real images compared to not using it. This demonstrates that Slerp-based noise perturbation helps stably maintain the trajectory and local neighborhood structure of reverse noise corresponding to real images.
In particular, it plays a key role in reducing the reconstruction discrepancy we observe between real and fake samples during reflow, thereby effectively mitigating the distributional drift that occurs in later iterations.



\paragraph{Effect of real pair injection and refreshing}
Incorporating real pairs into training improves trajectory fidelity and 1-step generation quality by anchoring learning closer to the target distribution. Notably, fixing the real pair only once leads to inferior performance, while periodically refreshing reverse noise achieves better FID, IS, and lower reconstruction errors. This highlights the importance of continuously updating real-sample guidance throughout training to maintain alignment with real data trajectories.

Additional experiments, including $k=4$ generation quality, precision/recall, low fake pair settings, and real-only supervision, are provided in Appendices~\ref{asec:more_reflow} and \ref{asec:extreme60k}.

\subsubsection{Slerp noise patterns and lerp}
\label{asec:slerp_pattern}
In this section, we show that our Slerp-based reflow method consistently improves 1-step generation quality, even when the Slerp noise pattern varies. To support this claim, we train 2-rectified flow models using different Slerp schedules, including patterns where the noise gradually increases, gradually decreases, or first increases and then decreases. We also include a baseline where Slerp is replaced with Lerp for comparison. All experiments are conducted with a batch size of 256, and training is performed for 300K iterations. Each setting uses 300K fake pairs and 60K real pairs for training on Cifar 10. Other configurations remain identical to those described in Section~\ref{sec:experiments} and Appendix~\ref{asec:config}.

\begin{table}[!ht]
\begin{minipage}{\textwidth}  
\small
\centering
\begin{tabular}{cc>{\centering\arraybackslash}p{6cm}cc}
\toprule
\textbf{Category} & \textbf{Method} & \textbf{Schedule Type} & \textbf{IS (↑)} & \textbf{FID (↓)} \\ 
\midrule
\multirow{3}{*}{\textbf{Slerp}} 
& Ours & $\zeta^{\text{max}}_{k}$: $\zeta^{\text{max}}/K \rightarrow \zeta^{\text{max}} \rightarrow \zeta^{\text{max}}/K$ & \textbf{8.72} & \textbf{6.63} \\
& Strictly Increasing & $\zeta^{\text{max}}_{k}$: $\zeta^{\text{max}}/K$ $\to$ $\zeta^{\text{max}}$ & \underline{8.48} & \underline{6.64} \\
& Strictly Decreasing & $\zeta^{\text{max}}_{k}$: $\zeta^{\text{max}}$ $\to$ $\zeta^{\text{max}}/K$ & 8.45 & 6.70 \\ 
\midrule
\textbf{Lerp} & Linear Interpolation & $(1{-}\zeta) \cdot v^{-1}(X_1) + \zeta \cdot \epsilon$, $\zeta \in [0, \zeta^{\text{max}}]$ & 8.46 & 7.50 \\ 
\bottomrule
\end{tabular}
\caption{Each method uses Slerp or Lerp-based noise trajectories. The best and second-best values are bolded and underlined, respectively.}
\label{tab:slerp_lerf_comparison}
\end{minipage}
\vspace{-0.5em}
\end{table}

As shown in Table~\ref{tab:slerp_lerf_comparison}, we observe three key findings. First, Slerp-based schedules consistently outperform Lerp in both FID and IS, suggesting that Slerp more effectively preserves the reverse noise trajectories of real images. Second, the strictly increasing noise schedule yields better performance than the decreasing one, indicating that progressive noise injection improves training stability. Third, our method achieves the highest IS while maintaining comparable FID to the best baseline, demonstrating improved sample diversity without compromising generation quality.

\subsection{\texorpdfstring{More complex dataset (Unconditional generation on ImageNet 64$\times$64)}{More complex dataset (Unconditional generation on ImageNet 64x64)}}
\label{sec:imagenet}

On ImageNet 64$\times$64 \citep{deng2009imagenet}, our method consistently improves unconditional generation quality over the original model. Using the same setup as CIFAR-10, we train a 1-rectified flow for 700K steps with batch size 256. The original uses \textbf{1M} fake pairs, while our method uses \textbf{600K} fake and 60K real pairs. As shown in Table~\ref{tab:imagenet}, Although 60K real images may be insufficient to fully cover the target distribution of ImageNet compared to CIFAR-10, our method still yields substantial gains in FID and Recall. This suggests that even limited real data can provide strong guidance in mitigating distributional drift. Moreover, it improves reconstruction error, perturbed reconstruction error, recall, and precision (see Table~\ref{tab:imagenet-precision} and Appendix~\ref{asec:imagenet_append}), demonstrating its ability to reduce distributional drift even on large-scale datasets.



\begin{table*}[t]

\scriptsize
\begin{minipage}[t]{0.5\textwidth}
    \vspace*{0pt}
    \resizebox{0.95\linewidth}{!}{
    \begin{tabular}{lcccc}
        \toprule
        \textbf{Metric}      & \multicolumn{2}{c}{\textbf{FID}} & \multicolumn{2}{c}{\textbf{IS}} \\ 
        \cmidrule(lr){2-3} \cmidrule(lr){4-5}
        \textbf{ODE Solver}  & \textbf{RK-45} & \textbf{Euler} & \textbf{RK-45} & \textbf{Euler} \\ \midrule
        1-rectified flow      & 23.1             & 369.8            & 12             & 1.1             \\ \midrule
        2-rectified flow      & 31.2             & 39.7             & 10.8           & \textbf{10.4}           \\
        \textbf{Ours}         & \textbf{28.2}    & \textbf{37.8}    & \textbf{11.4}  & 10.3  \\
        \bottomrule
    \end{tabular}
    }
    \parbox{0.9\linewidth}{\centering \captionof{table}{FID and IS across rectified flows using RK-45 and Euler solver on ImageNet.}
    \label{tab:imagenet}}
\end{minipage}
\hfill
\begin{minipage}[t]{0.5\textwidth}
    \vspace*{0pt}
    \resizebox{0.95\linewidth}{!}{
    \begin{tabular}{lccc}
        \toprule
        \textbf{Model} & \textbf{Solver} & \textbf{Precision} & \textbf{Recall} \\ 
        \midrule
        Original & Euler 1-step & \textbf{0.4129} & 0.4604 \\
        Ours     & Euler 1-step & 0.3812 & \textbf{0.5325} \\\midrule
        Original & RK-45        & \textbf{0.4717} & 0.5264 \\
        Ours     & RK-45        & 0.4432 & \textbf{0.6032} \\
        \bottomrule
    \end{tabular}
    }
    \parbox{0.85\linewidth}{\centering \captionof{table}{Precision and recall on ImageNet.}
    \label{tab:imagenet-precision}}
\end{minipage}
\vspace{-1em}
\end{table*}


\subsection{High-resolution image generation}
\label{sec:high}
\begin{figure*}[t!]  

\begin{minipage}[t]{0.47\textwidth}
    \vspace*{0pt}

    \begin{minipage}[b]{0.29\linewidth}
        \includegraphics[width=\linewidth]{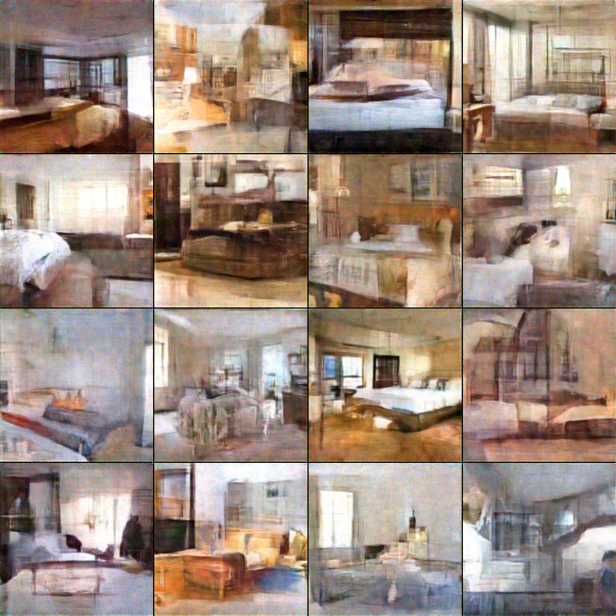}
    \end{minipage}%
    \hfill
    \begin{minipage}[b]{0.29\linewidth}
        \includegraphics[width=\linewidth]{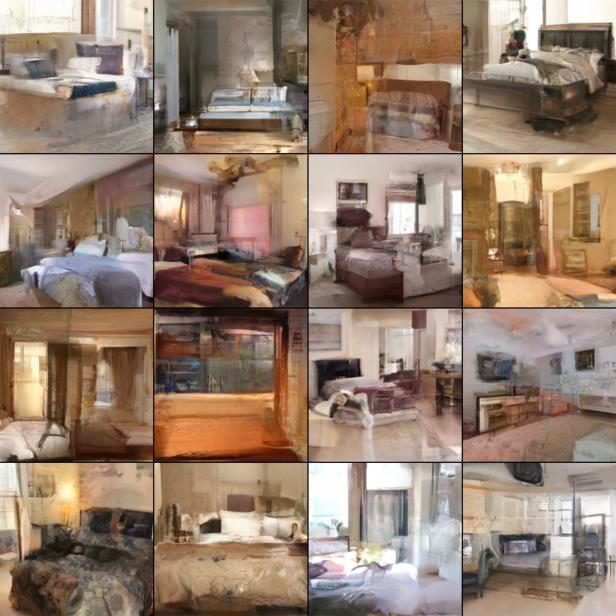}
    \end{minipage}%
    \hfill
    \begin{minipage}[b]{0.29\linewidth}
        \includegraphics[width=\linewidth]{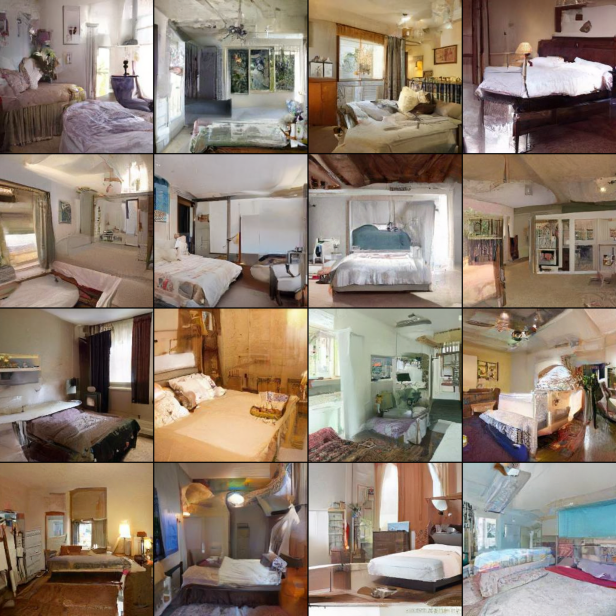}
    \end{minipage}

    \vspace{1mm}

    \begin{minipage}[b]{0.29\linewidth}
        \includegraphics[width=\linewidth]{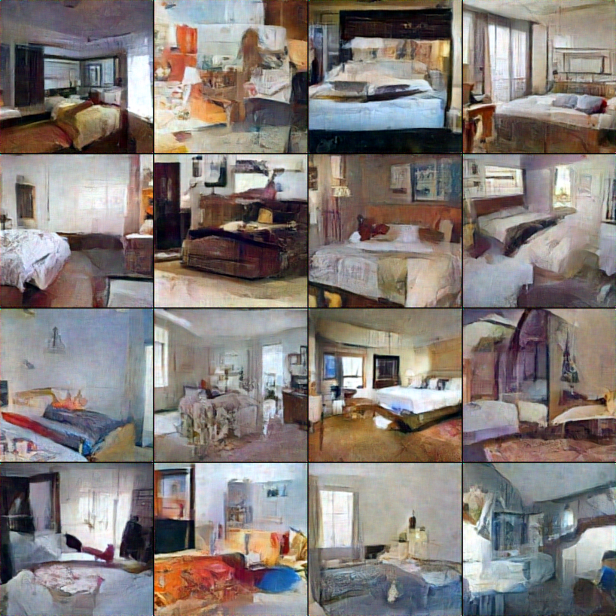}
        \vspace{-1em}
        \caption*{\scriptsize Euler 1-step}
    \end{minipage}%
    \hfill
    \begin{minipage}[b]{0.29\linewidth}
        \includegraphics[width=\linewidth]{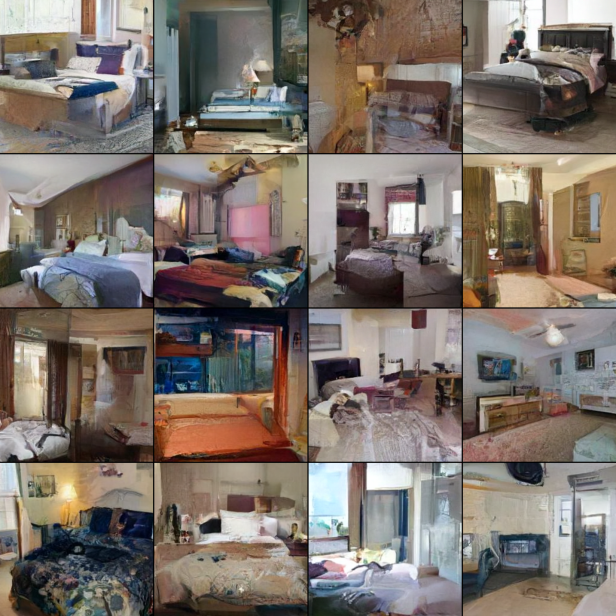}
        \vspace{-1em}
        \caption*{\scriptsize Euler 2-step}
    \end{minipage}%
    \hfill
    \begin{minipage}[b]{0.29\linewidth}
        \includegraphics[width=\linewidth]{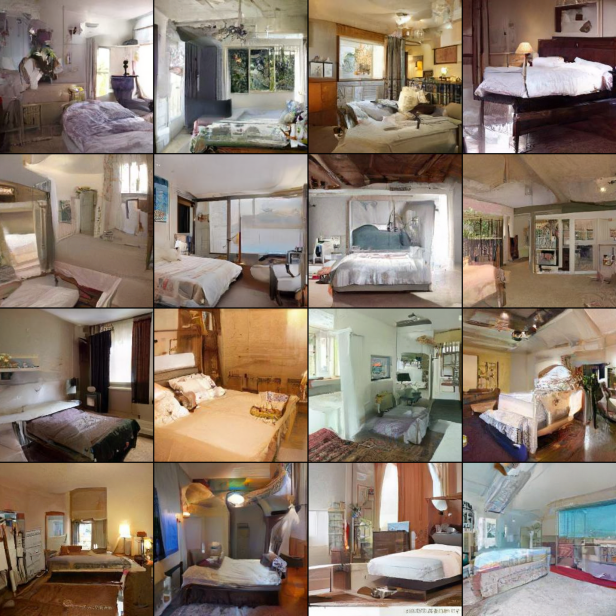}
        \vspace{-1em}
        \caption*{\scriptsize RK - 45}
    \end{minipage}
\end{minipage}
\begin{minipage}[t]{0.3\textwidth}
    \vspace*{0pt}
    \centering
    \scriptsize
    \setlength{\tabcolsep}{3pt}
    \begin{tabular}{lccc}
        \toprule
        \textbf{Solver} & \textbf{FID (Original/Ours)} & \textbf{Precision} & \textbf{Recall} \\
        \midrule
        1-step Euler &
        139.98 / \textbf{26.54} &
        0.0290 / \textbf{0.4822} &
        0.0220 / \textbf{0.2274} \\
        RK &
        24.76 / \textbf{24.14} &
        0.4525 / \textbf{0.4703} &
        0.2386 / \textbf{0.2388} \\
        \midrule
        \multicolumn{4}{c}{\textbf{IVD, Recon, P-Recon error (Fake/Real)}} \\
        \midrule
          & \textbf{IVD} & $\boldsymbol{L^{recon}_{2}}$ & $\boldsymbol{L^{p-recon}_{2}}$\\
        \midrule
        Original &
        1.1790 &
        0.0822 / 0.1147 &
        0.0820 / 0.1146 \\
        Ours &
        \textbf{0.9103} &
        \textbf{0.0487} / \textbf{0.0405} &
        \textbf{0.0486} / \textbf{0.0407} \\
        \midrule
          & $\boldsymbol{\Delta L^{recon}_{2}}$ & $\boldsymbol{\Delta L^{p-recon}_{2}}$ & \\
        \midrule
        Original & 0.0325 & 0.0326 & \\
        Ours     & \textbf{0.0083} & \textbf{0.0079} & \\
        \bottomrule
    \end{tabular}
    \label{tab:lsun_metrics_full}
\end{minipage}

\caption{Visual and quantitative comparison on LSUN. \textbf{Left}: 2-row layout showing original (\textbf{top}) and ours (\textbf{bottom}) for each solver (1-step, 2-step, RK). \textbf{Right}: evaluation metrics.}
\label{fig:lsun_qual_quant}
\vspace{-0.5em}
\end{figure*}

In this section, we assess the generalizability of our method on the LSUN Bedroom dataset \citep{yu2015lsun} at a resolution of 256$\times$256. We use the same hyperparameters, time schedule, and EMA settings as in the experiments by \citet{liu2022flow}. We use 60K fake pairs and 5K real pairs, while the original uses 120K fake pairs. Despite GPU limitations on the larger LSUN dataset, our method consistently outperformed the original in image quality. Figure~\ref{fig:lsun_qual_quant} shows superior few step generation quality, and with adaptive step sampling (RK45), our approach achieved comparable quality with significantly fewer fake pairs than the original. Further details on the training configurations are provided in Appendix~\ref{asec:config}. For comparisons under different random seeds, refer to Appendix~\ref{asec:lsun_appendix}.

\section{Related work}

\label{others}
Recent efforts to improve rectified flow models focus on modifying the time distribution, loss functions, or model architectures \citep{lee2024improving, kim2024simple}. Some works incorporate real data pairs or introduce discriminator-based regularization to reduce the effect of out-of-distribution samples \citep{kim2024simple}. Others analyze model degradation through the lens of denoising autoencoders, attributing performance drop to the vanishing of weight norms caused by repeated training on synthetic pairs \citep{zhu2024analyzing}. PerFlow takes a different approach by constructing piecewise linear intermediate paths to stabilize reflow training \citep{yan2024perflowpiecewiserectifiedflow}. In contrast to these works, our method introduces a perturbation-based supervision using real data inversions.

\section{Conclusion and future work}

We propose balanced conic reflow, a simple yet effective method that addresses key limitations of traditional rectified flow by incorporating real pairs through a Slerp-based perturbation strategy. Our approach improves generation quality across multiple settings, while requiring significantly fewer synthetic samples. Its plug-and-play nature makes it compatible with various flow-based generative models, such as InstaFlow and SD3 \citep{liu2023instaflow, sd3}.
Future directions include extending our method with additional loss functions or integrating it with diffusion-based synthetic supervision \citep{lee2024improving, kim2024simple}. Moreover, our strategy may complement recent frameworks such as Rectified Diffusion \citep{wang2024rectified}, which demonstrate that reflow-like improvements are possible even without retraining $v$-prediction-based flow matching models, where conic reflow can further help preserve real image trajectories and reduce reconstruction discrepancy without relying on linear interpolation paths or discriminator-based supervision.

\section*{Acknowledgment}
This work was partly supported by the National Research Foundation of Korea (RS-2023-00223062) and an IITP grant (RS-2020-II201361, RS-2024-00439762) funded by the Korean government (MSIT).

\newpage
\appendix

\setcounter{page}{1}

\renewcommand{\thetable}{A\arabic{table}}
\renewcommand{\thefigure}{A\arabic{figure}}
\setcounter{figure}{0}
\setcounter{table}{0}
\vspace*{0pt}

{\centering
\Large
\textbf{Balanced Conic Rectified Flow}\\
\vspace{0.5em}Supplementary Material \\
\vspace{1.0em}
}

\section{Detailed settings}
\label{asec:settings}
\paragraph{CPU and GPU settings}
All experiments were conducted on a machine with an Intel Core i9-10980XE CPU (18 cores, 3.00GHz) and four NVIDIA GeForce RTX 3090 GPUs (24GB VRAM each). The system was equipped with 128GB of DDR4 RAM.

\begin{figure}[ht]
    \centering
    \begin{minipage}{0.45\textwidth}
        \centering
        \includegraphics[width=\textwidth]{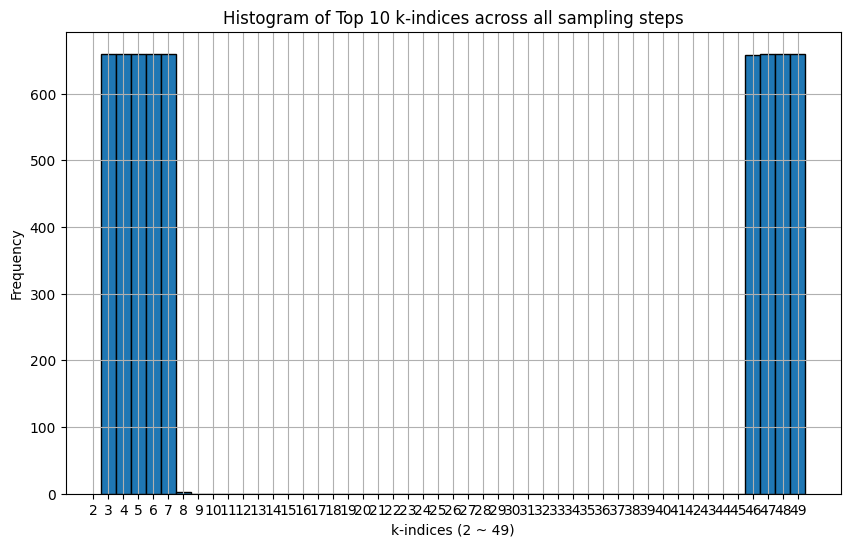}
        \subcaption{Top-K curvature indices.}
        \label{fig:top_k_cur1}
    \end{minipage}%
    \hspace{0.02\textwidth} 
    \begin{minipage}{0.45\textwidth}
        \centering
        \includegraphics[width=\textwidth]{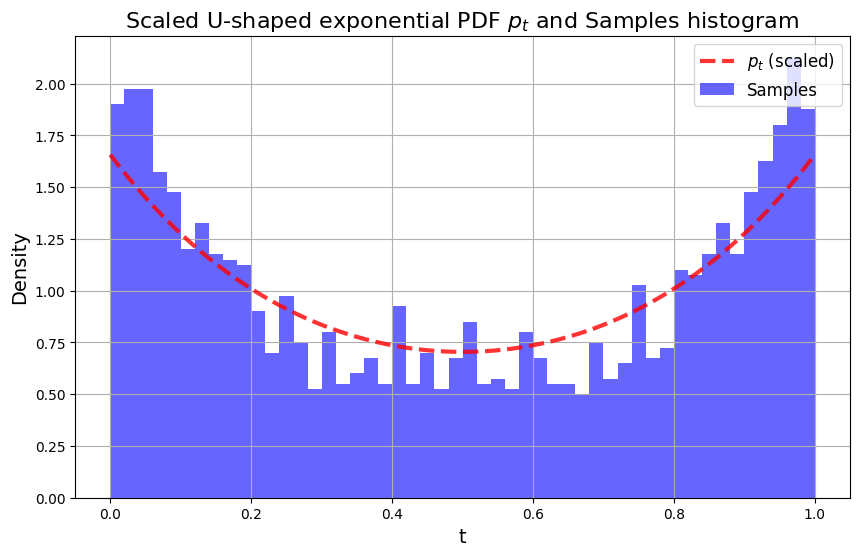}
        \subcaption{Time distribution and samples.}
        \label{fig:top_k_cur2}
    \end{minipage}
    \caption{Top-K (K=10) curvature indices and U-shape exponential time distribution with sampled data. For visualization convenience, the U-shaped probability density function $p_t$ was scaled so that the area under the curve equals 1.}
\end{figure}

\paragraph{Using exponential distribution of timesteps for training}
The trajectory crossovers during the reflow process are more frequent near the noise or image endpoints (i.e., when closer to $X_0$ or $X_1$) \cite{lee2024improving}. To focus the training more effectively on these regions with high crossover frequency, we employed an exponential distribution. This phenomenon can be clearly observed in \ref{fig:top_k_cur1}, where we computed the top-k indices of the predicted velocity during the sampling process. The curvature is significantly higher near the start and end indices, indicating more crossovers at those points. Based on this observation, we adopted an exponential distribution \footnote{More explicitly, we use $ p_t(u) \propto \exp(a u) + \exp(-a u) \,\, on\,\,  u \in [0, 1]  \,\, with\,\, a = 3 $} rather than a uniform distribution, as illustrated in \ref{fig:top_k_cur2}.

\paragraph{$U_{\text{fake}} \, \text{and} \,U_{\text{real}}$}
The training schedule for conic reflow and original reflow is split into two phases. In the first half, conic and original reflows alternate to prevent bias toward either real or fake pairs. In the second half, only the original reflow is used to address the data imbalance, as fake pairs are far more abundant than real pairs.
For instance, if total training step is $\mathbb{N}=100$, then $U_{\text{real}} = \{1,3,5,7,...,49\}$ and  $U_{\text{fake}} = \{2,4,6,8,...,50,51,52...,100\}$.



\section{Recall and Precision}
\label{asec:recall}
In this section, we evaluate the performance of our method compared to the original rectified flow using recall and precision metrics. These metrics allow us to analyze how well the generated data covers the real data distribution (recall) and how accurate the generated samples are compared to the real data (precision) \citep{kynkaanniemi2019improved}.

We conduct the evaluation on the CIFAR-10 dataset, which consists of 60,000 real images combined with 50,000 synthetic images. For sampling, we utilize Euler sampling to ensure consistency across experiments.

As shown in \ref{tab:recall}, the results highlight key differences between multi-step and 1-step sampling:
\begin{itemize}
    \item \textbf{Multi-step sampling}: Both our method and the original exhibit similar precision and recall values, indicating comparable performance in this setting.
    \item \textbf{1-step sampling}: Precision shows only a slight difference of approximately 0.85\% on average, suggesting that both methods perform nearly identically in terms of precision. However, recall reveals a more significant advantage for our method:
    \begin{itemize}
        \item On average, our method achieves 4.5\% higher recall across all settings.
        \item Specifically, for 2-Rectified Flow, our method outperforms the original by 5.5\% in recall.
    \end{itemize}
\end{itemize}

These findings indicate that while the precision of our method is nearly identical to the original, its higher recall demonstrates superior coverage of the real data distribution. Therefore, we can interpret these results as evidence that our method produces a more comprehensive and balanced representation of the underlying data distribution compared to the original.

\begin{figure}[!ht]
    \centering
    \begin{minipage}[c]{0.8\textwidth}  
        \centering
        \small
        \begin{tabular}{lccc}
        \toprule
        \textbf{Method} & \textbf{NFE} & \textbf{Precision (↑)} & \textbf{Recall (↑)} \\ \midrule

        \multicolumn{4}{l}{\textbf{2-Rectified Flow}} \\
        \multicolumn{4}{l}{\textit{Full Step Generation}} \\
        Ours & 104 & 0.691 & \textbf{0.605} (+0.005) \\
        Original & 104 & \textbf{0.696} (+0.005) & 0.600 \\ \midrule

        \multicolumn{4}{l}{\textit{One step Generation}} \\
        Ours & 1 & 0.687 & \textbf{0.583} (+0.055) \\
        Original & 1 & \textbf{0.695} (+0.008) & 0.528 \\ \midrule

        \multicolumn{4}{l}{\textbf{3-Rectified Flow}} \\
        \multicolumn{4}{l}{\textit{Full Step Generation}} \\
        Ours & 104 & 0.691 & \textbf{0.599} (+0.007) \\
        Original & 104 & \textbf{0.698} (+0.007) & 0.592 \\\midrule

        \multicolumn{4}{l}{\textit{One step Generation}} \\
        Ours & 1 & 0.682 & \textbf{0.592} (+0.03) \\
        Original & 1 & \textbf{0.691} (+0.009) & 0.562 \\

        \bottomrule
        \end{tabular}
        \captionof{table}{Comparison of 2- and 3-Rectified Flows under different NFE settings, highlighting the better results in bold.}
        \label{tab:recall}
    \end{minipage}
\end{figure}

\section{\texorpdfstring{Extreme number of reflow process $(k=4)$}{Extreme number of reflow process (k=4)}}
\label{asec:more_reflow}

In this section, we compare the generative quality, curvature, IVD of our method and the original under a setting with an extreme number of reflow processes (\(k=4\)), which is higher than the typical settings of \(k=2\) or \(k=3\).

\begin{table}[!ht]
\centering
\begin{minipage}[t]{0.48\textwidth}
    \centering
    \tiny
    \begin{tabular}{lccccc}
    \toprule
    \textbf{Method} & \textbf{NFE} & \textbf{IS (↑)} & \textbf{FID (↓)} & \textbf{Precision (↑)} & \textbf{Recall (↑)} \\ 
    \midrule

    \multicolumn{6}{l}{\textbf{4-Rectified Flow}} \\
    Ours & 100 & \textbf{9.076} & \textbf{4.195} & 0.696 & \textbf{0.585} \\
    Original & 100 & 8.951 & 4.490 & \textbf{0.705} & 0.584 \\ 
    \midrule

    Ours & 1 & \textbf{8.808} & \textbf{5.662} & \textbf{0.690} & \textbf{0.581} \\
    Original & 1 & 8.597 & 6.580 & 0.688 & 0.576 \\ 

    \bottomrule
    \end{tabular}
    \caption{Comparison of IS, FID, precision, and recall for 4-Rectified Flow between Ours and Original. The better values are highlighted in \textbf{bold}.}
    \label{tab:nfe_is_fid}
\end{minipage}%
\hfill
\begin{minipage}[t]{0.48\textwidth}
    \centering
    \small
    \begin{tabular}{lcc}
    \toprule
    \textbf{Method} & \textbf{Curvature (↓)} & \textbf{IVD (↓)} \\ 
    \midrule

    \multicolumn{3}{l}{\textbf{4-Rectified Flow}} \\
    Ours & \textbf{0.00176} & \textbf{0.20787} \\
    Original & 0.00186 & 0.21812 \\ 

    \bottomrule
    \end{tabular}
    \caption{Comparison of curvature and IVD for 4-Rectified Flow between Ours and Original rectified flow. The better values are highlighted in \textbf{bold}.}
    \label{tab:curvature_ivd}
\end{minipage}
\end{table}

As shown in Table~\ref{tab:nfe_is_fid} and \ref{tab:curvature_ivd}:
\begin{itemize}
    \item \textbf{1-Step generation quality}: Our method outperforms the original in both FID and IS, demonstrating superior generative performance.
    \item \textbf{Multi-step generation quality}: Similarly, our method shows better results compared to the original.
    \item \textbf{Additional metrics}: Our method achieves better curvature and IVD compared to the original. This indicates that our method forms a velocity field that enables the solution trajectory to be straighter during the reflow process and better preserves the direction of the initial velocity, ensuring it aligns more closely with the overall trajectory.
\end{itemize}

These findings show that our method preserves the distribution of real images while preventing bias toward fake images. Even with $k > 3$, our method improves the reflow process and makes it robust for extreme reflow step settings.

\section{Fine-tuning with real pairs improves a pre-trained rectified flow model}
\label{asec:finetune}
\begin{figure}[H]
\small
\begin{center}
\begin{minipage}{0.45\textwidth}
    \centering
    \includegraphics[width=\textwidth]{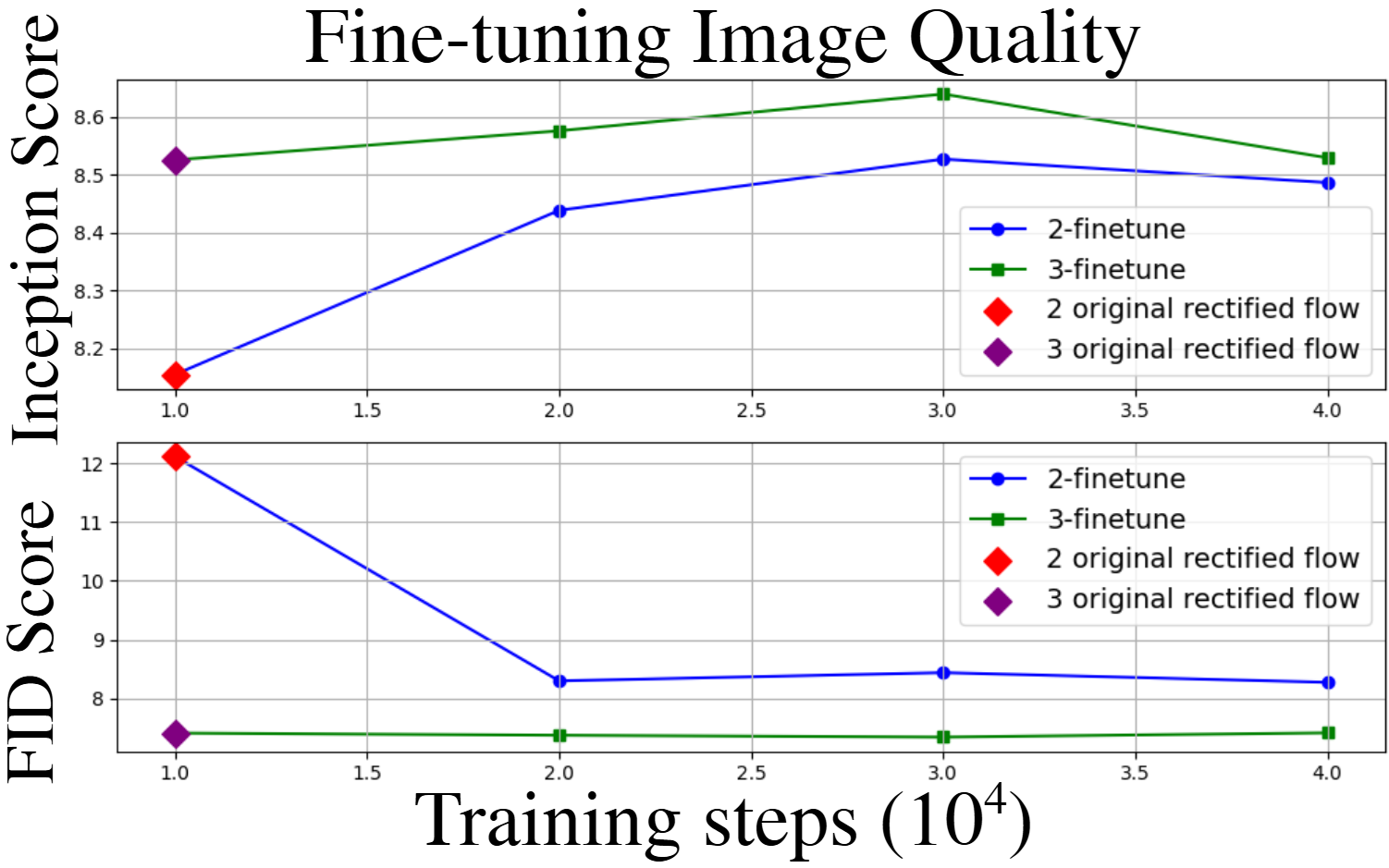}
\end{minipage}%
\hspace{0.01\textwidth} 
\begin{minipage}{0.45\textwidth}
    \centering
    \includegraphics[width=\textwidth]{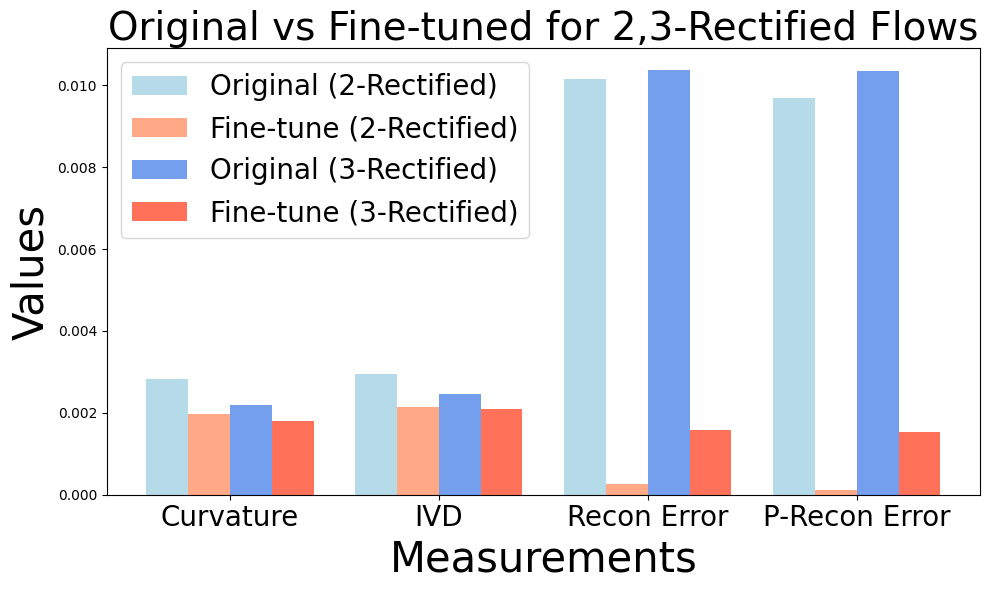}
\end{minipage}
\caption{(a) Comparison of image quality between the original rectified model and our fine-tuned model across training steps \textbf{(left)}. (b) Comparison of measurements for original and fine-tuned models for 2- and 3-rectified flows \textbf{(right)}.}
\label{fig:reflow_comparison_1}
\end{center}
\end{figure}

We demonstrate the effect of fine-tuning a pretrained rectified flow model using only 60,000 real pairs. For this experiment, we used the official rectified flow CIFAR-10 checkpoints available on GitHub.\footnote{Checkpoints are available at: \href{https://github.com/gnobitab/RectifiedFlow}{https://github.com/gnobitab/RectifiedFlow}.}. With minimal additional training, we observe a noticeable improvement in 1-step quality as shown in Figure~\ref{fig:reflow_comparison_1}(a). Additionally, both curvature and IVD values decrease rapidly as illustrated in Figure~\ref{fig:reflow_comparison_1}(b). These results show that applying our method to a 2- or 3-rectified flow model, previously trained with standard techniques, effectively improves 1-step generation quality even with a small number of real pairs. Figure~\ref{fig:reflow_comparison_1}(b) also shows fine-tuned version has lower curvature and IVD than the original, indicating that our method is straighter than the original. Furthermore, the fine-tuned version has lower recon and p-recon differences between real and fake images, indicating that our method reduces the bias toward fake samples. For visualization convenience, IVD was scaled by $10^{-2}$.

\section{Using even fewer fake samples and reflow just real pair}
\label{asec:extreme60k}
\paragraph{Extreme case (60k fake pair(original)  and + 60k fake pair Real pair(ours))}

\begin{figure}[H]
\begin{center}
\begin{minipage}{0.45\textwidth}
    \centering
    \includegraphics[width=\textwidth]{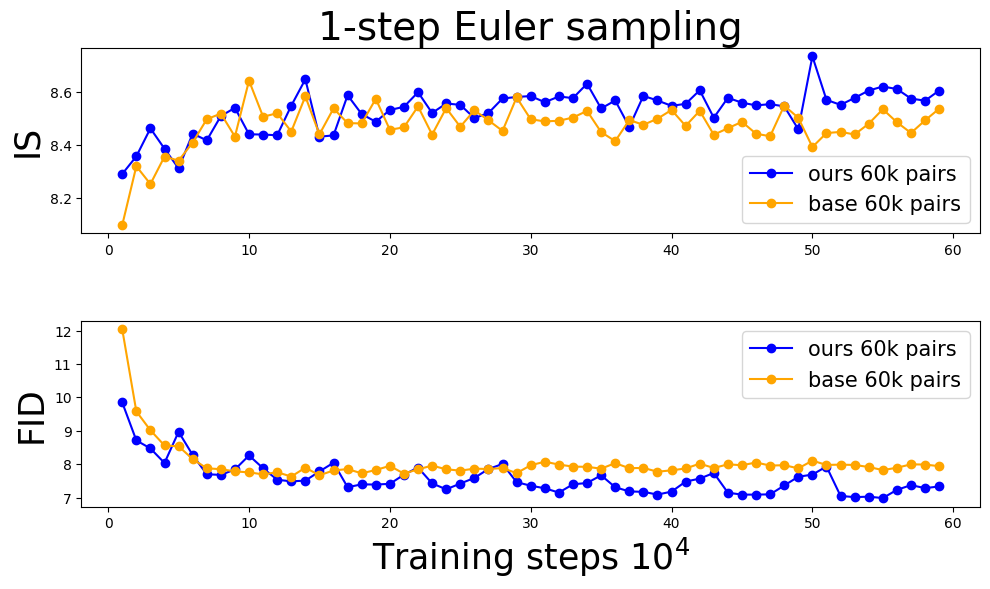}
    \text{1-step Euler sampling}
\end{minipage}%
\hspace{0.05\textwidth} 
\begin{minipage}{0.45\textwidth}
    \centering
    \includegraphics[width=\textwidth]{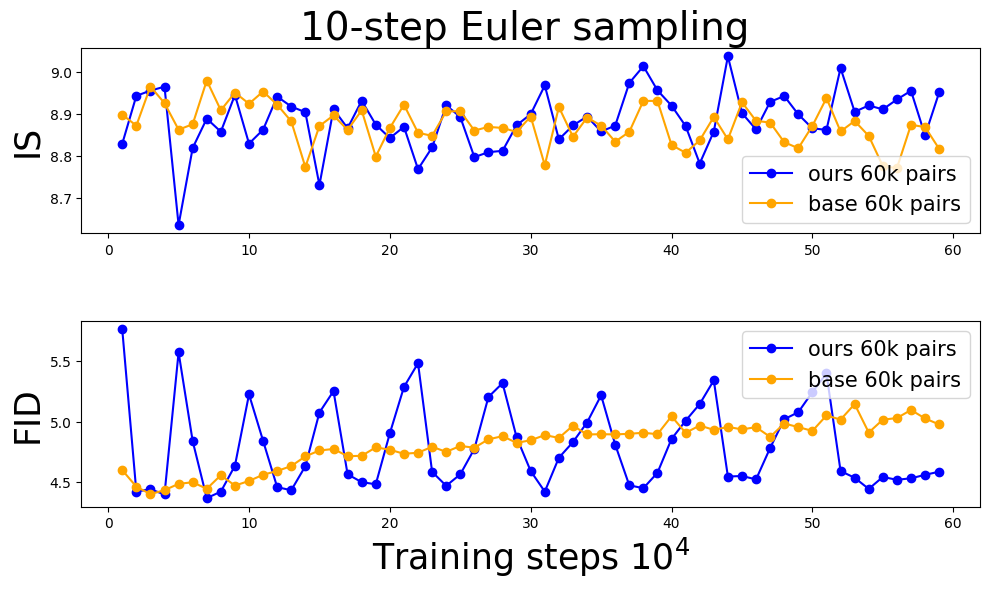}
    \text{10-step Euler sampling}
\end{minipage}
\end{center}
\caption{FID and IS score for original and ours 2-rectified flow (using 60k fake pairs)}

\vspace{-1em}
\label{fig:extreme}
\end{figure}
Figure~\ref{fig:extreme} compares the generation quality of the 2-rectified flow trained with an extremely small number of fake pairs. As shown in the figure, when the number of fake pairs is very limited, the 1-step image generation quality shows slight improvement. However, reflow training under such limited data coverage tends to overfit to the narrow fake distribution, which leads to a decline in generation quality over multiple steps, despite the gains observed in 1-step performance.

In contrast, our approach preserves the path for real images during the conic reflow step, even with an extremely small number of fake pairs (60k). This results in not only improved 1-step quality but also mitigates the degradation in multi-step generation quality compared to the original.

\paragraph{Reflow just using real pair}
We also conduct experiments under the extreme setting where reflow is trained using only real image pairs, excluding any synthetic samples. While this setup leads to slight improvements in 1-step generation quality, we observe progressive degradation in full-step performance as training continues. This limitation arises from the fact that real pairs offer accurate supervision only in the local neighborhood of the data manifold. Without the diversity and coverage provided by synthetic samples, the model struggles to generalize its velocity field across the entire generation trajectory.

To validate this observation, we compare the following three settings: (1) Real only (no Slerp), (2) Real only with Slerp-based supervision, and (3) Our full method combining real and fake pairs with Slerp. As shown in Table~\ref{tab:real_only}, setting (2) improves over (1), but both underperform compared to our full method:

\begin{table}[H]
\centering
\begin{tabular}{lcc}
\toprule
\textbf{Setting} & \textbf{1-step (IS / FID)} & \textbf{RK-step (IS / FID)} \\
\midrule
(1) Real only (no Slerp) & 8.21 / 6.93 & 8.65 / 5.13 \\
(2) Real only (+ Slerp) & 8.53 / 6.71 & 9.02 / 4.47 \\
(3) Ours (real + fake + Slerp) & \textbf{8.79} / \textbf{4.16} & \textbf{9.30} / \textbf{3.24} \\
\bottomrule
\end{tabular}
\caption{Performance comparison using only real pairs, real pairs with Slerp supervision, and our full method.}
\label{tab:real_only}
\end{table}

These results show that while Slerp-based perturbed supervision enhances the effectiveness of real image, the combination of real and synthetic samples achieves better 1-step and multi-step generation quality.

\section{\texorpdfstring{Generalization to complex dataset: Imagenet 64$\times$64}{Generalization to complex dataset: Imagenet 64x64}}

\label{asec:imagenet_append}
In this section, we evaluate how well our method generalizes on the ImageNet 64×64 dataset. We compare our method with the original by using 1) reconstruction and perturbed reconstruction errors, and 2) their difference. The evaluation shows that our approach is not only limited to smaller datasets like CIFAR-10 but also extends effectively to more challenging datasets. All precision and recall values are computed using 100k real images and 50k synthetic images. Reconstruction and perturbed reconstruction errors are obtained using a 2-step Euler solver, where reverse noise is perturbed with Slerp by $\epsilon = 0.1$. Both reconstruction and perturbed reconstruction errors are computed over a total of 12k images.

\begin{figure}[H]
\centering
\begin{minipage}{0.3\textwidth}
    \centering
    \includegraphics[width=\textwidth]{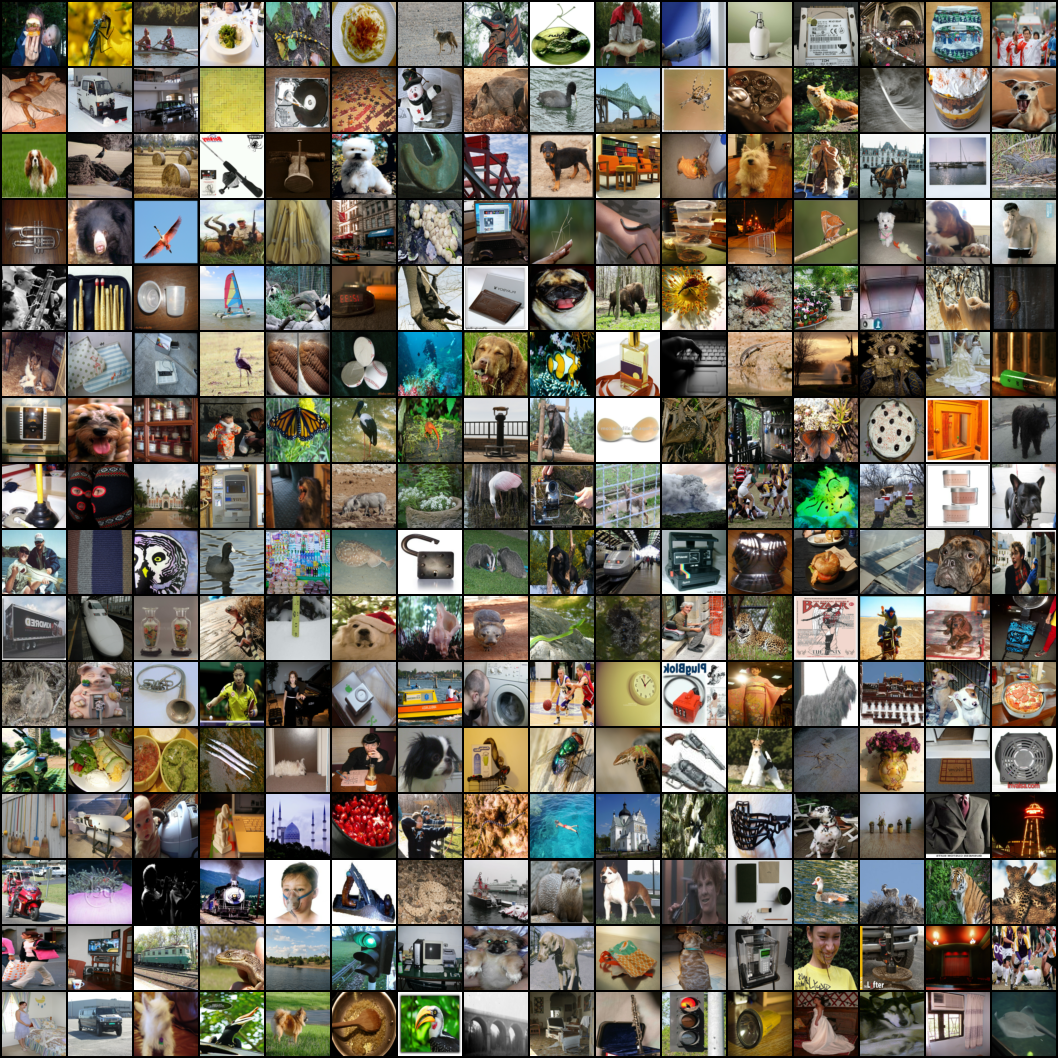}
    \subcaption{Real Image}
    \label{fig:batch_generation_ours}
\end{minipage}%
\hfill
\begin{minipage}{0.3\textwidth}
    \centering
    \includegraphics[width=\textwidth]{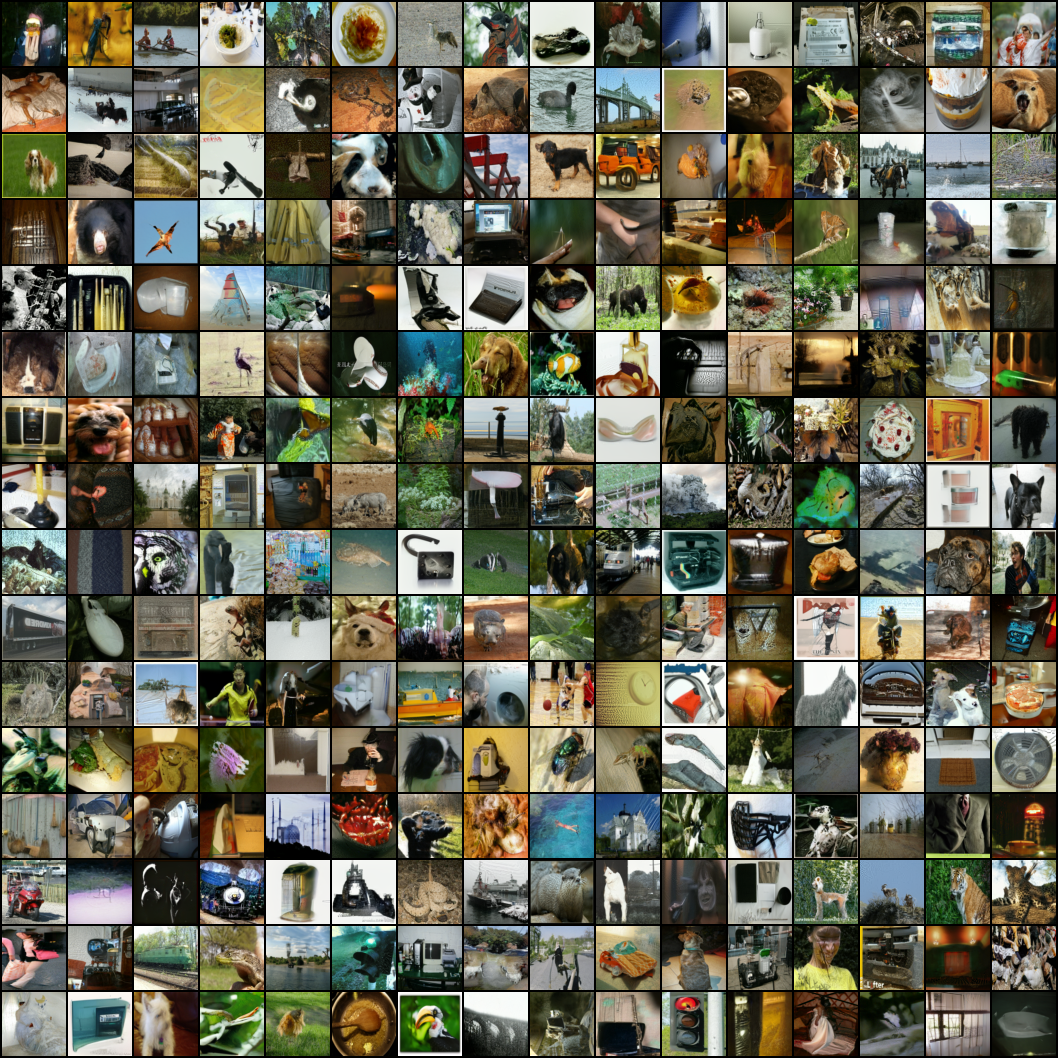}
    \subcaption{Reconstruction Image (Ours)}
    \label{fig:real_reconstruction_ours}
\end{minipage}%
\hfill
\begin{minipage}{0.3\textwidth}
    \centering
    \includegraphics[width=\textwidth]{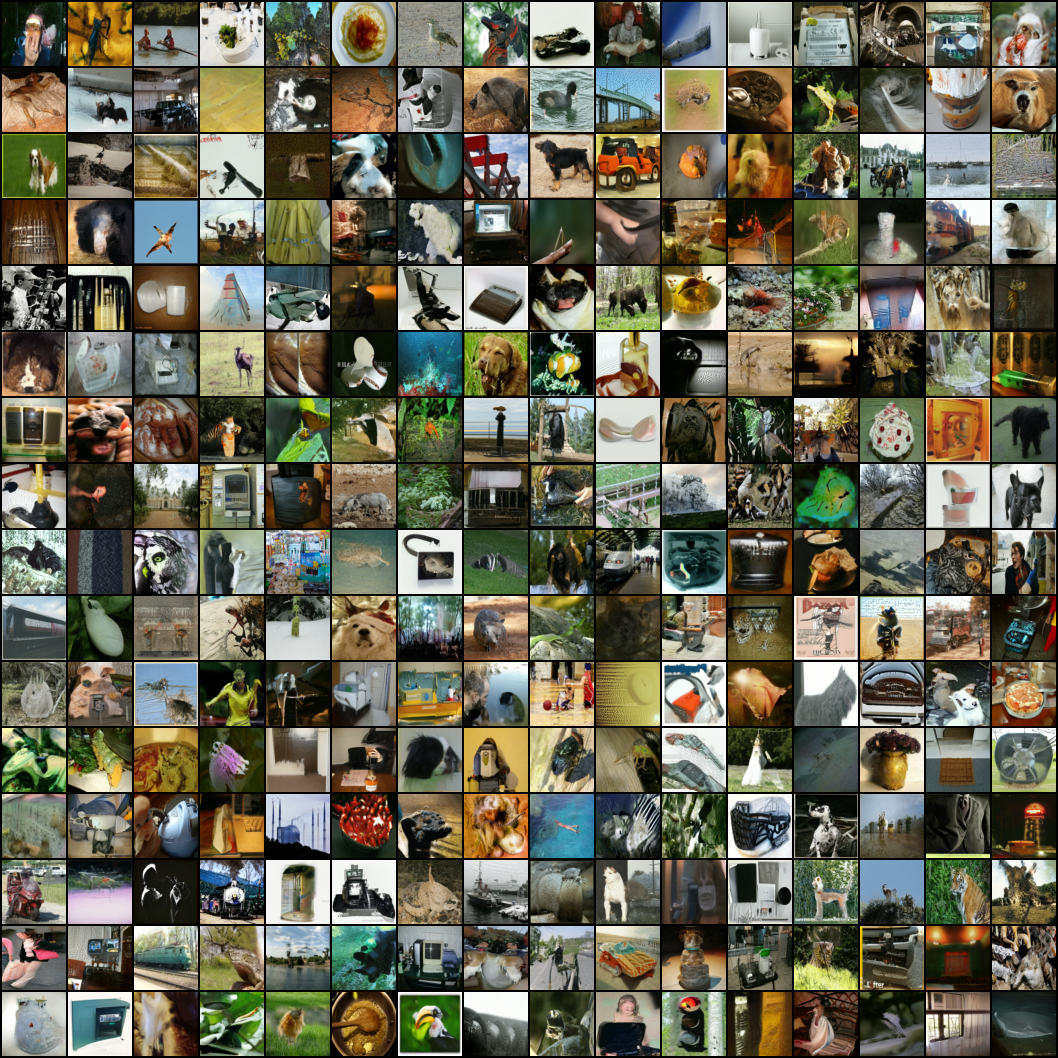}
    \subcaption{Perturbed Recon Image (Ours)}
    \label{fig:perturbed_reconstruction_ours}
\end{minipage}

\vspace{1em} 
\begin{minipage}{0.3\textwidth}
    \centering
    \includegraphics[width=\textwidth]{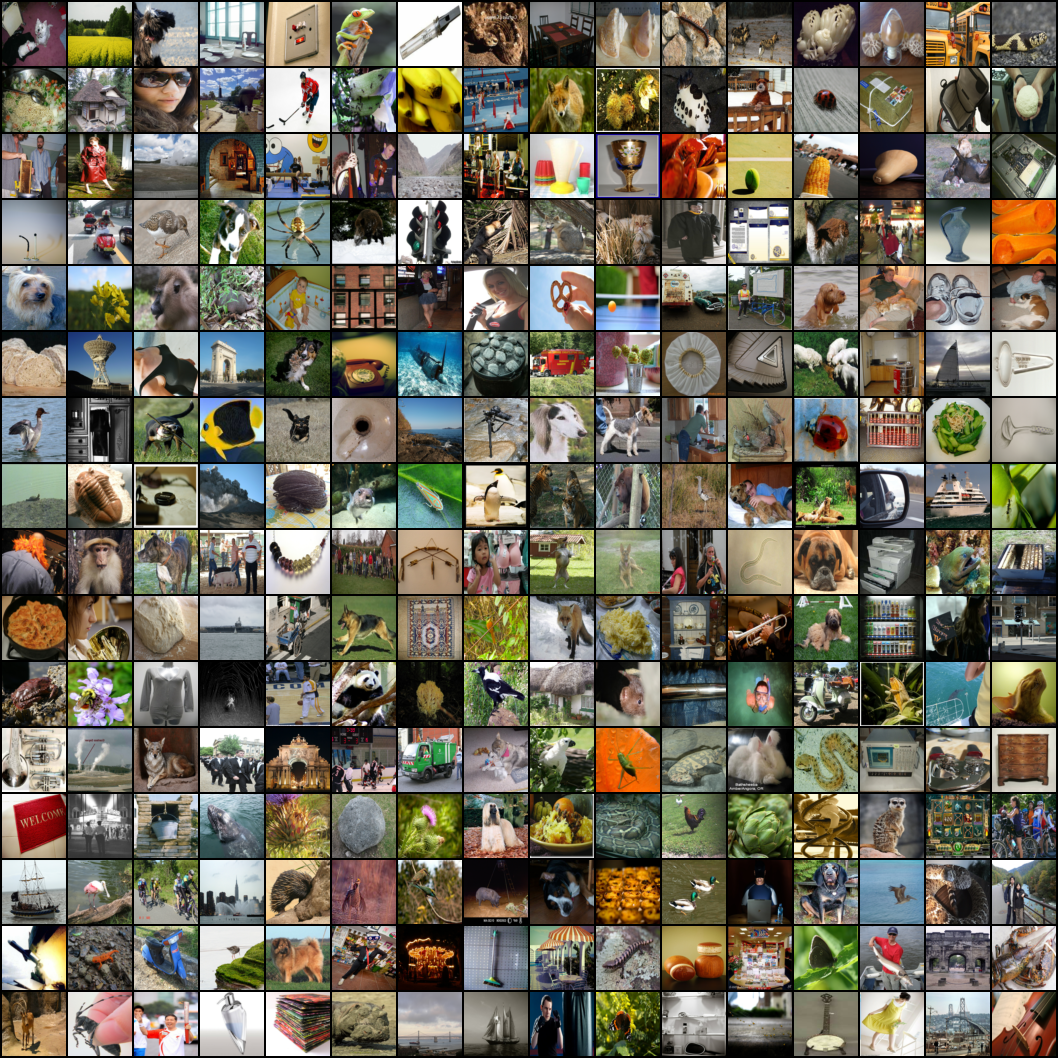}
    \subcaption{Real Image}
    \label{fig:batch_generation_base}
\end{minipage}%
\hfill
\begin{minipage}{0.3\textwidth}
    \centering
    \includegraphics[width=\textwidth]{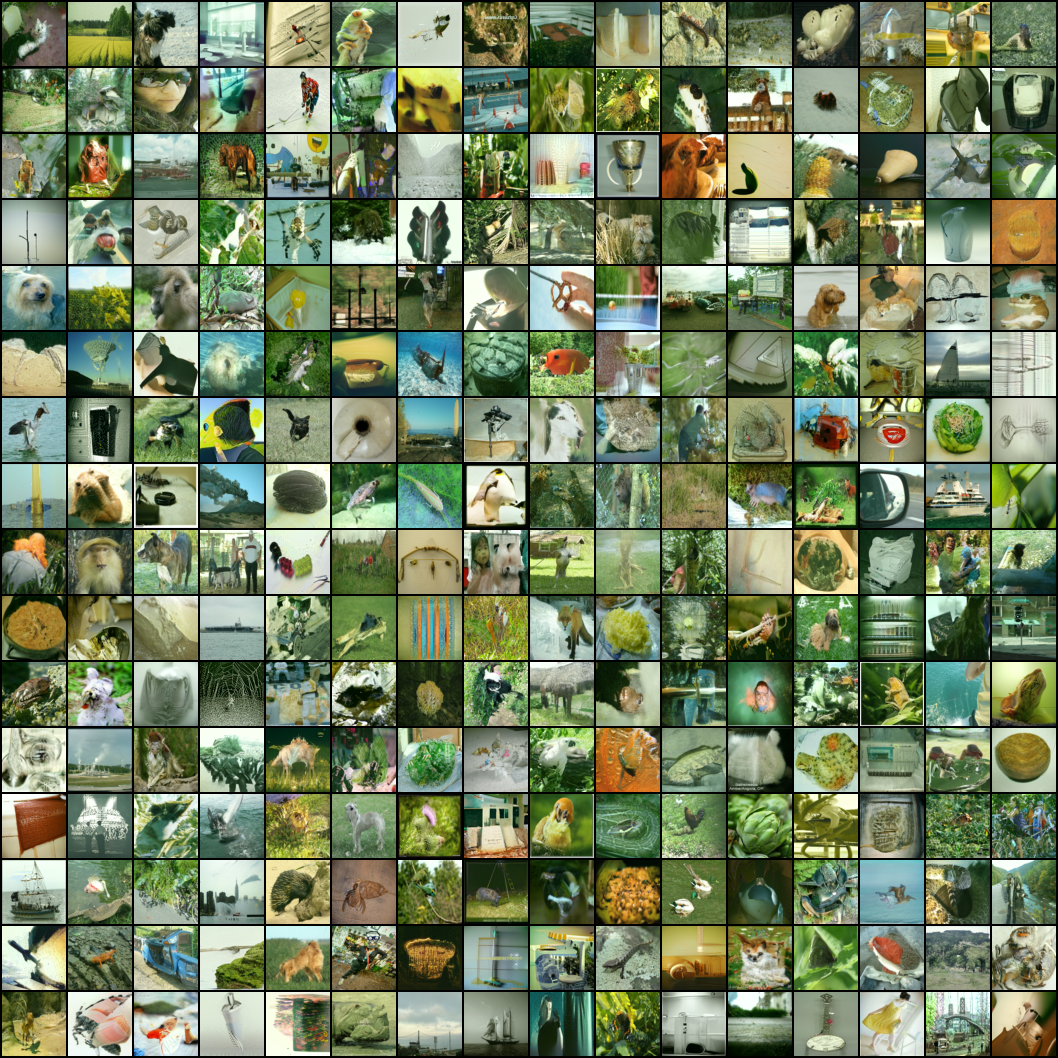}
    \subcaption{Reconstruction Image (Base)}
    \label{fig:real_reconstruction_base}
\end{minipage}%
\hfill
\begin{minipage}{0.3\textwidth}
    \centering
    \includegraphics[width=\textwidth]{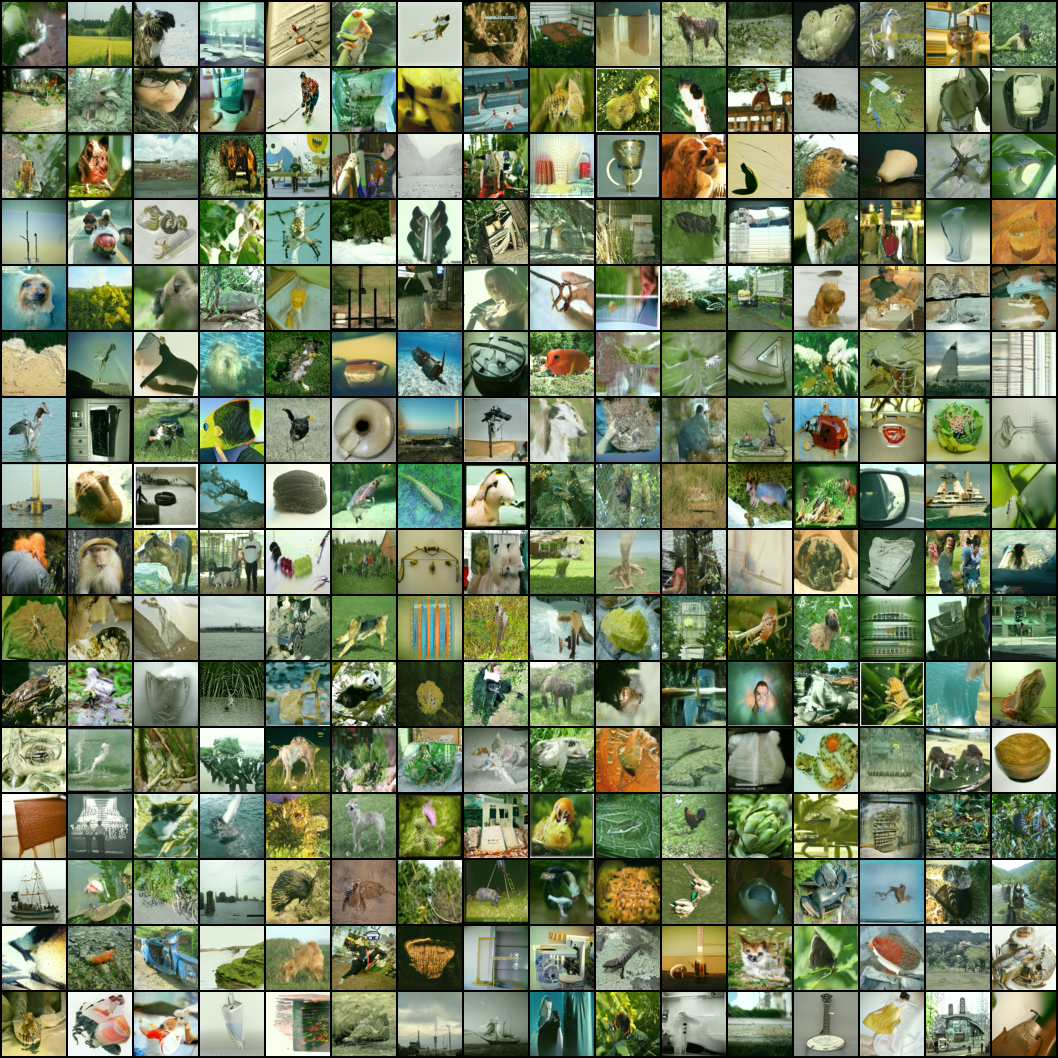}
    \subcaption{Perturbed Recon Image (Base)}
    \label{fig:perturbed_reconstruction_base}
\end{minipage}

\caption{Qualitative comparison of \textbf{real} image reconstruction and perturbed reconstruction results between the baseline and our method using the 2-rectified flow trained on the ImageNet dataset. The \textbf{bottom} row presents the original method, while the \textbf{top} row presents our method.}
\label{fig:imagenet1}
\end{figure}

\begin{figure}[H]
\centering
\begin{minipage}{0.3\textwidth}
    \centering
    \includegraphics[width=\textwidth]{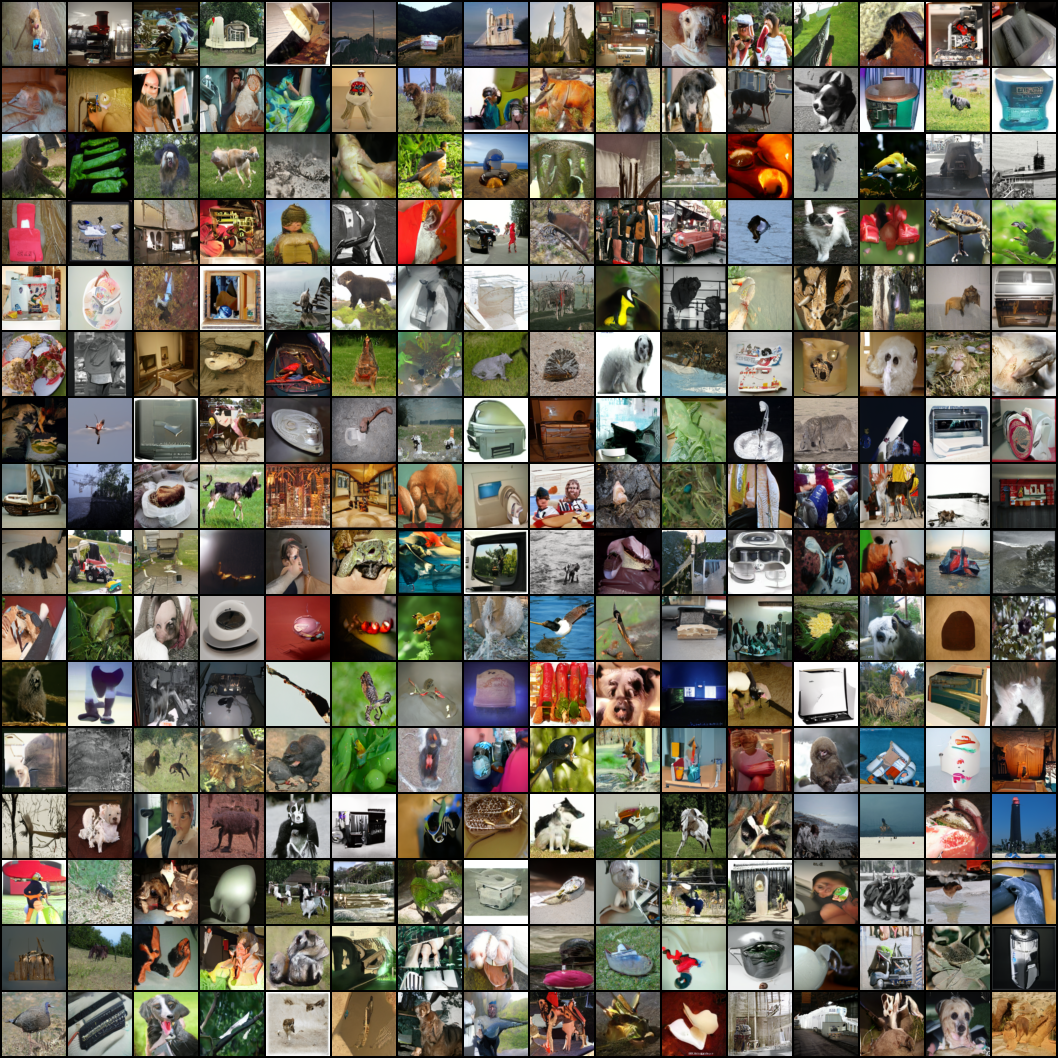}
    \subcaption{Fake Image}
    \label{fig:fake_batch_generation_ours}
\end{minipage}%
\hfill
\begin{minipage}{0.3\textwidth}
    \centering
    \includegraphics[width=\textwidth]{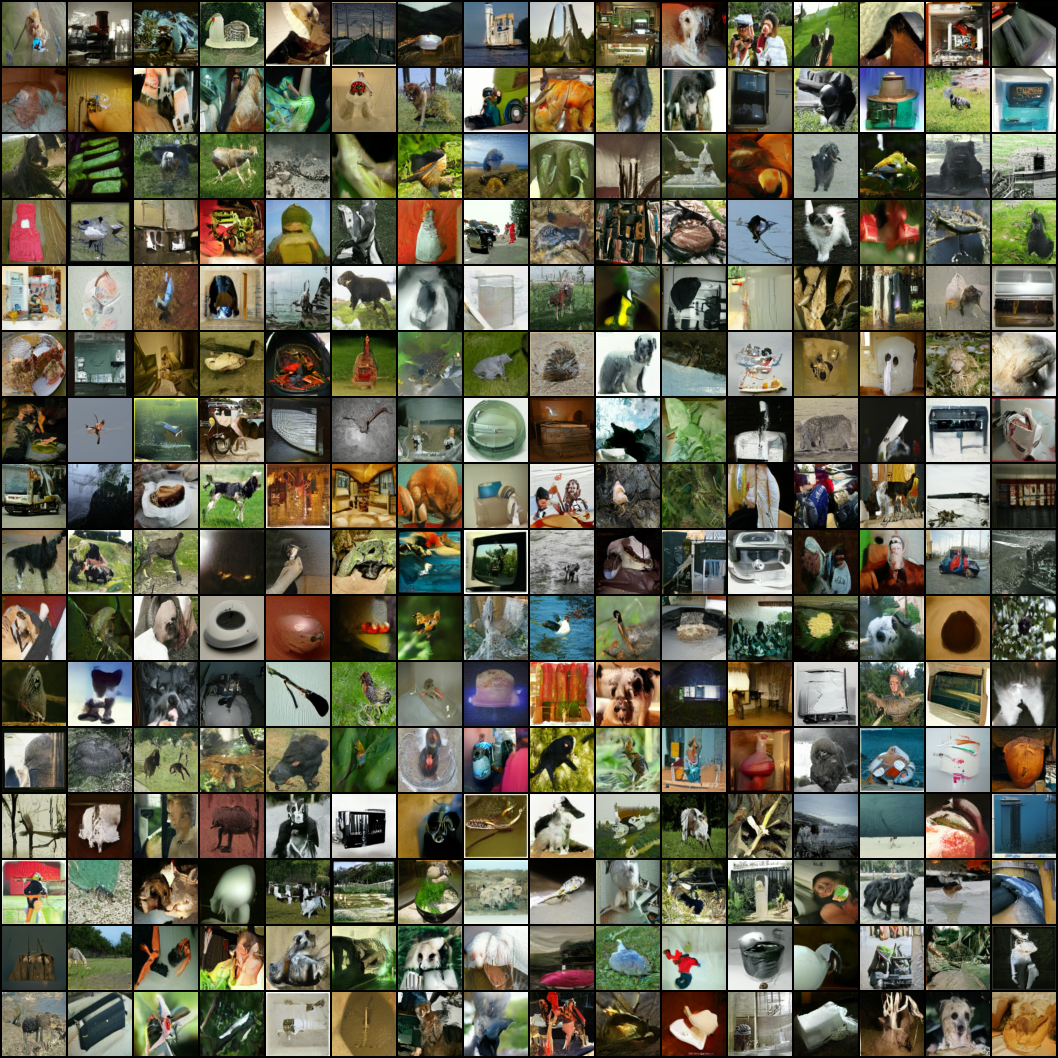}
    \subcaption{Reconstruction Image (Ours)}
    \label{fig:fake_reconstruction_ours}
\end{minipage}%
\hfill
\begin{minipage}{0.3\textwidth}
    \centering
    \includegraphics[width=\textwidth]{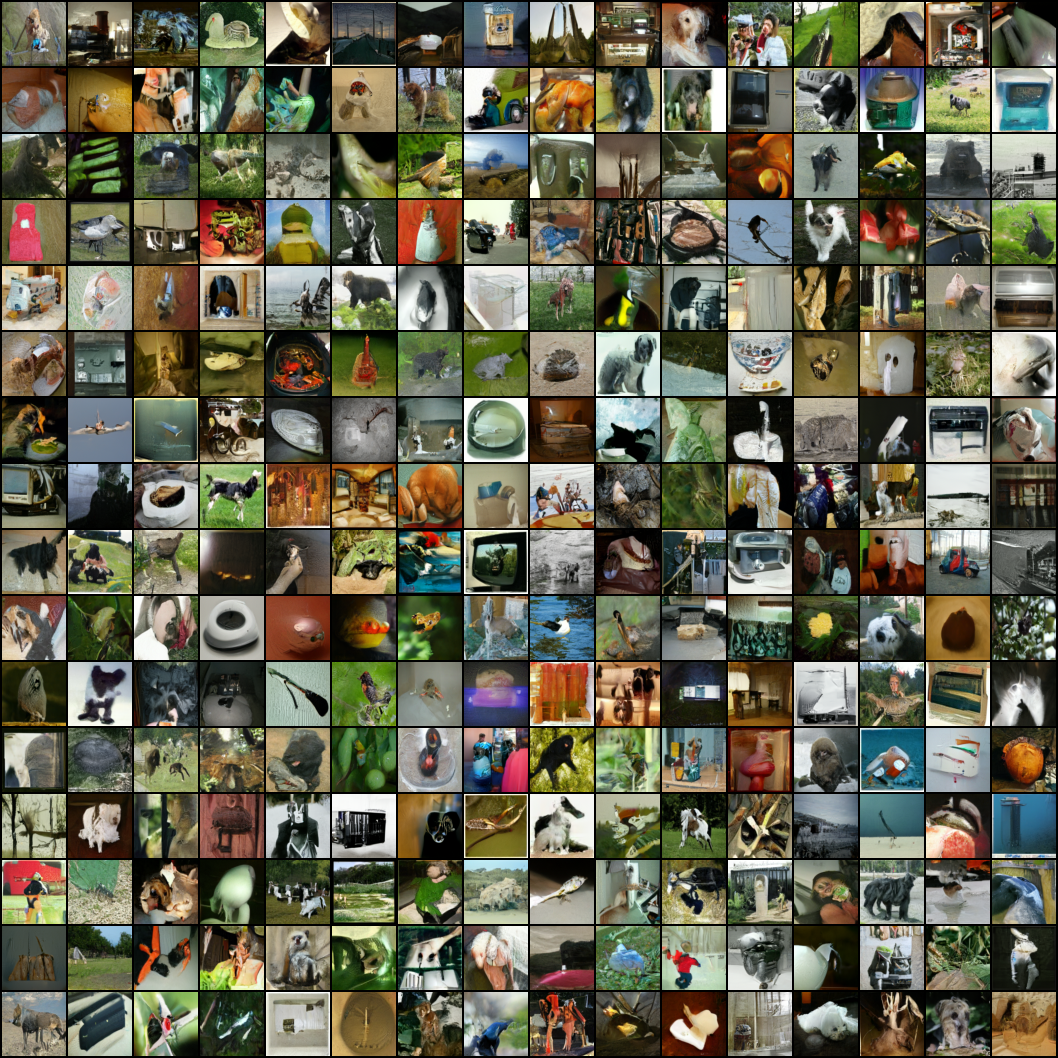}
    \subcaption{Perturbed Recon Image (Ours)}
    \label{fig:fake_perturbed_reconstruction_ours}
\end{minipage}

\vspace{1em} 

\begin{minipage}{0.3\textwidth}
    \centering
    \includegraphics[width=\textwidth]{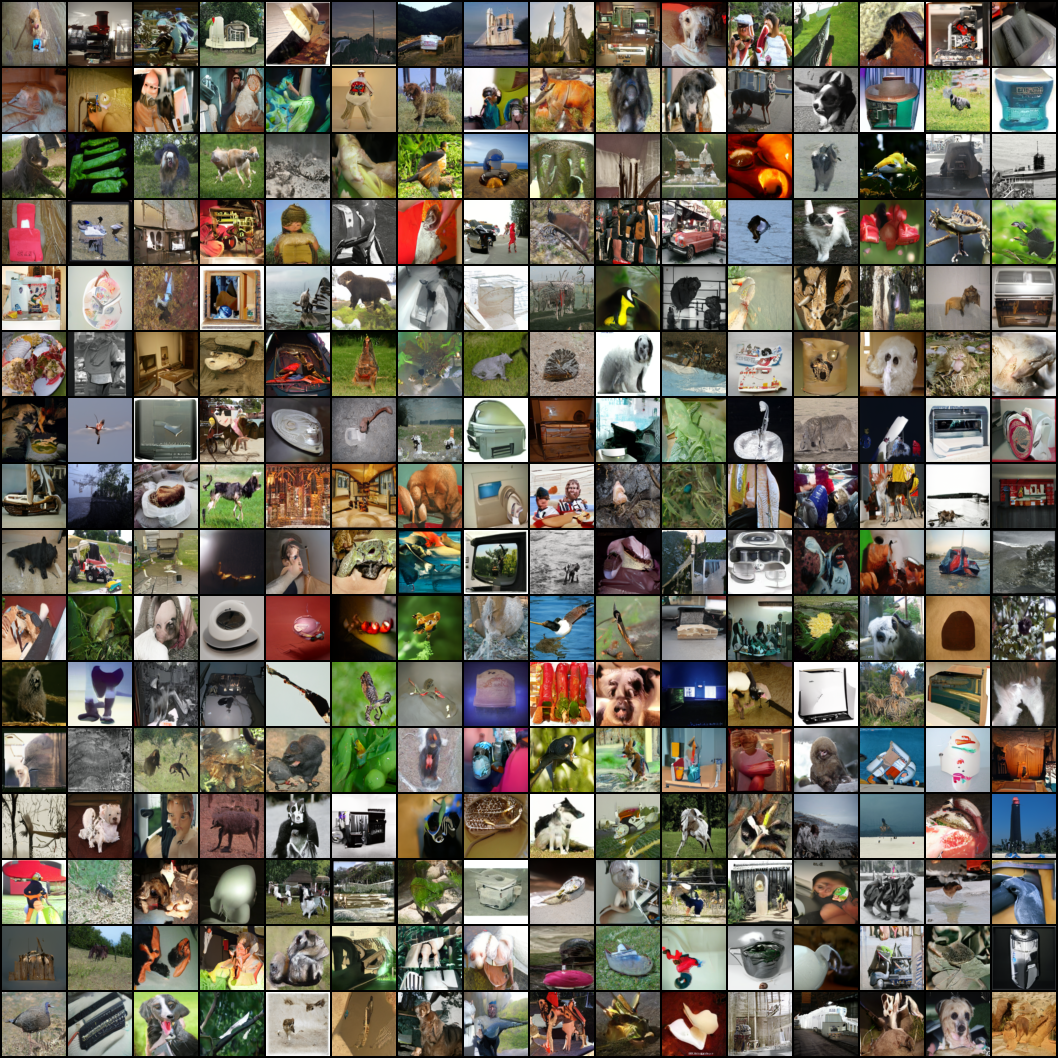}
    \subcaption{Fake Image}
    \label{fig:fake_batch_generation_base}
\end{minipage}%
\hfill
\begin{minipage}{0.3\textwidth}
    \centering
    \includegraphics[width=\textwidth]{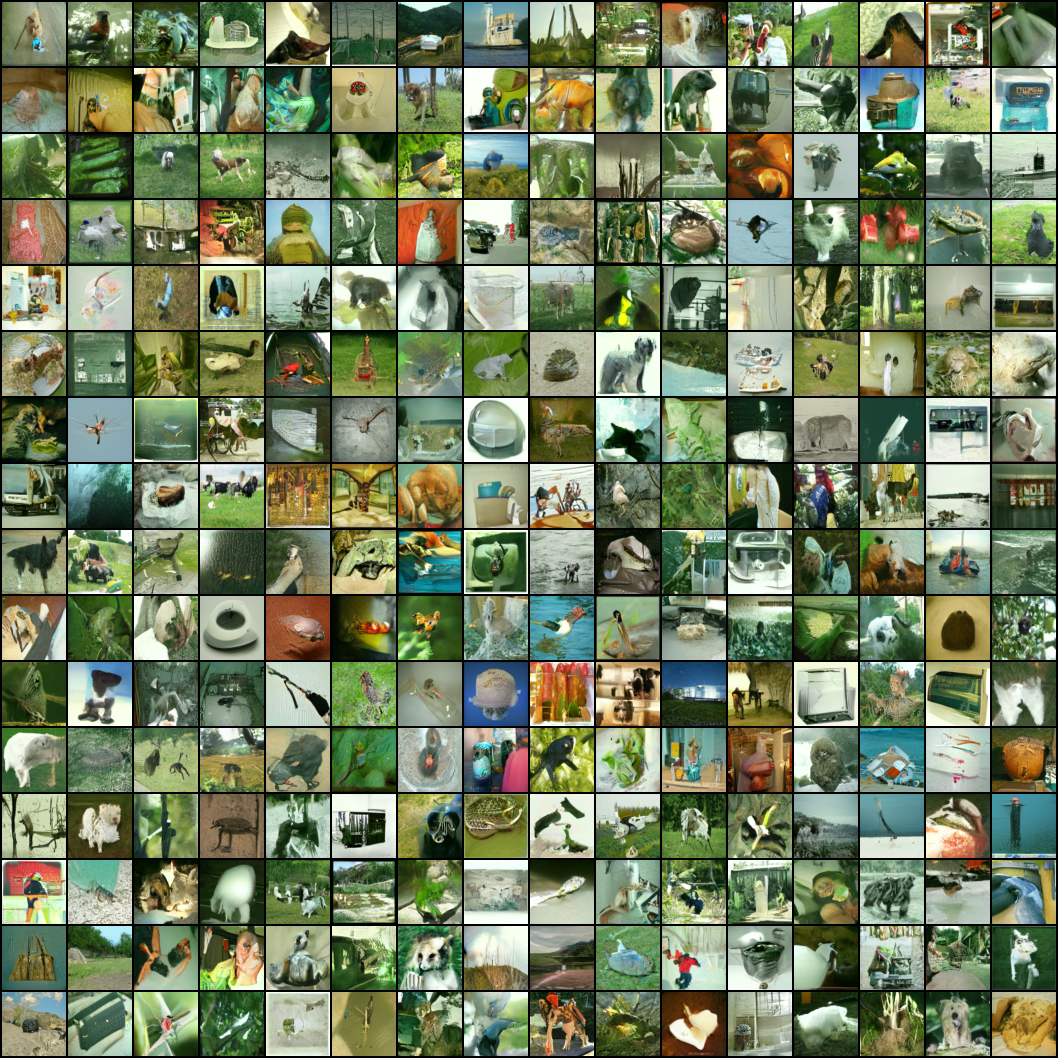}
    \subcaption{Reconstruction Image (Base)}
    \label{fig:fake_reconstruction_base}
\end{minipage}%
\hfill
\begin{minipage}{0.3\textwidth}
    \centering
    \includegraphics[width=\textwidth]{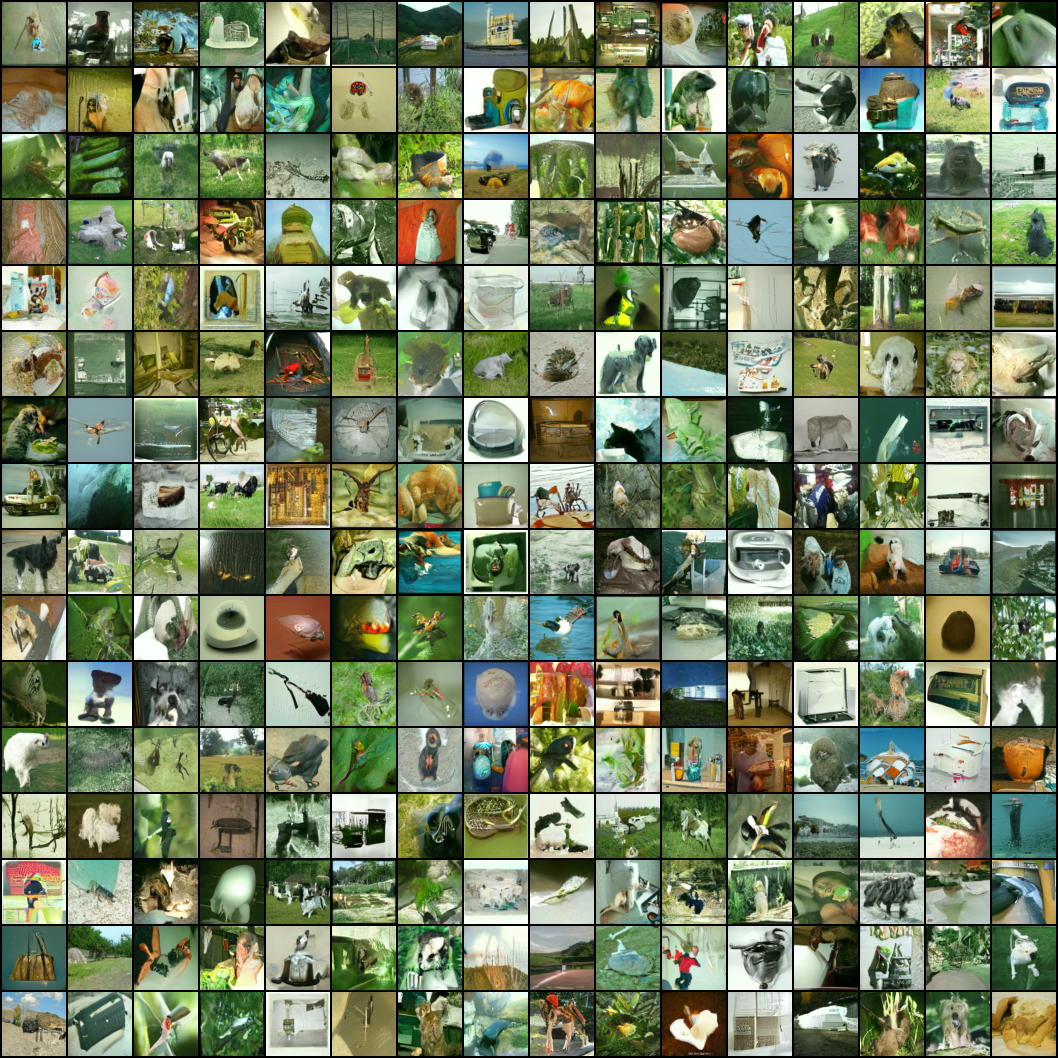}
    \subcaption{Perturbed Recon Image (Base)}
    \label{fig:fake_perturbed_reconstruction_base}
\end{minipage}

\caption{Qualitative comparison of \textbf{fake} image reconstruction and perturbed reconstruction results between the original and our method using the 2-rectified flow trained on the ImageNet dataset. The \textbf{bottom} row presents the original method, while the \textbf{top} row presents our method.}
\label{fig:imagenet2}
\end{figure}

As shown in Table~\ref{tab:imagenet-l2-errors}, our method significantly reduces reconstruction errors for both real and fake images compared to the original. This includes perturbed cases, where the gap between real and fake reconstruction errors (Difference and Difference Perturbed) is notably smaller in our approach. These results demonstrate that our method effectively mitigates the increasing discrepancy between real and fake images during the reflow process. For qualitative results on reconstruction and perturbed reconstruction images for both fake and real images, please refer to Figure~\ref{fig:imagenet1} and \ref{fig:imagenet2}.

\begin{table}[h]
    \tiny
    \centering
    \resizebox{0.9\textwidth}{!}{ 
        \begin{tabular}{lccccccc}
            \toprule
            & \textbf{Fake Recon} & \textbf{Real Recon} & \textbf{Perturbed Fake Recon} & \textbf{Perturbed Real Recon} & \textbf{Difference} & \textbf{Difference Perturbed} \\ 
            \midrule
            \textbf{Ours (Euler 2-step)} & \textbf{0.01727} & \textbf{0.02138} & \textbf{0.02735} & \textbf{0.03076} & \textbf{0.00410} & \textbf{0.00340} \\
            \textbf{Original (Euler 2-step)} &  0.02566 & 0.03273 & 0.03665 & 0.04261 & 0.00707 & 0.00596 \\
            \bottomrule
        \end{tabular}
    }
    \caption{Comparison of Reconstruction errors and differences between Original and Ours on ImageNet 64x64 using the Euler 2-step solver.}
    \label{tab:imagenet-l2-errors}
\end{table}
\section{RF++\textdagger{} and RF++\textdagger{}(+Ours) extended training results (500K steps)}
\label{asec:aftet500k}

In this section, we report extended training results beyond 300K iterations to demonstrate that our method continues to outperform the original RF++$^\dagger$.

As shown in Table~\ref{tab:rfpp_500k}, RF++$^\dagger$(+Ours) achieves better FID and IS scores compared to RF++$^\dagger$. This suggests that our Slerp-based perturbation method leads to more efficient reflow training, effectively preserving straight paths to real image distributions. All experiments use the same training configurations as described in Appendix~\ref{asec:config}.

\vspace{0.5em}
\begin{table}[h]
\centering
\small
\begin{tabular}{lcccc}
\toprule
\textbf{Model} & \textbf{Steps} & \textbf{IS (↑)} & \textbf{FID (↓)} & \textbf{Solver} \\
\midrule
RF++$^\dagger$(\textbf{+Ours})      & 1-step  & \textbf{9.15} & \textbf{3.84} & Heun 2nd \\
RF++$^\dagger$             & 1-step  & 9.04 & 4.14 & Heun 2nd \\
RF++$^\dagger$(\textbf{+Ours})      & 2-step  & \textbf{9.36} & \textbf{3.03} & Heun 2nd \\
RF++$^\dagger$             & 2-step  & 9.24 & 3.16 & Heun 2nd \\
\bottomrule
\end{tabular}
\caption{Extended training results (500K steps) for RF++$^\dagger$ and RF++$^\dagger$(+Ours).}
\label{tab:rfpp_500k}
\end{table}

\section{Qualitative results of reconstruction and perturbed reconstruction}
\label{asec:qual2}

\begin{figure}[H]
    \begin{center}
        \begin{minipage}{0.7\textwidth}
            \centering
            \includegraphics[width=\textwidth]{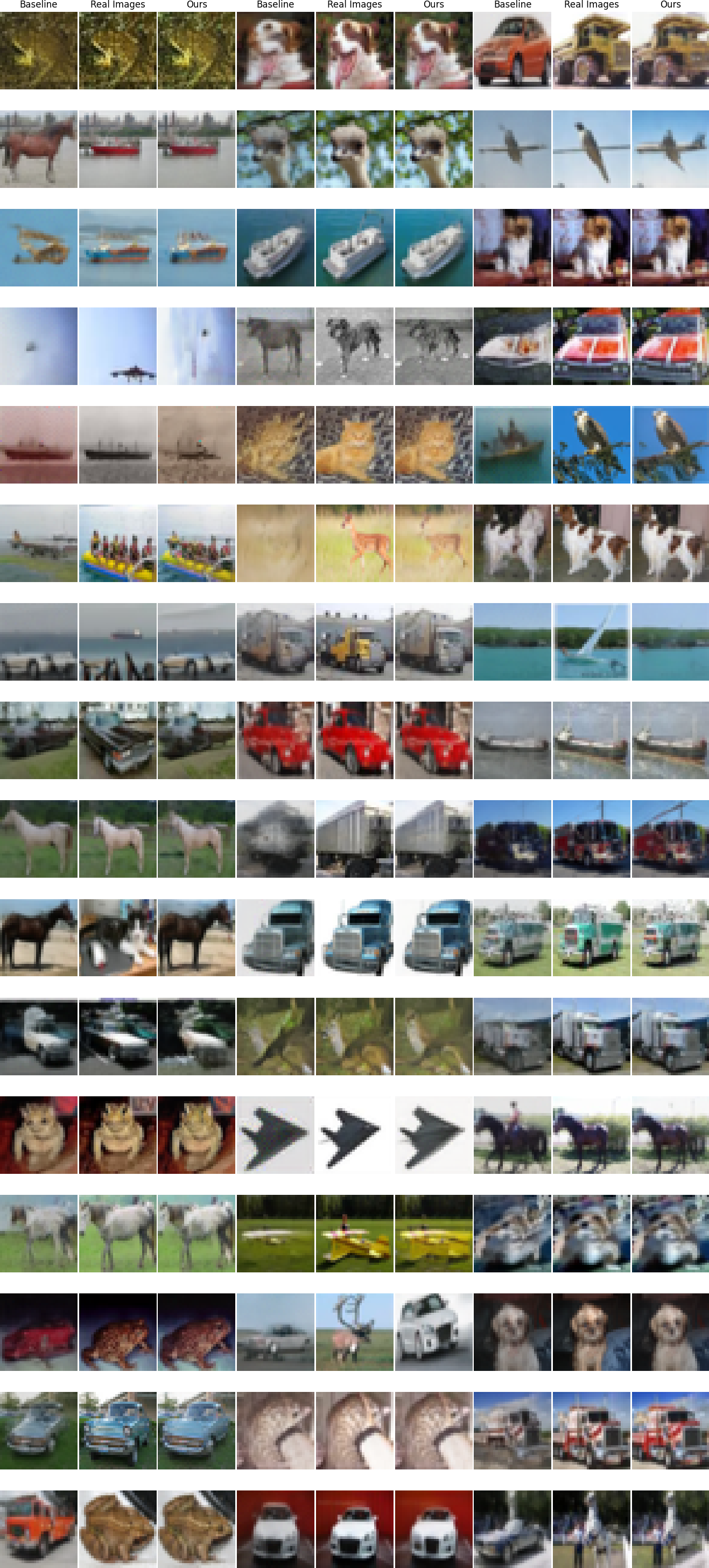}
        \end{minipage}
        \caption{Compare with reconstruction image to original (1-step Euler)}
        \label{fig:appendix-cifer10-1}
    \end{center}
\end{figure}

\begin{figure}[H]
    \begin{center}
        \begin{minipage}{0.7\textwidth}
            \centering
            \includegraphics[width=\textwidth]{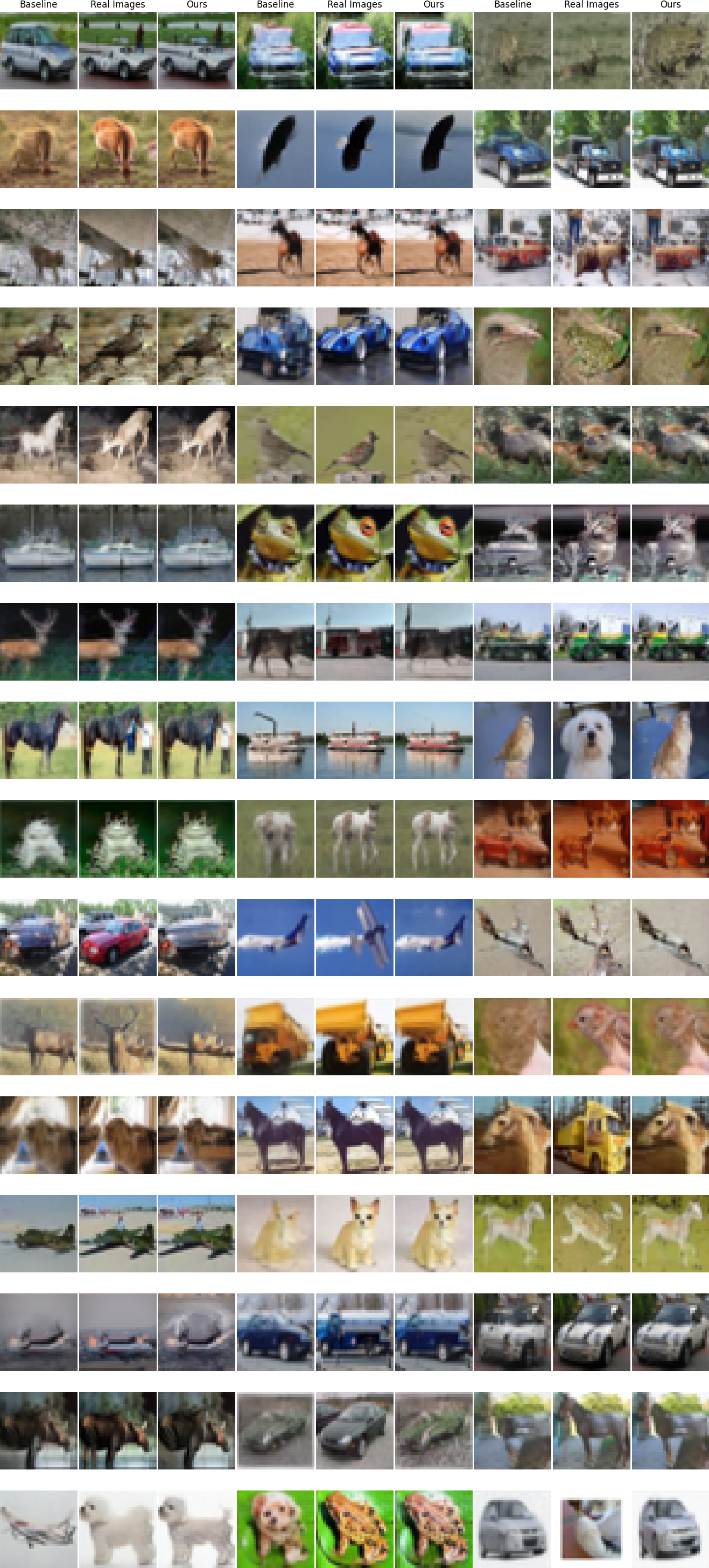}
        \end{minipage}
        \caption{Compare with reconstruction image to original (1-step Euler)}
         \label{fig:appendix-cifer10-2}
    \end{center}
\end{figure}

\begin{figure}[H]
    \begin{center}
        \begin{minipage}{0.7\textwidth}
            \centering
            \includegraphics[width=\textwidth]{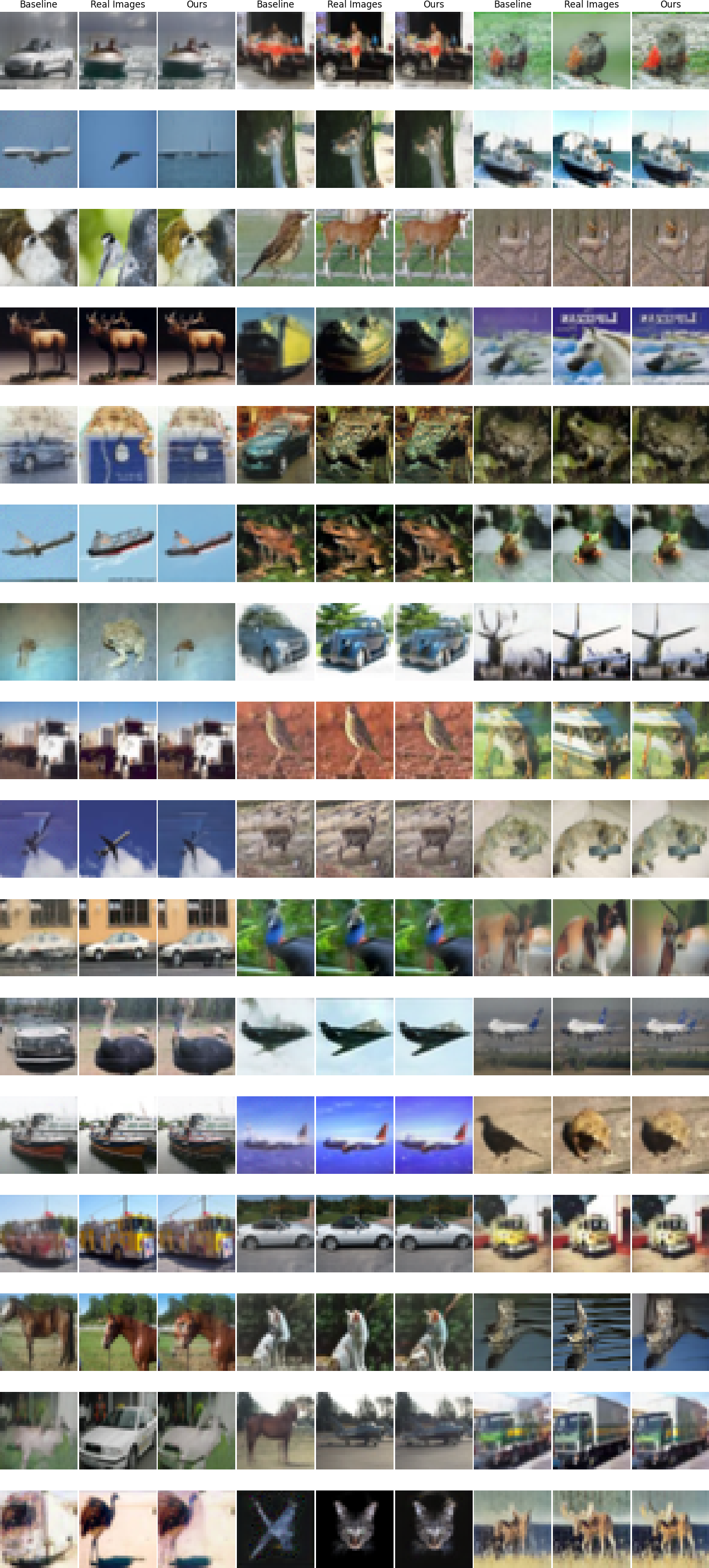}
        \end{minipage}
        \caption{Compare with reconstruction image to original (1-step Euler)}
        \label{fig:appendix-cifer10-3}
    \end{center}
\end{figure}

\begin{figure}[H]
    \begin{center}
        \begin{minipage}{0.7\textwidth}
            \centering
            \includegraphics[width=\textwidth]{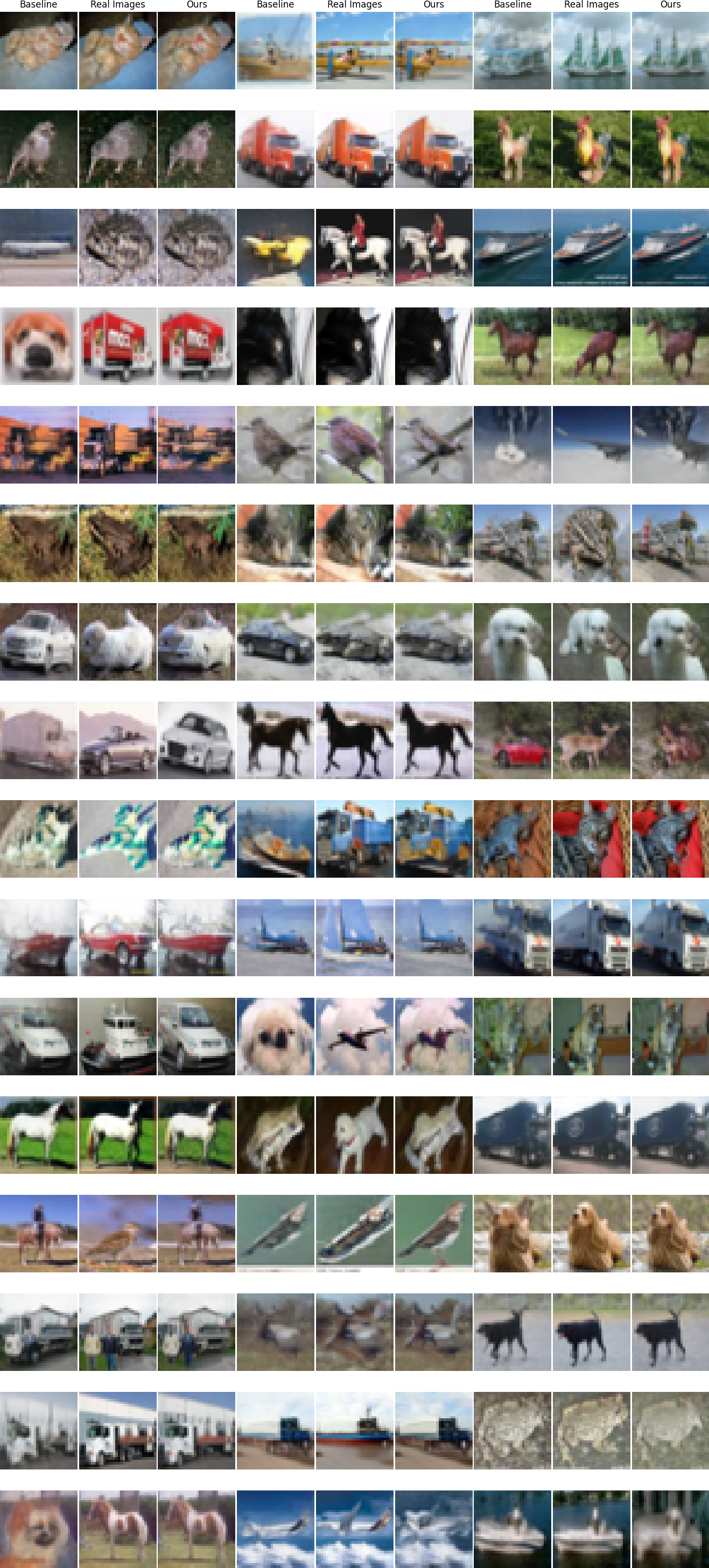}
        \end{minipage}
        \caption{Compare with perturbed ($\varepsilon = 0.04$) reconstruction image to original (1-step Euler)}
        \label{fig:appendix-cifer10-p1}
    \end{center}
\end{figure}

Figure~\ref{fig:appendix-cifer10-1}, \ref{fig:appendix-cifer10-2}, \ref{fig:appendix-cifer10-3} provide a qualitative comparison of reconstruction between the original and ours. Figure~\ref{fig:appendix-cifer10-p1} provide a qualitative comparison of the perturbed reconstruction between the original and ours (Cifar 10). As shown in the figures, the rectified flow model trained with our reflow procedure preserves the structure of real objects more effectively, whereas the original generates images that deviate significantly from the original object after undergoing reconstruction or perturbed reconstruction.


\section{Qualitative comparison on Lsun dataset}
\label{asec:lsun_appendix}
\begin{figure}[H]
\begin{center}
\begin{minipage}{0.75\textwidth}
    \centering
    \includegraphics[width=\textwidth]{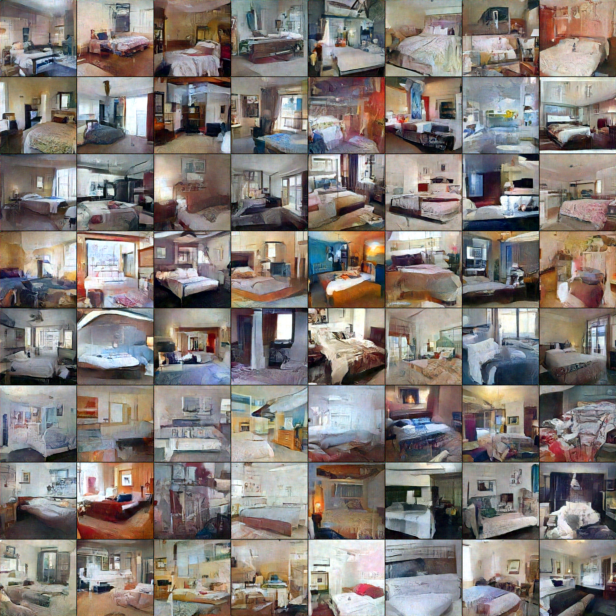}
\end{minipage}

\vspace{1em} 
\rule{0.75\textwidth}{0.5pt} 
\vspace{1em} 

\begin{minipage}{0.75\textwidth}
    \centering
    \includegraphics[width=\textwidth]{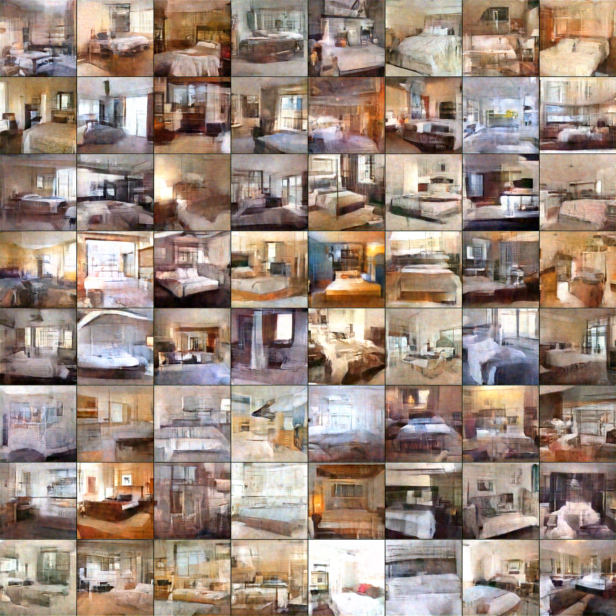}
\end{minipage}%
    \caption{Ours (\textbf{up}) and Original (\textbf{down}) 2-rectified flow (1-step, seed 1)}
    \label{fig:lsun1}
\end{center}
\end{figure}

\begin{figure}[H]
\begin{center}

\begin{minipage}{0.75\textwidth}
    \centering
    \includegraphics[width=\textwidth]{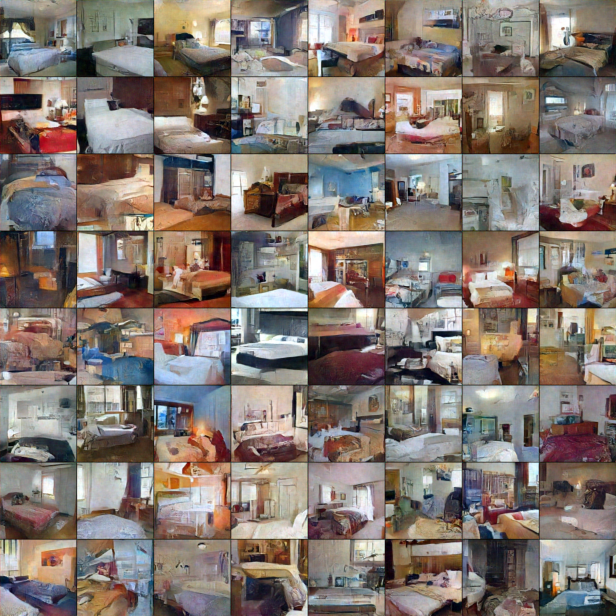}
\end{minipage}

\vspace{1em} 
\rule{0.75\textwidth}{0.5pt} 
\vspace{1em} 

\begin{minipage}{0.75\textwidth}
    \centering
    \includegraphics[width=\textwidth]{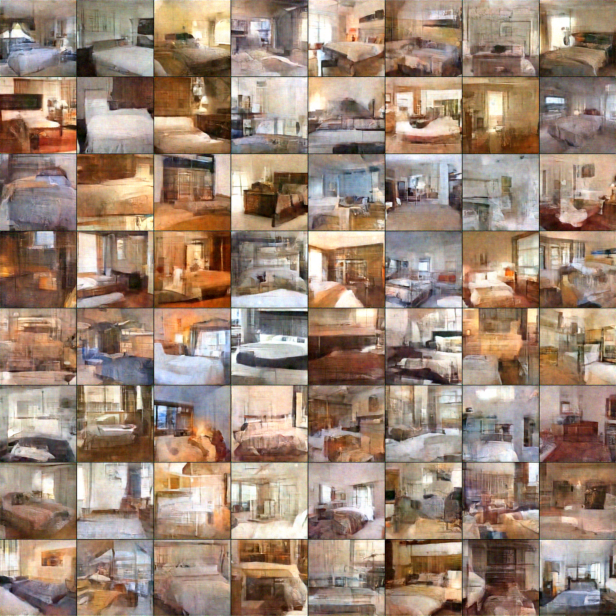}
\end{minipage}%

\caption{Ours (\textbf{up}) and Original (\textbf{down}) 2-rectified flow (1-step, seed 2)}
\label{fig:lsun2}
\end{center}
\end{figure}

\begin{figure}[H]
\begin{center}
\begin{minipage}{0.75\textwidth}
    \centering
    \includegraphics[width=\textwidth]{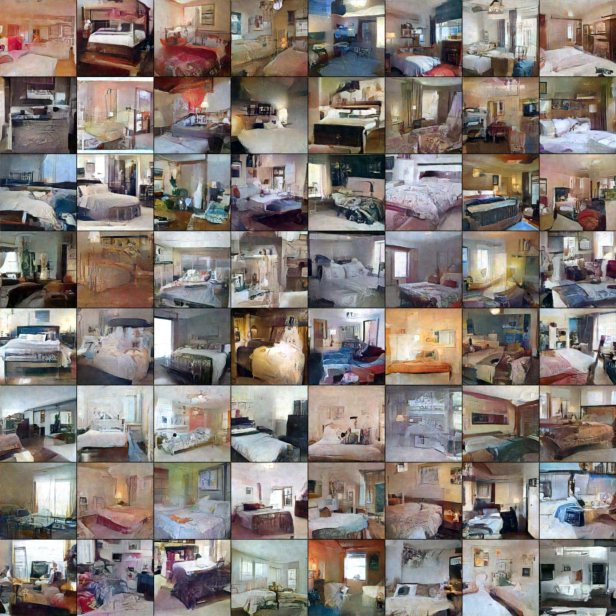}
\end{minipage}

\vspace{1em} 
\rule{0.75\textwidth}{0.5pt} 
\vspace{1em} 

\begin{minipage}{0.75\textwidth}
    \centering
    \includegraphics[width=\textwidth]{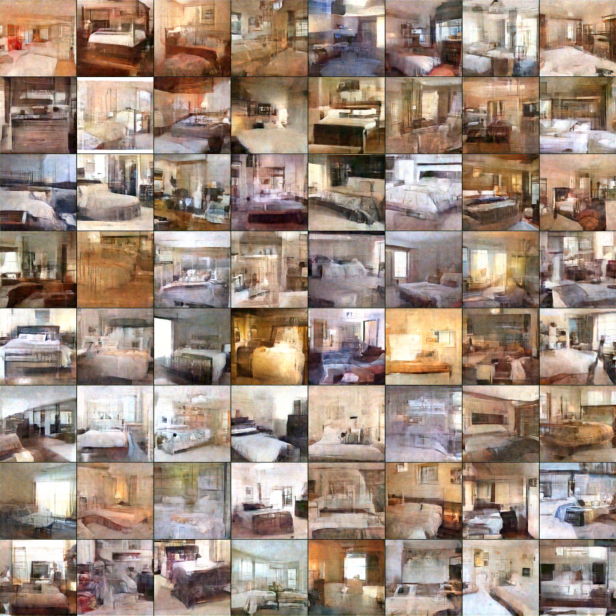}
\end{minipage}

\caption{Ours (\textbf{up}) and Original (\textbf{down}) 2-rectified flow (1-step, seed 3)}
\label{fig:lsun3}
\end{center}
\end{figure}

\begin{figure}[H]
\begin{center}
\begin{minipage}{0.75\textwidth}
    \centering
    \includegraphics[width=\textwidth]{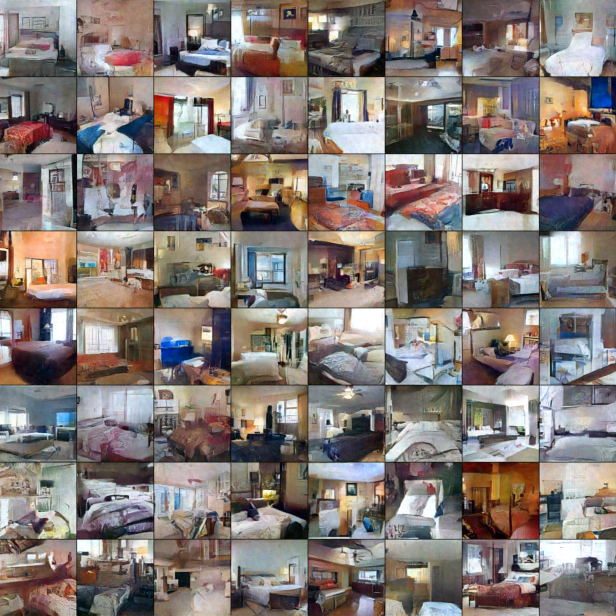}
\end{minipage}

\vspace{1em} 
\rule{0.75\textwidth}{0.5pt} 
\vspace{1em} 

\begin{minipage}{0.75\textwidth}
    \centering
    \includegraphics[width=\textwidth]{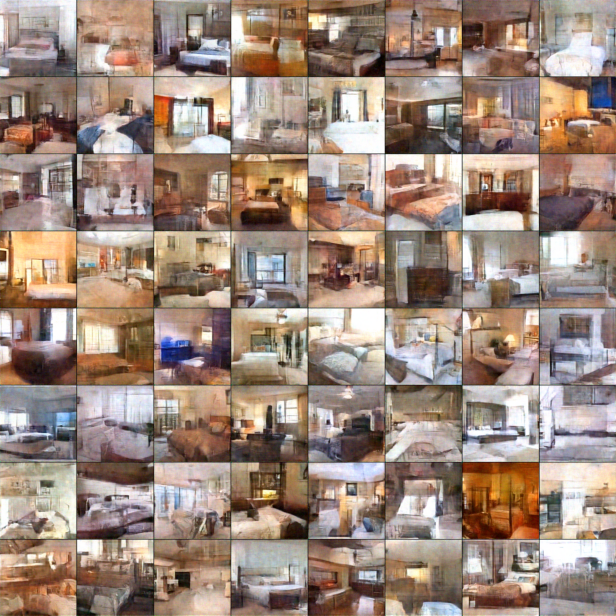}
\end{minipage}

\caption{Ours (\textbf{up}) and Original (\textbf{down}) 2-rectified flow (1-step, seed 333)}
\label{fig:lsun4}
\end{center}
\end{figure}

\begin{figure}[H]
\begin{center}
\begin{minipage}{0.75\textwidth}
    \centering
    \includegraphics[width=\textwidth]{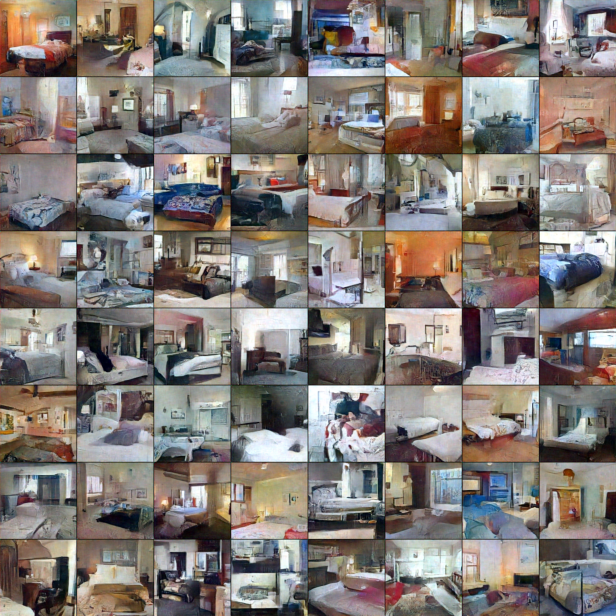}
\end{minipage}

\vspace{1em} 
\rule{0.75\textwidth}{0.5pt} 
\vspace{1em} 

\begin{minipage}{0.75\textwidth}
    \centering
    \includegraphics[width=\textwidth]{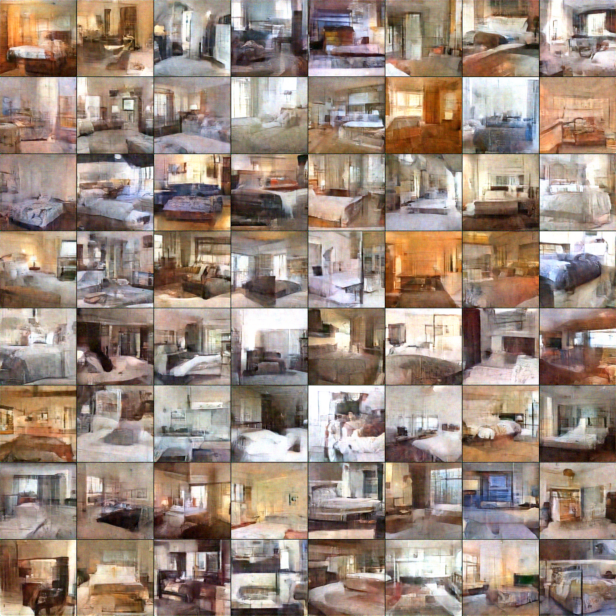}
\end{minipage}%

\caption{Ours (\textbf{up}) and Original (\textbf{down}) 2-rectified flow (1-step, seed 555)}
\label{fig:lsun5}
\end{center}
\end{figure}

\begin{figure}[H]
\begin{center}
\begin{minipage}{0.75\textwidth}
    \centering
    \includegraphics[width=\textwidth]{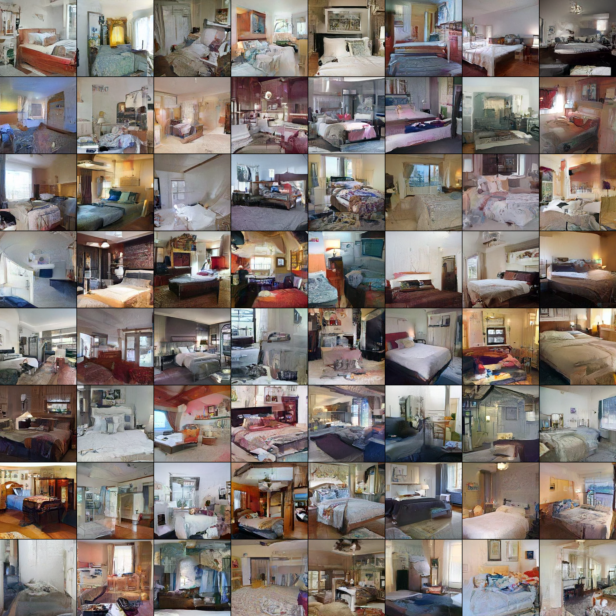}
\end{minipage}

\vspace{1em} 
\rule{0.75\textwidth}{1pt} 
\vspace{1em} 

\begin{minipage}{0.75\textwidth}
    \centering
    \includegraphics[width=\textwidth]{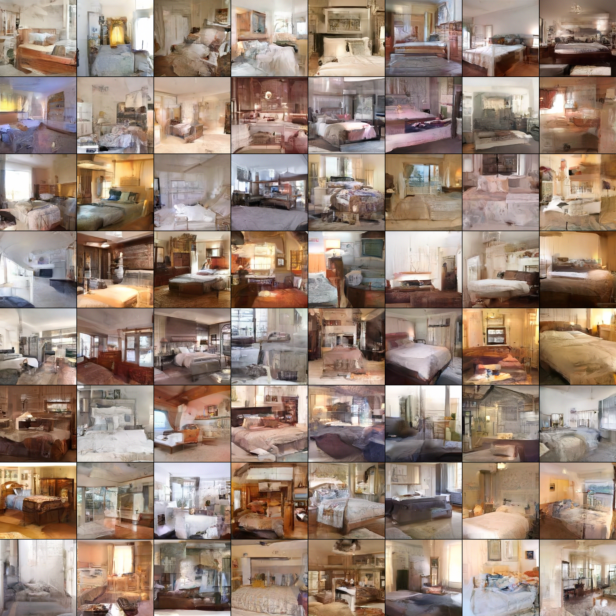}
\end{minipage}%
\end{center}
\caption{Ours (\textbf{up}) and Original (\textbf{down}) 2-rectified flow (2-step, seed 785)}
\label{fig:lsun6}
\end{figure}

We provide additional qualitative results in Figure~\ref{fig:lsun1},\ref{fig:lsun2}, \ref{fig:lsun3}, \ref{fig:lsun4}, \ref{fig:lsun5}, \ref{fig:lsun6} demonstrating that our method achieves better generation quality than the original under few-step sampling settings.

\section{Misc. configurations}
\label{asec:config}
Our experiments on CIFAR10 are configured as follows.
For rectified flow, we utilize the same network architecture as the rectified flow model based on DDPM++ in \citep{song2020score,liu2022flow} with batch size 256. The training process is smoothed using an exponential moving average (EMA) with a decay rate of 0.999999, following the approach in \citet{song2020score}. The Adam optimizer \citep{diederik2014adam} is used with a learning rate of 2e-4 and a dropout rate of 0.15 is applied.

We train RF++$^\dagger$ following the configuration (F) from \citet{lee2024improving}, which includes EDM initialization, an exponential time distribution, and LPIPS-Huber-$1/t$ loss, with a batch size of 128. For training, we use 800K fake pairs for RF++ and 600K fake pairs with an additional 50K real pairs for RF++$^\dagger$(+ours). Both models are optimized using Adam with an EMA decay of 0.9999. We generate 50,000 synthetic images to compute FID using the official RF++ implementation, which employs Heun’s second-order solver \cite{lee2024improving}. For evaluation, we report results using the best checkpoint selected during 300K training iterations. The best FID and IS score obtained with 500K iterations (see Appendix~\ref{asec:aftet500k}) is also based on the same configuration, and evaluated using the official RF++ and guided-diffusion \cite{A3} implementation without modification.

For the high-resolution (LSUN), both our method and the original 2-rectified flow were trained for 600,000 steps with a batch size of 16. Training started from the same 1-rectified flow checkpoints from the official rectified flow repository. During inference, all models used fixed seeds and identical noise conditions corresponding to the sampling steps.

\section{Pseudocode}
\label{asec:algorithm}
\begin{algorithm}[H]
\small
\caption{Full Algorithm (\textcolor{red}{Red} text indicates our addition)}
\label{alg:full}

\KwIn{Coupling $(X_0, X_1)$ of $\pi_0$ and $\pi_1$; velocity model $v_\theta: \mathbb{R}^d \to \mathbb{R}^d$}
\KwOut{$v_{\hat{\theta}_{\text{conic}}}$}

\vspace{0.5em}
\textbf{Training:} \\
$\hat{\theta} = \arg \min_{\theta} \mathbb{E} \left[ \|X_1 - X_0 - v_{\theta}(tX_1 + (1 - t)X_0, t)\|^2 \right]$, where $t \sim \text{Uniform}([0, 1])$

\vspace{0.5em}
\textbf{Sampling:} \\
Draw forward flow: $dZ_{t,F} = v_{\hat{\theta}}(Z_{t,F}, t)\, dt$, starting from $Z_{0,F} \sim \pi_0$ \\
\textcolor{red}{Draw backward flow: $(Z_{0,R}, X_1)$ following $dZ_{t,R} = v_{\hat{\theta}}^{-1}(Z_{t,R}, t)\, dt$, with $X_1 \sim \pi_1$}

\vspace{0.5em}
\textcolor{red}{
\textbf{Balanced Conic Reflow:} \\
Initialize : Total training step = $\mathbb{N}$, Repairing step = $\mathcal{T}$, $cnt = 0$, $\chi_{\text{fake}}$, $\chi_{\text{real}}$}\\
\textcolor{red}{
\For{$t \in (0, \mathbb{N})$}{
    \If{$t == t_{\zeta^{\max}}$}{
        Compute $\zeta^{\max}$ using Eqn.~\ref{mrd}
    }
    \If{$cnt == \mathcal{T}$}{
        Generate new real pair: $(Z_{0,R}, X_1) \leftarrow (v_{\theta_{\text{conic}}}^{-1}(X_1), X_1)$ \\
        $\zeta^{\max}_{k} \leftarrow \zeta^{max}/k$\\
        $cnt \leftarrow 0$
    }
    \Else{
        \textbf{Train:} Update $\hat{\theta}_{\text{conic}}$
        \begin{align*}
        \arg \min_{\theta} \,\, \mathbb{E}\Big[ \|\chi_{\text{fake}} \cdot (\dot{Z}_{t,F} - v_{\theta}(Z_{t,F})) 
        + \chi_{\text{real}} \cdot (X_1 - \text{slerp}(Z_{0,R}, \epsilon, \zeta) 
        - v_{\theta}(\text{Conic}(X_1, \epsilon, \zeta, t)))\|^2 \Big]
        \end{align*}
    }
    cnt+=1
}}
\textbf{(Optional) Distillation:} Learn neural network $\hat{T}$ to distill the $k$-rectified flow, so that $Z^k_{1,F} \approx \hat{T}(Z^k_{0,F})$
\end{algorithm}

\section{Limitation}
\label{asec:limitation}

While our method demonstrates strong improvements in reflow process, one limitation remains in the choice of the maximum Slerp noise magnitude $\zeta^{\text{max}}$, which is dependent on the training dataset. We select $\zeta^{\text{max}}$ by measuring reconstruction discrepancy between real and fake samples after a warm-up phase, and choosing the noise scale that maximizes this gap. While this aligns well with our problem formulation and intuition, providing a theoretically explicit solution remains challenging due to the unknown intrinsic dimension of the training dataset. It may be beneficial to incorporate additional factors into the decision process. For example, from the perspective of information theory, the ratio between real and fake pairs may serve as a guideline by comparing their information content (e.g., entropy), and the intrinsic dimension of the data manifold may also provide useful signals for determining an appropriate noise scale~\cite{A1,A2}.

\newpage
\section{Implementation details}
\label{asec:core_code}

\vspace{0.5em}
\paragraph{Slerp.} Below is the Python implementation of our spherical linear interpolation (Slerp), used to interpolate between the reverse noise $Z_{0,R}$ and a randomly sampled noise $\epsilon \sim \mathcal{N}(0, I)$ with ratio $\zeta$:

\begin{lstlisting}[language=Python, label={lst:slerp}]
def slerp_ours(t, v0, v1, DOT_THRESHOLD=0.9995):
    c = True
    if not isinstance(v0, np.ndarray):
        c = True
        v0 = v0.detach().cpu().numpy()
    if not isinstance(v1, np.ndarray):
        c = True
        v1 = v1.detach().cpu().numpy()

    v0_copy = np.copy(v0)
    v1_copy = np.copy(v1)

    v0 = v0 / np.linalg.norm(v0)
    v1 = v1 / np.linalg.norm(v1)
    dot = np.sum(v0 * v1)

    if np.abs(dot) > DOT_THRESHOLD:
        print("do lerf")
        return torch.lerp(t, v0_copy, v1_copy)

    theta_0 = np.arccos(dot)
    sin_theta_0 = np.sin(theta_0)
    theta_t = theta_0 * t
    sin_theta_t = np.sin(theta_t)

    s0 = np.sin(theta_0 - theta_t) / sin_theta_0
    s1 = sin_theta_t / sin_theta_0
    v2 = s0 * v0_copy + s1 * v1_copy

    return torch.from_numpy(v2).to("cuda") if c else v2
\end{lstlisting}

\paragraph{Noise scheduler.}  
We define the nonlinear Slerp noise schedule as:

\begin{lstlisting}[language=Python,label={lst:scheduler}]
def noise_scheduler(max_steps, max_noise):
    schedule = []
    for step in range(max_steps):
        t = step / max_steps
        noise = max_noise * 2 * (1 - 1 / (1 + t**2))
        schedule.append(noise)
    return schedule
\end{lstlisting}

\paragraph{Scheduled reflow during training.}  
During training, conic reflow is periodically triggered and the noise schedule is dynamically adjusted:

\begin{lstlisting}[language=Python]
if i_effective == warmup_steps:
    max_noise = compute_zeta_max(0.01, 0.5) 
    # Equation (8): starting noise magnitude = 0.01, end = 0.5
(...)
if i_effective % 100000 == 0:
    add_noise = True
    c_n = 0
    patial_max_noise = max_noise / noise_decay
    noise_decay += decay_direction
    # Start decay_direction is -1
    if noise_decay == 1:
        decay_direction = 1
    elif noise_decay == k:
        decay_direction = -1
    alpha = noise_scheduler(max_steps + 5, patial_max_noise)
\end{lstlisting}

\paragraph{Applying conic reflow.}  
At scheduled noise injection steps, we replace $z$ using Slerp between the original $z$ and Gaussian noise:

\begin{lstlisting}[language=Python]
if conic_rf:
    if add_noise:
        x, z, c, _ = next(train_iter_conic)
        noise_batch = torch.randn_like(z).to(device) * (1 - 1e-5)
        z = slerp_ours(alpha[-1 - c_n], z, noise_batch)
        if cnt % gradient_accumulation_steps == 0:
            print('conic reflow(slerp):', alpha[-1 - c_n], 'noise_step:', c_n)
            add_noise = False
(...)
# Sample t, zt
t = sample_t(exponential_distribution, x.shape[0], arg.a).to(device)
zt = (1 - t).view(-1, 1, 1, 1) * x + t.view(-1, 1, 1, 1) * z
target = z - x

# Forward pass
pred = model(zt, t, c)
# Predicted x 
pred_x = zt - pred * t.view(-1, 1, 1, 1)
\end{lstlisting}

\newpage
\section{Unconditional generation model quality}
\label{asec:diffusion}
\begin{table}[H]
    \centering
    \begin{tabular}{lccc}
        \toprule
        \textbf{Method} & \textbf{NFE (↓)} & \textbf{IS (↑)} & \textbf{FID (↓)} \\ \midrule

        \multicolumn{4}{l}{\textbf{\textit{Full Simulation (Euler Solver, N=2000)}}} \\
        VP SDE \citep{song2020score} & 2000 & 9.58 & 2.55 \\
        sub-VP SDE \citep{song2020score} & 2000 & 9.56 & 2.61 \\
        NCSN++ (VE SDE) \citep{song2020score} & 2000 & 9.83 & 2.31 \\
        DDPM \citep{ho2020denoising} & 1000 & 9.46 & 3.21 \\ \midrule

        \multicolumn{4}{l}{\textbf{\textit{Adaptive Step Simulation (Runge–Kutta (RK45), Adaptive N)}}} \\
        VP ODE \citep{song2020score} & 140 & 9.37 & 3.93 \\
        sub-VP ODE \citep{song2020score} & 146 & \textcolor{red}{9.46} & 3.16 \\
        NCSN++ (VE ODE) \citep{song2020score} & 176 & \textcolor{red}{9.35} & 5.38 \\
        LSGM \citep{vahdat2021score} & 147 & - & \textcolor{red}{2.10} \\
        PFGM \citep{xu2022poisson} & 110 & \textcolor{red}{9.68} & \textcolor{red}{2.35} \\
        EDM \citep{karras2022elucidating} & 35 & \textcolor{red}{\textbf{9.84}} & \textcolor{red}{\textbf{2.04}} \\
        1-Rectified Flow & 127 & \textcolor{red}{9.60} & \textcolor{red}{2.58} \\
        2-Rectified Flow & 110 & 9.24 & 3.36 \\
        \textbf{2-Rectified Flow Ours} & 104 & 9.30 & 3.24 \\
        3-Rectified Flow & 104 & 9.01 & 3.96 \\
        \textbf{3-Rectified Flow Ours} & 98 & 9.14 & 3.70 \\ 
        FM-OT \citep{lipman2023flowmatchinggenerativemodeling} &142&-&6.35 \\
        OT-CFM \citep{tong2024improvinggeneralizingflowbasedgenerative} &100&-&4.44 \\
        Simple ReFlow \citep{kim2024simple} &9&-&\textcolor{red}{2.23}\\
        
        \midrule

        \multicolumn{4}{l}{\textbf{\textit{One-Step Simulation (Euler Solver, N=1)}}} \\
        VP ODE (\textit{+Distill}) \citep{song2020score}  & 1 & 1.20 (8.73) & 451 (16.23) \\
        sub-VP ODE (\textit{+Distill}) \citep{song2020score}  & 1 & 1.21 (8.80) & 451 (14.32) \\
        NCSN++ (VE ODE) (\textit{+Distill}) \citep{song2020score}  & 1 & 1.18 (2.57) & 461 (254) \\
        1-Rectified Flow (\textit{+Distill}) & 1 & 1.13 ({9.08}) & 378 (6.18) \\
        2-Rectified Flow (\textit{+Distill}) & 1 & 8.08 (9.01) & 12.21 (4.85) \\
        2-Rectified Flow++$^\dagger$\citep{lee2024improving}& 1 & 9.03 & 4.14 \\
        2-Rectified Flow++$^\dagger$\textbf{(+Ours)}& 1 & \textcolor{blue}{9.15} & 3.84 \\
        \textbf{2-Rectified Flow Ours (\textit{+Distill})} & 1 & 8.79 (9.11) & 5.98 (4.16) \\
        3-Rectified Flow (\textit{+Distill}) & 1 & 8.47 (8.79) & 8.15 (5.21) \\
        \textbf{3-Rectified Flow Ours (\textit{+Distill})} & 1 & 8.84 (8.96) & 5.48 (4.68) \\ 
        \midrule

        \multicolumn{4}{l}{\textbf{\textit{Diffusion + Distillation}}} \\
        DDIM Distillation \citep{luhman2021knowledge} & 1 & 8.36 & 9.36 \\
        DMD \citep{yin2024one} & 1 & - & 3.77 \\
        Diff-Instruct \citep{luo2024diff} & 1 & \textcolor{blue}{9.89} & 4.53 \\
        PD \citep{salimans2022progressive} & 1 & 8.69 & 8.34 \\ 
        DFNO \citep{zheng2023fast} & 1 & - & 4.15\\
        SlimFlow (EDM teacher) \citep{zhu2024slimflowtrainingsmalleronestep}&1&-&4.53\\
        SlimFlow (1-Rectified flow teacher)&1&-&5.81\\
        SID, $(\alpha=1.2)$ \citep{zhou2024score} & 1 & \textcolor{blue}{\textbf{9.98}} & \textcolor{blue}{\textbf{1.92}}\\\midrule

        \multicolumn{4}{l}{\textbf{\textit{Consistency Model}}} \\
        CD \citep{song2023consistency} & 1 & \textcolor{blue}{9.48} & 3.55 \\
        CT \citep{song2023consistency} & 1 & 8.49 & 8.70 \\
        ICT \citep{song2023improved}& 1 & \textcolor{blue}{9.54} & \textcolor{blue}{2.83} \\
        ECT \citep{geng2024ect} &1&-&3.60\\
        sCT \citep{lu2025sct}&1&-&\textcolor{blue}{2.97}\\
        IMM \citep{imm}&1&-&3.40\\
        MeanFlow \citep{geng2025meanflowsonestepgenerative} &1&-&\textcolor{blue}{2.92}\\
        CTM \citep{kim2023consistency} & 1 & - & 5.19 \\
        CTM + GAN \citep{kim2023consistency} & 1 & - & \textcolor{blue}{1.98} \\

        \bottomrule
    \end{tabular}
    \caption{Unconditional generation quality with various diffusion-based models on CIFAR-10. \textcolor{blue}{Blue} rows highlight the top-5 baselines for 1-NFE, and \textcolor{red}{red} rows for Adaptive NFE (RK-45). The lowest FID and highest IS scores in each setting are \textbf{bolded}.}
    \label{tab:method_comparison}
\end{table}

\vspace{-10em}

\centering

\centering







\newpage

\small
\bibliographystyle{plainnat}
\bibliography{references}

\begin{thebibliography}{57}
\providecommand{\natexlab}[1]{#1}
\providecommand{\url}[1]{\texttt{#1}}
\expandafter\ifx\csname urlstyle\endcsname\relax
  \providecommand{\doi}[1]{doi: #1}\else
  \providecommand{\doi}{doi: \begingroup \urlstyle{rm}\Url}\fi

\bibitem[Adler et~al.(2021)Adler, Kobler, Lunz, Sch{\"o}nlieb, and Arridge]{adler2021task}
Jonas Adler, Emil Kobler, Sebastian Lunz, Carola-Bibiane Sch{\"o}nlieb, and Simon Arridge.
\newblock Task adapted reconstruction for inverse problems.
\newblock In \emph{International Conference on Machine Learning}, pages 74--84. PMLR, 2021.

\bibitem[Ambrosio et~al.(2008)Ambrosio, Crippa, De~Lellis, Otto, Westdickenberg, Ambrosio, and Crippa]{ambrosio2008existence}
Luigi Ambrosio, Gianluca Crippa, Camillo De~Lellis, Felix Otto, Michael Westdickenberg, Luigi Ambrosio, and Gianluca Crippa.
\newblock Existence, uniqueness, stability and differentiability properties of the flow associated to weakly differentiable vector fields.
\newblock \emph{Transport equations and multi-D hyperbolic conservation laws}, pages 3--57, 2008.

\bibitem[Antun et~al.(2020)Antun, Renna, Poon, Adcock, and Hansen]{antun2020instabilities}
Velin Antun, Francesco Renna, Clarice Poon, Ben Adcock, and Anders~C Hansen.
\newblock On instabilities of deep learning in image reconstruction and the potential costs of ai.
\newblock \emph{Proceedings of the National Academy of Sciences}, 117\penalty0 (48):\penalty0 30088--30095, 2020.

\bibitem[Candes et~al.(2006)Candes, Romberg, and Tao]{candes2006stable}
Emmanuel~J Candes, Justin~K Romberg, and Terence Tao.
\newblock Stable signal recovery from incomplete and inaccurate measurements.
\newblock \emph{Communications on Pure and Applied Mathematics: A Journal Issued by the Courant Institute of Mathematical Sciences}, 59\penalty0 (8):\penalty0 1207--1223, 2006.

\bibitem[Chen et~al.(2023)Chen, Lee, and Lu]{chen2023improved}
Hongrui Chen, Holden Lee, and Jianfeng Lu.
\newblock Improved analysis of score-based generative modeling: User-friendly bounds under minimal smoothness assumptions.
\newblock In \emph{International Conference on Machine Learning}, pages 4735--4763. PMLR, 2023.

\bibitem[De~Bortoli et~al.(2021)De~Bortoli, Thornton, Heng, and Doucet]{de2021diffusion}
Valentin De~Bortoli, James Thornton, Jeremy Heng, and Arnaud Doucet.
\newblock Diffusion schr{\"o}dinger bridge with applications to score-based generative modeling.
\newblock \emph{Advances in Neural Information Processing Systems}, 34:\penalty0 17695--17709, 2021.

\bibitem[Deng et~al.(2009)Deng, Dong, Socher, Li, Li, and Fei-Fei]{deng2009imagenet}
Jia Deng, Wei Dong, Richard Socher, Li-Jia Li, Kai Li, and Li~Fei-Fei.
\newblock Imagenet: A large-scale hierarchical image database.
\newblock In \emph{2009 IEEE Conference on Computer Vision and Pattern Recognition (CVPR)}, pages 248--255. IEEE, 2009.

\bibitem[Diederik(2014)]{diederik2014adam}
P~Kingma Diederik.
\newblock Adam: A method for stochastic optimization.
\newblock \emph{(No Title)}, 2014.

\bibitem[Esser et~al.(2024)Esser, Kulal, Blattmann, Entezari, M{\"u}ller, Saini, Levi, Lorenz, Sauer, Boesel, et~al.]{sd3}
Patrick Esser, Sumith Kulal, Andreas Blattmann, Rahim Entezari, Jonas M{\"u}ller, Harry Saini, Yam Levi, Dominik Lorenz, Axel Sauer, Frederic Boesel, et~al.
\newblock Scaling rectified flow transformers for high-resolution image synthesis.
\newblock In \emph{Forty-first International Conference on Machine Learning}, 2024.

\bibitem[Figalli and Glaudo(2021)]{OT3}
Alessio Figalli and Federico Glaudo.
\newblock An invitation to optimal transport, wasserstein distances, and gradient flows.
\newblock \emph{Springer}, 2021.

\bibitem[Flamary et~al.(2016)Flamary, Courty, Tuia, and Rakotomamonjy]{OT4}
R~Flamary, N~Courty, D~Tuia, and A~Rakotomamonjy.
\newblock Optimal transport for domain adaptation.
\newblock \emph{IEEE Trans. Pattern Anal. Mach. Intell}, 2016.

\bibitem[Geng et~al.(2024)Geng, Pokle, Luo, Lin, and Kolter]{geng2024ect}
Zhengyang Geng, Ashwini Pokle, William Luo, Justin Lin, and J.~Zico Kolter.
\newblock Consistency models made easy, 2024.
\newblock URL \url{https://arxiv.org/abs/2406.14548}.

\bibitem[Geng et~al.(2025)Geng, Deng, Bai, Kolter, and He]{geng2025meanflowsonestepgenerative}
Zhengyang Geng, Mingyang Deng, Xingjian Bai, J.~Zico Kolter, and Kaiming He.
\newblock Mean flows for one-step generative modeling, 2025.
\newblock URL \url{https://arxiv.org/abs/2505.13447}.

\bibitem[Heusel et~al.(2017)Heusel, Ramsauer, Unterthiner, Nessler, and Hochreiter]{heusel2017gans}
Martin Heusel, Hubert Ramsauer, Thomas Unterthiner, Bernhard Nessler, and Sepp Hochreiter.
\newblock Gans trained by a two time-scale update rule converge to a local nash equilibrium.
\newblock \emph{Advances in neural information processing systems}, 30, 2017.

\bibitem[Ho et~al.(2020)Ho, Jain, and Abbeel]{ho2020denoising}
Jonathan Ho, Ajay Jain, and Pieter Abbeel.
\newblock Denoising diffusion probabilistic models.
\newblock \emph{Advances in Neural Information Processing Systems}, 33:\penalty0 6840--6851, 2020.

\bibitem[Jang et~al.(2024)Jang, Huynh, Shah, Chen, and Lim]{jang2024spherical}
Young~Kyun Jang, Dat Huynh, Ashish Shah, Wen-Kai Chen, and Ser-Nam Lim.
\newblock Spherical linear interpolation and text-anchoring for zero-shot composed image retrieval.
\newblock \emph{arXiv preprint arXiv:2405.00571}, 2024.

\bibitem[Karras et~al.(2022)Karras, Aittala, Aila, and Laine]{karras2022elucidating}
Tero Karras, Miika Aittala, Timo Aila, and Samuli Laine.
\newblock Elucidating the design space of diffusion-based generative models.
\newblock \emph{arXiv preprint arXiv:2206.00364}, 2022.

\bibitem[Kim et~al.(2024)Kim, Hsieh, Klein, Cuturi, Ye, Kawar, and Thornton]{kim2024simple}
Beomsu Kim, Yu-Guan Hsieh, Michal Klein, Marco Cuturi, Jong~Chul Ye, Bahjat Kawar, and James Thornton.
\newblock Simple reflow: Improved techniques for fast flow models.
\newblock \emph{arXiv preprint arXiv:2410.07815}, 2024.

\bibitem[Kim et~al.(2023)Kim, Lai, Liao, Murata, Takida, Uesaka, He, Mitsufuji, and Ermon]{kim2023consistency}
Dongjun Kim, Chieh-Hsin Lai, Wei-Hsiang Liao, Naoki Murata, Yuhta Takida, Toshimitsu Uesaka, Yutong He, Yuki Mitsufuji, and Stefano Ermon.
\newblock Consistency trajectory models: Learning probability flow ode trajectory of diffusion.
\newblock \emph{arXiv preprint arXiv:2310.02279}, 2023.

\bibitem[Krizhevsky et~al.(2009)Krizhevsky, Hinton, et~al.]{krizhevsky2009learning}
Alex Krizhevsky, Geoffrey Hinton, et~al.
\newblock Learning multiple layers of features from tiny images.
\newblock Technical report, University of Toronto, Toronto, ON, Canada, 2009.
\newblock URL \url{https://www.cs.toronto.edu/~kriz/learning-features-2009-TR.pdf}.
\newblock Technical report.

\bibitem[Kurtz(2011)]{kurtz2011equivalence}
Thomas~G Kurtz.
\newblock Equivalence of stochastic equations and martingale problems.
\newblock \emph{Stochastic analysis 2010}, pages 113--130, 2011.

\bibitem[Kynk{\"a}{\"a}nniemi et~al.(2019)Kynk{\"a}{\"a}nniemi, Karras, Laine, Lehtinen, and Aila]{kynkaanniemi2019improved}
Tuomas Kynk{\"a}{\"a}nniemi, Tero Karras, Samuli Laine, Jaakko Lehtinen, and Timo Aila.
\newblock Improved precision and recall metric for assessing generative models.
\newblock \emph{Advances in neural information processing systems}, 32, 2019.

\bibitem[Lee et~al.(2023)Lee, Kim, and Ye]{lee2023minimizing}
Sangyun Lee, Beomsu Kim, and Jong~Chul Ye.
\newblock Minimizing trajectory curvature of ode-based generative models.
\newblock In \emph{International Conference on Machine Learning}, pages 18957--18973. PMLR, 2023.

\bibitem[Lee et~al.(2024)Lee, Lin, and Fanti]{lee2024improving}
Sangyun Lee, Zinan Lin, and Giulia Fanti.
\newblock Improving the training of rectified flows.
\newblock \emph{arXiv preprint arXiv:2405.20320}, 2024.

\bibitem[Li et~al.(2024)Li, Chu, Shi, and Wang]{li2024flowdreamer}
Hangyu Li, Xiangxiang Chu, Dingyuan Shi, and Lin Wang.
\newblock Flowdreamer: exploring high fidelity text-to-3d generation via rectified flow, 2024.
\newblock URL \url{https://arxiv.org/abs/2408.05008}.

\bibitem[Lipman et~al.(2023)Lipman, Chen, Ben-Hamu, Nickel, and Le]{lipman2023flowmatchinggenerativemodeling}
Yaron Lipman, Ricky T.~Q. Chen, Heli Ben-Hamu, Maximilian Nickel, and Matt Le.
\newblock Flow matching for generative modeling, 2023.
\newblock URL \url{https://arxiv.org/abs/2210.02747}.

\bibitem[Liu et~al.(2022)Liu, Gong, and Liu]{liu2022flow}
Xingchao Liu, Chengyue Gong, and Qiang Liu.
\newblock Flow straight and fast: Learning to generate and transfer data with rectified flow.
\newblock \emph{arXiv preprint arXiv:2209.03003}, 2022.

\bibitem[Liu et~al.(2023)Liu, Zhang, Ma, Peng, et~al.]{liu2023instaflow}
Xingchao Liu, Xiwen Zhang, Jianzhu Ma, Jian Peng, et~al.
\newblock Instaflow: One step is enough for high-quality diffusion-based text-to-image generation.
\newblock In \emph{The Twelfth International Conference on Learning Representations}, 2023.

\bibitem[Lu and Song(2025)]{lu2025sct}
Cheng Lu and Yang Song.
\newblock Simplifying, stabilizing and scaling continuous-time consistency models, 2025.
\newblock URL \url{https://arxiv.org/abs/2410.11081}.

\bibitem[Luhman and Luhman(2021)]{luhman2021knowledge}
Eric Luhman and Troy Luhman.
\newblock Knowledge distillation in iterative generative models for improved sampling speed.
\newblock \emph{arXiv preprint arXiv:2101.02388}, 2021.

\bibitem[Luo et~al.(2024)Luo, Hu, Zhang, Sun, Li, and Zhang]{luo2024diff}
Weijian Luo, Tianyang Hu, Shifeng Zhang, Jiacheng Sun, Zhenguo Li, and Zhihua Zhang.
\newblock Diff-instruct: A universal approach for transferring knowledge from pre-trained diffusion models.
\newblock \emph{Advances in Neural Information Processing Systems}, 36, 2024.

\bibitem[Pedregosa et~al.(2011)Pedregosa, Varoquaux, Gramfort, Michel, Thirion, Grisel, Blondel, Prettenhofer, Weiss, Dubourg, Vanderplas, Passos, Cournapeau, Brucher, Perrot, and Duchesnay]{scikit-learn}
Fabian Pedregosa, Gael Varoquaux, Alexandre Gramfort, Vincent Michel, Bertrand Thirion, Olivier Grisel, Mathieu Blondel, Peter Prettenhofer, Ron Weiss, Vincent Dubourg, Jake Vanderplas, Alexandre Passos, David Cournapeau, Matthieu Brucher, Matthieu Perrot, and \'E.douard Duchesnay.
\newblock Scikit-learn: Machine learning in python.
\newblock \emph{Journal of Machine Learning Research}, 12:\penalty0 2825--2830, 2011.

\bibitem[Ravishankar et~al.(2019)Ravishankar, Ye, and Fessler]{ravishankar2019image}
Saiprasad Ravishankar, Jong~Chul Ye, and Jeffrey~A Fessler.
\newblock Image reconstruction: From sparsity to data-adaptive methods and machine learning.
\newblock \emph{Proceedings of the IEEE}, 108\penalty0 (1):\penalty0 86--109, 2019.

\bibitem[Salimans and Ho(2022)]{salimans2022progressive}
Tim Salimans and Jonathan Ho.
\newblock Progressive distillation for fast sampling of diffusion models.
\newblock \emph{arXiv preprint arXiv:2202.00512}, 2022.

\bibitem[Song et~al.(2020{\natexlab{a}})Song, Meng, and Ermon]{song2020denoising}
Jiaming Song, Chenlin Meng, and Stefano Ermon.
\newblock Denoising diffusion implicit models.
\newblock \emph{arXiv preprint arXiv:2010.02502}, 2020{\natexlab{a}}.

\bibitem[Song and Dhariwal(2023)]{song2023improved}
Yang Song and Prafulla Dhariwal.
\newblock Improved techniques for training consistency models.
\newblock \emph{arXiv preprint arXiv:2310.14189}, 2023.

\bibitem[Song et~al.(2020{\natexlab{b}})Song, Sohl-Dickstein, Kingma, Kumar, Ermon, and Poole]{song2020score}
Yang Song, Jascha Sohl-Dickstein, Diederik~P Kingma, Abhishek Kumar, Stefano Ermon, and Ben Poole.
\newblock Score-based generative modeling through stochastic differential equations.
\newblock \emph{arXiv preprint arXiv:2011.13456}, 2020{\natexlab{b}}.

\bibitem[Song et~al.(2023)Song, Dhariwal, Chen, and Sutskever]{song2023consistency}
Yang Song, Prafulla Dhariwal, Mark Chen, and Ilya Sutskever.
\newblock Consistency models.
\newblock \emph{arXiv preprint arXiv:2303.01469}, 2023.

\bibitem[Stetter et~al.(1973)]{stetter1973analysis}
Hans~J Stetter et~al.
\newblock \emph{Analysis of discretization methods for ordinary differential equations}, volume~23.
\newblock Springer, 1973.

\bibitem[Tong et~al.(2024)Tong, Fatras, Malkin, Huguet, Zhang, Rector-Brooks, Wolf, and Bengio]{tong2024improvinggeneralizingflowbasedgenerative}
Alexander Tong, Kilian Fatras, Nikolay Malkin, Guillaume Huguet, Yanlei Zhang, Jarrid Rector-Brooks, Guy Wolf, and Yoshua Bengio.
\newblock Improving and generalizing flow-based generative models with minibatch optimal transport, 2024.
\newblock URL \url{https://arxiv.org/abs/2302.00482}.

\bibitem[Tzen and Raginsky(2019)]{tzen2019theoretical}
Belinda Tzen and Maxim Raginsky.
\newblock Theoretical guarantees for sampling and inference in generative models with latent diffusions.
\newblock In \emph{Conference on Learning Theory}, pages 3084--3114. PMLR, 2019.

\bibitem[Vahdat et~al.(2021)Vahdat, Kreis, and Kautz]{vahdat2021score}
Arash Vahdat, Karsten Kreis, and Jan Kautz.
\newblock Score-based generative modeling in latent space.
\newblock \emph{Advances in Neural Information Processing Systems}, 34:\penalty0 11287--11302, 2021.

\bibitem[Vargas et~al.(2021)Vargas, Thodoroff, Lamacraft, and Lawrence]{vargas2021solving}
Francisco Vargas, Pierre Thodoroff, Austen Lamacraft, and Neil Lawrence.
\newblock Solving schr{\"o}dinger bridges via maximum likelihood.
\newblock \emph{Entropy}, 23\penalty0 (9):\penalty0 1134, 2021.

\bibitem[Villani(2009)]{OT2}
Cedric Villani.
\newblock Optimal transport: old and new.
\newblock \emph{Springer}, 2009.

\bibitem[Villani(2021)]{OT1}
Cedric Villani.
\newblock Topics in optimal transportation.
\newblock \emph{American Mathematical Soc}, 2021.

\bibitem[Virtanen et~al.(2020)Virtanen, Gommers, Oliphant, Haberland, Reddy, Cournapeau, Burovski, Peterson, Weckesser, Bright, et~al.]{virtanen2020scipy}
Pauli Virtanen, Ralf Gommers, Travis~E Oliphant, Matt Haberland, Tyler Reddy, David Cournapeau, Evgeni Burovski, Pearu Peterson, Warren Weckesser, Jonathan Bright, et~al.
\newblock Scipy 1.0: fundamental algorithms for scientific computing in python.
\newblock \emph{Nature methods}, 17\penalty0 (3):\penalty0 261--272, 2020.

\bibitem[Wang and Golland(2023)]{wang2023interpolating}
Clinton Wang and Polina Golland.
\newblock Interpolating between images with diffusion models, 2023.
\newblock URL \url{https://openreview.net/pdf?id=L2D9Gybx0P}.
\newblock OpenReview preprint.

\bibitem[Wang et~al.(2024)Wang, Yang, Huang, Wang, and Li]{wang2024rectified}
Fu-Yun Wang, Ling Yang, Zhaoyang Huang, Mengdi Wang, and Hongsheng Li.
\newblock Rectified diffusion: Straightness is not your need in rectified flow.
\newblock \emph{arXiv preprint arXiv:2410.07303}, 2024.

\bibitem[Xu et~al.(2022)Xu, Liu, Tegmark, and Jaakkola]{xu2022poisson}
Yilun Xu, Ziming Liu, Max Tegmark, and Tommi Jaakkola.
\newblock Poisson flow generative models.
\newblock \emph{Advances in Neural Information Processing Systems}, 35:\penalty0 16782--16795, 2022.

\bibitem[Yan et~al.(2024)Yan, Liu, Pan, Liew, Liu, and Feng]{yan2024perflowpiecewiserectifiedflow}
Hanshu Yan, Xingchao Liu, Jiachun Pan, Jun~Hao Liew, Qiang Liu, and Jiashi Feng.
\newblock Perflow: Piecewise rectified flow as universal plug-and-play accelerator, 2024.
\newblock URL \url{https://arxiv.org/abs/2405.07510}.

\bibitem[Yin et~al.(2024)Yin, Gharbi, Zhang, Shechtman, Durand, Freeman, and Park]{yin2024one}
Tianwei Yin, Micha{\"e}l Gharbi, Richard Zhang, Eli Shechtman, Fredo Durand, William~T Freeman, and Taesung Park.
\newblock One-step diffusion with distribution matching distillation.
\newblock In \emph{Proceedings of the IEEE/CVF Conference on Computer Vision and Pattern Recognition}, pages 6613--6623, 2024.

\bibitem[Yu et~al.(2015)Yu, Seff, Zhang, Song, Funkhouser, and Xiao]{yu2015lsun}
Fisher Yu, Ari Seff, Yinda Zhang, Shuran Song, Thomas Funkhouser, and Jianxiong Xiao.
\newblock Lsun: Construction of a large-scale image dataset using deep learning with humans in the loop.
\newblock \emph{arXiv preprint arXiv:1506.03365}, 2015.

\bibitem[Zheng et~al.(2023)Zheng, Nie, Vahdat, Azizzadenesheli, and Anandkumar]{zheng2023fast}
Hongkai Zheng, Weili Nie, Arash Vahdat, Kamyar Azizzadenesheli, and Anima Anandkumar.
\newblock Fast sampling of diffusion models via operator learning.
\newblock In \emph{International conference on machine learning}, pages 42390--42402. PMLR, 2023.

\bibitem[Zhou et~al.(2025)Zhou, Ermon, and Song]{imm}
Linqi Zhou, Stefano Ermon, and Jiaming Song.
\newblock Inductive moment matching, 2025.
\newblock URL \url{https://arxiv.org/abs/2503.07565}.

\bibitem[Zhou et~al.(2024)Zhou, Zheng, Wang, Yin, and Huang]{zhou2024score}
Mingyuan Zhou, Huangjie Zheng, Zhendong Wang, Mingzhang Yin, and Hai Huang.
\newblock Score identity distillation: Exponentially fast distillation of pretrained diffusion models for one-step generation.
\newblock In \emph{Forty-first International Conference on Machine Learning}, 2024.

\bibitem[Zhu et~al.(2024{\natexlab{a}})Zhu, Wang, Ding, Qu, and Zhu]{zhu2024analyzing}
Huminhao Zhu, Fangyikang Wang, Tianyu Ding, Qing Qu, and Zhihui Zhu.
\newblock Analyzing and improving model collapse in rectified flow models.
\newblock \emph{arXiv preprint arXiv:2412.08175}, 2024{\natexlab{a}}.

\bibitem[Zhu et~al.(2024{\natexlab{b}})Zhu, Liu, and Liu]{zhu2024slimflowtrainingsmalleronestep}
Yuanzhi Zhu, Xingchao Liu, and Qiang Liu.
\newblock Slimflow: Training smaller one-step diffusion models with rectified flow, 2024{\natexlab{b}}.
\newblock URL \url{https://arxiv.org/abs/2407.12718}.

\end{thebibliography}


\begin{thebibliography}{99}
\bibitem[A1]{A1}
Stanczuk, Jan Pawel, Georgios Batzolis, Teo Deveney, and Carola-Bibiane Schönlieb.  
\textit{Diffusion models encode the intrinsic dimension of data manifolds}.  
International Conference on Machine Learning (ICML), 2024.
\bibitem[A2]{A2}
Mingi Kwon, Shin seong Kim, Jaeseok Jeong. Yi Ting Hsiao, and Youngjung Uh. Tcfg: \textit{Tangential damping classifier-free guidance}. 2025. URL https://arxiv.org/abs/2503.18137.
\bibitem[A3]{A3}
Dhariwal, Prafulla and Alexander Nichol.  
\textit{Diffusion models beat GANs on image synthesis}.  
Advances in Neural Information Processing Systems (NeurIPS), 34:8780--8794, 2021.

\end{thebibliography}

\renewcommand{\refname}{Additional References} 

\newpage
\section*{NeurIPS Paper Checklist}
\begin{enumerate}

\item {\bf Claims}
    \item[] Question: Do the main claims made in the abstract and introduction accurately reflect the paper's contributions and scope?
    \item[] Answer: \answerYes{} 
    \item[] Justification: We have summarized our contributions at the end of the introduction.

\item {\bf Limitations}
    \item[] Question: Does the paper discuss the limitations of the work performed by the authors?
    \item[] Answer: \answerYes{} 
    \item[] Justification: See Appendix~\ref{asec:limitation}.

\item {\bf Theory assumptions and proofs}
    \item[] Question: For each theoretical result, does the paper provide the full set of assumptions and a complete (and correct) proof?
    \item[] Answer: \answerNA{} 
    \item[] Justification: The paper does not present formal theoretical results or proofs. Instead, we focus on empirically analyzing and mitigating distributional drift during the reflow process. 
    
\item {\bf Experimental result reproducibility}
    \item[] Question: Does the paper fully disclose all the information needed to reproduce the main experimental results of the paper to the extent that it affects the main claims and/or conclusions of the paper (regardless of whether the code and data are provided or not)?
    \item[] Answer: \answerYes{} 
    \item[] Justification: We describe the full training mechanism of our method, including hyperparameter settings, in the Section~\ref{sec:detail} and Appendix~\ref{asec:settings}. Additionally, we provide the pseudo algorithm and detailed training configurations for each dataset in the Appendix~\ref{asec:algorithm}.

\item {\bf Open access to data and code}
    \item[] Question: Does the paper provide open access to the data and code, with sufficient instructions to faithfully reproduce the main experimental results, as described in supplemental material?
    \item[] Answer: \answerYes{} 
    \item[] Justification: We provide a partial implementation with core components in Appendix~\ref{asec:core_code}, and plan to release the full codebase and instructions by the camera-ready deadline.

\item {\bf Experimental setting/details}
    \item[] Question: Does the paper specify all the training and test details (e.g., data splits, hyperparameters, how they were chosen, type of optimizer, etc.) necessary to understand the results?
    \item[] Answer: \answerYes{} 
    \item[] Justification: We provide clear guidelines for choosing the perturbation magnitude in the Section~\ref{sec:detail}, and describe all model configurations including the optimizer in both the Experiment Setup Section~\ref{sec:experiments} and the Appendix~\ref{asec:config}.

\item {\bf Experiment statistical significance}
    \item[] Question: Does the paper report error bars suitably and correctly defined or other appropriate information about the statistical significance of the experiments?
    \item[] Answer: \answerYes{} 
    \item[] Justification: We explicitly state the number of data points used to compute the mean values of curvature, initial velocity delta, and reconstruction error in Section~\ref{sec:experiments} and Appendix~\ref{asec:config}. The magnitude of the epsilon used to calculate the perturbed reconstruction error is also clearly described. Additionally, we specify the number of synthetic samples used for evaluating generation quality metrics such as recall, precision, and FID.

\item {\bf Experiments compute resources}
    \item[] Question: For each experiment, does the paper provide sufficient information on the computer resources (type of compute workers, memory, time of execution) needed to reproduce the experiments?
    \item[] Answer: \answerYes{} 
    \item[] Justification: We provide the details of our CPU and GPU resources in the appendix~\ref{asec:settings}. Additionally, the number of training iterations, batch size, and the number of fake and real samples used in each experiment are specified in the experiment section~\ref{sec:experiments} and Appendix~\ref{asec:config}.
    
\item {\bf Code of ethics}
    \item[] Question: Does the research conducted in the paper conform, in every respect, with the NeurIPS Code of Ethics \url{https://neurips.cc/public/EthicsGuidelines}?
    \item[] Answer: \answerYes{} 
    \item[] Justification: We confirm that the research conducted in this paper adheres to the NeurIPS Code of Ethics.

\item {\bf Broader impacts}
    \item[] Question: Does the paper discuss both potential positive societal impacts and negative societal impacts of the work performed?
    \item[] Answer: \answerNo{}{} 
    \item[] Justification: This work analyzes distributional drift in reflow and proposes a technical solution for flow-based generative models. Since it does not involve application-specific deployment or sensitive content, we found no broader societal impacts to discuss.
    
\item {\bf Safeguards}
    \item[] Question: Does the paper describe safeguards that have been put in place for responsible release of data or models that have a high risk for misuse (e.g., pretrained language models, image generators, or scraped datasets)?
    \item[] Answer: \answerNA{} 
    \item[] Justification: : This work does not involve any models or datasets that pose a foreseeable risk of misuse or harm.

\item {\bf Licenses for existing assets}
    \item[] Question: Are the creators or original owners of assets (e.g., code, data, models), used in the paper, properly credited and are the license and terms of use explicitly mentioned and properly respected?
    \item[] Answer: \answerYes{} 
    \item[] Justification: We used the official repositories of existing flow models \citep{liu2022flow} and clearly cited the datasets used for training (CIFAR-10 \citep{krizhevsky2009learning}, LSUN \citep{yu2015lsun}, ImageNet \citep{deng2009imagenet}). We also specified the version of the Runge-Kutta sampling tool (from SciPy ) used during sampling \citep{virtanen2020scipy}.

\item {\bf New assets}
    \item[] Question: Are new assets introduced in the paper well documented and is the documentation provided alongside the assets?
    \item[] Answer: \answerNA{} 
    \item[] Justification: We do not introduce any new assets. All experiments are conducted using widely used public datasets such as CIFAR-10, ImageNet, and LSUN Bedroom \citep{krizhevsky2009learning, yu2015lsun, deng2009imagenet}.

\item {\bf Crowdsourcing and research with human subjects}
    \item[] Question: For crowdsourcing experiments and research with human subjects, does the paper include the full text of instructions given to participants and screenshots, if applicable, as well as details about compensation (if any)? 
    \item[] Answer: \answerNA{} 
    \item[] Justification: This paper does not involve crowdsourcing or research with human subjects.

\item {\bf Institutional review board (IRB) approvals or equivalent for research with human subjects}
    \item[] Question: Does the paper describe potential risks incurred by study participants, whether such risks were disclosed to the subjects, and whether Institutional Review Board (IRB) approvals (or an equivalent approval/review based on the requirements of your country or institution) were obtained?
    \item[] Answer: \answerNA{} 
    \item[] Justification:  This paper does not involve crowdsourcing nor research with human subjects.

\item {\bf Declaration of LLM usage}
    \item[] Question: Does the paper describe the usage of LLMs if it is an important, original, or non-standard component of the core methods in this research? Note that if the LLM is used only for writing, editing, or formatting purposes and does not impact the core methodology, scientific rigorousness, or originality of the research, declaration is not required.
    \item[] Answer: \answerNA{} 
    \item[] Justification: We did not use LLMs for any part of the core methodology or scientific contribution of the paper.

\end{enumerate}


\end{document}